\documentclass{article}

\usepackage{algorithm}
\usepackage{algorithmic}
\usepackage{textcomp}
\usepackage{csquotes}
\usepackage{microtype}
\usepackage{graphicx}
\usepackage{booktabs}
\usepackage{hyperref}
\usepackage{amsmath}
\usepackage{amssymb}
\usepackage{mathtools}
\usepackage{amsthm}
\usepackage[capitalize,noabbrev]{cleveref}
\usepackage[title]{appendix}
\usepackage{subcaption}
\usepackage{wrapfig}
\usepackage[utf8]{inputenc} 
\usepackage[T1]{fontenc}    
\usepackage{url}            
\usepackage{xspace}
\usepackage[export]{adjustbox}
\usepackage{amsfonts}       
\usepackage{nicefrac}       
\usepackage{microtype}      
\usepackage{wasysym}
\usepackage{enumitem}
\usepackage[american]{babel}
\usepackage{multirow}
\usepackage{adjustbox}
\usepackage{hyperref}       
\usepackage{booktabs}       
\usepackage{xcolor}         
\usepackage{longtable}

\usepackage[accepted]{icml2025}

\usepackage{amsmath,amsfonts,bm}

\def\eqref#1{equation~\ref{#1}}

\def\1{\bm{1}}

\def\vx{{\bm{x}}}
\def\vy{{\bm{y}}}

\def\mX{{\bm{X}}}
\def\mY{{\bm{Y}}}

\DeclareMathAlphabet{\mathsfit}{\encodingdefault}{\sfdefault}{m}{sl}
\SetMathAlphabet{\mathsfit}{bold}{\encodingdefault}{\sfdefault}{bx}{n}

\def\gD{{\mathcal{D}}}

\def\gN{{\mathcal{N}}}

\def\gX{{\mathcal{X}}}
\def\gY{{\mathcal{Y}}}

\def\sD{{\mathbb{D}}}

\def\sR{{\mathbb{R}}}

\newcommand{\R}{\mathbb{R}}

\DeclareMathOperator*{\argmax}{arg\,max}

\newcommand{\method}[1]{\textsc{#1}}
\def\ours{{\method{TabPTM}}\xspace}

\icmltitlerunning{Rethinking Pre-Training in Tabular Data: A Neighborhood Embedding Perspective}

\theoremstyle{plain}

\theoremstyle{definition}

\theoremstyle{remark}

\makeatletter
\DeclareRobustCommand\onedot{\futurelet\@let@token\@onedot}
\def\@onedot{\ifx\@let@token.\else.\null\fi\xspace}

\def\eg{\emph{e.g}\onedot} 
\def\ie{\emph{i.e}\onedot}

\makeatother

\begin{document}

\twocolumn[
\icmltitle{Rethinking Pre-Training in Tabular Data: \\ A Neighborhood Embedding Perspective}

\icmlsetsymbol{equal}{*}

\begin{icmlauthorlist}
\icmlauthor{Han-Jia Ye}{nju}
\icmlauthor{Qile Zhou}{nju}
\icmlauthor{Huai-Hong Yin}{nju}
\icmlauthor{De-Chuan Zhan}{nju}
\icmlauthor{Wei-Lun Chao}{osu}
\end{icmlauthorlist}

\icmlaffiliation{nju}{School of Artificial Intelligence, Nanjing University, China National Key Laboratory for Novel Software Technology}
\icmlaffiliation{osu}{The Ohio State University}

\icmlcorrespondingauthor{Han-Jia Ye}{yehj@lamda.nju.edu.cn}

\icmlkeywords{Machine Learning, ICML}

\vskip 0.3in
]
\printAffiliationsAndNotice{} 

\begin{abstract}
Pre-training is prevalent in deep learning for vision and text data, leveraging knowledge from other datasets to enhance downstream tasks. However, for tabular data, the inherent heterogeneity in attribute and label spaces across datasets complicates the learning of shareable knowledge. We propose \textbf{Tab}ular data \textbf{P}re-\textbf{T}raining via \textbf{M}eta-representation (\ours), aiming to pre-train a general tabular model over diverse datasets. The core idea is to embed data instances into a shared feature space, where each instance is represented by its distance to a fixed number of nearest neighbors and their labels. This ``meta-representation'' transforms heterogeneous tasks into homogeneous local prediction problems, enabling the model to infer labels (or scores for each label) based on neighborhood information. As a result, the pre-trained \ours can be {\em applied directly to new datasets, regardless of their diverse attributes and labels, without further fine-tuning.} Extensive experiments on 101 datasets confirm TabPTM's effectiveness in both classification and regression tasks, with and without fine-tuning.
\end{abstract}

\section{Introduction}

Pre-training has driven recent advances in AI \citep{devlin2018bert,liu2019roberta,Kirillov2023SAM,Dosovitskiy2021ViT}, with foundation models in vision and language benefiting from pre-training on large, diverse datasets of images and documents \citep{Zhou2023Foundation,oquab2023dinov2,Radford2021Learning}. Once trained, these models exhibit impressive generalizability to new tasks, often without the need for fine-tuning. However, such success has yet to be realized for tabular data, despite its ubiquity in many real-world applications, such as financial prediction~\citep{Cao2001Financial}, recommendation systems~\citep{richardson2007predicting}, and healthcare~\citep{Ogunleye2020XGBoost}.

A key challenge in tabular data is its \emph{inherent heterogeneity} across different sources and tasks. Typically represented as a matrix, with rows as instances and columns as attributes (features)~\citep{Borisov2022Deep}, tabular datasets can differ significantly in both dimensionality (\ie, the number of columns) and the semantic meaning of each dimension, even within the same application.  For example, different healthcare datasets may capture varying granularities and aspects of patient information. Even within the same feature entry (\eg, the $d$-th column), the meaning can differ (\eg, ``age'' vs.~``height''). This is in stark contrast to vision and text data (within the same language), where different data sources typically share the same ``vocabulary'' (\eg,  pixel, patch, or sub-word) and similar relationships between vocabulary ``elements'' (\eg, neighboring pixels often share colors). The lack of shared vocabulary and relationships in tabular data makes it difficult to jointly train a model across multiple datasets, let alone apply the pre-trained model directly to new downstream tasks.

\begin{displayquote}
\vspace{-2.5mm}
\textit{We hypothesize that to pre-train a useful model for tabular data, one must first answer two key questions: What is the shared vocabulary across datasets, and what transferable knowledge can be learned from them?}
\vspace{-2.5mm}
\end{displayquote}

Some recent work addresses this by leveraging the semantic meanings of columns (\ie, attributes). By converting data instances into textual form (\eg, ``Age is 20, height is 170, ...''), tabular data can be tackled using language models \citep{Hegselmann2022TabLLM,Wang2022TransTab,Liu2022PTab,Wang2023AnyPredict}. When semantic meanings are inaccessible or ill-defined, as often happens in domains like healthcare or measurement data, some methods learn dataset-specific token representations for each attribute, transforming different datasets into a shared feature space \citep{Iwata2020Meta,Kumagai2022Few,Liu2022Distribution,Wydmanski2023HyperTab,zhu2023xtab}. However, for new downstream tasks, these approaches require re-training task-specific token representations before applying the pre-trained model.

\begin{figure*}
    \centering
    \includegraphics[width=\textwidth]{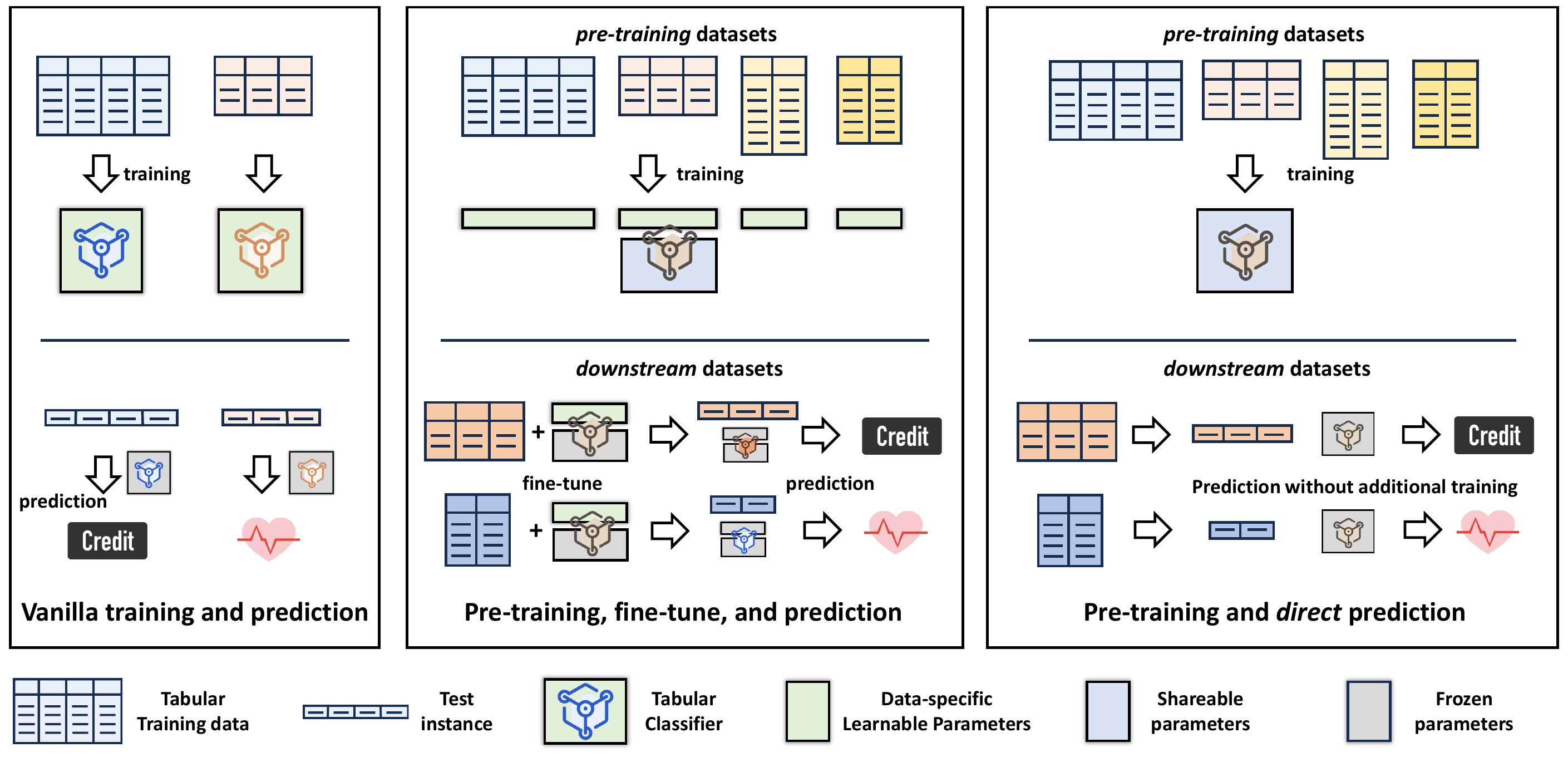}
    \caption{ \textbf{An illustration of \ours vs.~other training strategies on tabular data.} Left: the vanilla training and prediction pipeline, where tabular models are trained on each dataset separately. Middle: pre-training of a joint model on top of the learned dataset-specific token representations (of each attribute), in which the representation must be re-trained for downstream datasets. Right: \ours unifies heterogeneous tabular datasets via the meta-representation using neighborhood information, allowing pre-training a shareable model to predict labels based on such information, even without fine-tuning it on downstream datasets.}
    \label{fig:teaser}
\end{figure*}

Recognizing the limitations of current approaches, we explore a novel direction for pre-training a general model for tabular data. We draw inspiration from traditional methods in \emph{dimensionality reduction} and \emph{manifold learning}, as well as recent work in \emph{comparing neural network representations}, to address the question of \textbf{shared vocabulary across heterogeneous tabular datasets}. In multidimensional scaling, different tabular datasets can be embedded into a shared, low-dimensional matrix if they produce the same pairwise distance matrix between instances \citep{davison2000multidimensional,carroll1998multidimensional,torgerson1952multidimensional}. Similarly, manifold learning enables a high-dimensional dataset to be re-represented by a lower-dimensional one if they preserve the same neighborhood relationships \citep{weinberger2004learning,yan2006graph,van2008visualizing,hinton2002stochastic}. 
Recent work also suggests that two pre-trained neural networks, even with different architectures, encode similar knowledge if their affinity matrices across data instances are similar~\citep{kornblith2019similarity,huh2024platonic}. In short, \textbf{the relationships among instances} within a dataset can serve as the shared vocabulary across diverse tabular datasets, regardless of their dimensionality or semantic meanings.

\textbf{What transferable knowledge} can we learn from the shared vocabulary, \ie, the relationships among instances? We draw inspiration from the well-established \emph{nearest-neighbor-based algorithms} \citep{boiman2008defense,wang2019simpleshot,sanchez1997use,chaudhuri1996new,zhang2006svm,chao2013facial}. The key insight is that an instance's relationship to a particular label can be inferred from its relationships to the nearest neighbors of that label.

Putting it all together, we propose \textbf{Tab}ular data \textbf{P}re-\textbf{T}raining via \textbf{M}eta-representation (\ours), a novel pre-training approach for general tabular models using heterogeneous datasets. \ours represents each data instance (\ie, a row) {\em by the local context of its nearest neighbors in the training set.} Specifically, each instance is ``meta-represented'' as a fixed-dimensional vector encoding its distances to a set of nearest neighbors (ordered by distance) and their labels, while other methods may apply. This ``meta-representation'' standardizes diverse datasets, \emph{unifying} instances into a consistent form with uniform dimensionality and semantics, and transforming heterogeneous tasks into \emph{homogeneous local prediction tasks}. \ours then trains a joint neural network on the meta-representations across multiple datasets, learning to map the neighborhood information of each instance to the ground-truth label. This ability to \emph{predict the label of a given instance in its local context} can be transferred to new downstream tasks. Namely, the pre-trained model can be directly applied to new tabular datasets (represented by the meta-representation) without fine-tuning, though fine-tuning may enhance performance. (See \autoref{fig:teaser} for a comparison with existing tabular model strategies.)

We extensively validate \ours on 101 datasets across various scenarios. The results demonstrate \ours's effectiveness in acquiring shareable knowledge from diverse datasets, enabling strong generalizability on new classification and regression tasks. Our contributions are two-fold:
\begin{itemize}[topsep=-2pt,itemsep=-2pt,leftmargin=*] 
\item We propose meta-representations to reduce attribute heterogeneity and facilitate pre-training of a general model across diverse tabular datasets. \item Using the meta-representations, the pre-trained \ours can directly generalize to new datasets without additional training while achieving state-of-the-art accuracy in many datasets after fine-tuning, as shown by extensive experiments in both classification and regression tasks. 
\end{itemize}

\noindent{\bf Remark.} Our approach, \ours, is conceptually related to TabPFN \citep{Hollmann2022TabPFN}, which learns a general in-context model for tabular classification tasks. 
Given a downstream dataset, TabPFN inputs the (whole) training set and each test instance into the pre-trained model, using the learned interactions between training and test data to make predictions. Our \ours is particularly advantageous in three aspects. 
First, unlike TabPFN, which generates in-context tasks in every pre-training step and feeds all the corresponding instances into the model at once, \ours pre-meta-represents each instance, enabling smaller batch sizes and more efficient pre-training. Second, a similar property applies to testing: \ours can pre-meta-represent each test instance, allowing for more efficient testing by inputting only the meta-representations into the model. Third, putting these two points together, \citet{Hollmann2022TabPFN} claimed that TabPFN is limited to small datasets with fewer than a thousand instances, while \ours can handle and be applied to much larger tabular datasets.

\section{Related Work (More in~\autoref{sec:related_appendix})}

Tabular data is ubiquitous in many fields \citep{richardson2007predicting,vanschoren2014openml,superconductivty}.
Recent efforts have extended deep learning to the tabular domain \citep{Cheng2016Wide,WangFFW17DCN,GorishniyRKB21Revisiting,Huang2020TabTransformer,ZivA22Tabular,Grinsztajn2022Why}, and explore the feasibility of general discriminative models that can adapt to a broad array of downstream datasets.
The potential for in-distribution generalization of pre-trained tabular models has been demonstrated in contexts such as multi-task learning \citep{Argyriou2006Multitask,Zhang2022Multitask,Rubachev2022revisiting,Luetto2023One} and self-supervised learning~\citep{UcarHE21SubTab,BahriJTM22Scarf}, where data from various sources are in a consistent format. 
Some recent approaches utilize the deep neural network to pre-train a more generalizable tabular model, taking the difference in attributes and labels into account. The main difficulty lies in the ambiguity of shareable vocabulary and transferable knowledge across datasets.
One representative kind of approaches addresses this by harnessing the semantic meanings of column names (\ie, attribute names) to transform instances into text, thereby leveraging large language models to enhance prediction capabilities across diverse datasets \citep{Liu2022PTab,Hegselmann2022TabLLM,Zhang2023Generative,Wang2023AnyPredict}. 
Concurrently, other researchers have focused on learning shared components, such as attribute-agnostic transformations, to provide effective initial weights for adapting models to new tasks~\citep{Iwata2020Meta,Liu2022Distribution,Zhang2023Meta,shen2023cross,zhu2023xtab}. 
\citet{Hollmann2022TabPFN} addressed dataset heterogeneity by padding attributes across datasets into the same size and utilizing the contextual learning capabilities of transformers for classification tasks.
Our proposed \ours standardizes heterogeneous datasets into a uniform format using meta-representations. This enables the effective pre-training of a general model that can be directly applied to or minimally fine-tuned for downstream datasets without necessitating additional parameters.

\section{Preliminary}
\label{sec:prelim}
\noindent{\bf Learning from a single tabular dataset}.
We denote a tabular classification dataset as $\gD=\{(\vx_{i}, y_i)\}_{i=1}^N$, which has $N$ examples (rows in the table) and $C$ classes. The label $y_i\in\{1,\ldots,C\}$ can be represented by a one-hot form as $\vy_i\in\{0,1\}^C$.
For regression tasks, the label becomes a real number, \ie, $y_i\in\R$.

Each instance $\vx_i\in\gX$ comprises $d$ attributes (columns) and is represented by a vector\footnote{We assume all attributes of an instance are numerical (continuous). If there exist categorical (discrete) attributes, we transform them into the one-hot forms in advance.}. 
The goal is to estimate the posterior probability given an instance, \ie, $\Pr(y_i \mid \vx_i, \gD)$, often implemented by a model $f$ mapping from $\gX$ to the label space $\gY$.
The generalization ability of $f$ is measured by the prediction accuracy on an unseen instance sampled from the same distribution as $\gD$.

\noindent{\bf Pre-training across multiple tabular datasets}.
Tabular datasets could be collected from various sources. Assume there are $T$ datasets $\sD=\{\gD_1, \ldots, \gD_T\}$, in which the attribute and label spaces of the $t$-th dataset are denoted by $\gX_t$ and $\gY_t$. 
Due to the inherent heterogeneity across datasets, we denote the number of instances, attributes (\ie, the dimension of an instance), and classes in dataset $\gD_t$ as $N_t$, $d_t$, and $C_t$, respectively. Unlike multi-task or multi-view learning, which assumes one of $\gX_t$ and $\gY_t$ is homogeneous (\ie, identical) across all datasets, we consider {\em heterogeneous} datasets where the meanings of attributes and classes vary from one dataset to another.
By pre-training a model $f$ over $T$ heterogeneous datasets, we expect $f$ to deal with different attribute and label spaces and generalize its learned ability to an unseen downstream task $\gD_u$.
There are two primary challenges to pre-train $f$. The first is to determine the shared ``vocabulary'' across datasets so that $f$ can learn from $\sD$ and be applied to any heterogeneous dataset.
The second is to identify and encode the ``transferable knowledge'' from $\sD$ into $f$, enhancing the generalizability of $f$ such that it can improve an unseen downstream task.

There are two types of generalization. The first is the {\em direct} generalization, which means the learned $f$ can directly predict the label of an unseen instance $\vx_*^u$ in task $\gD_u$ without additional training, \ie, $\hat{y}_*^u = f(\vx_*^u\; |\; \gD_u)$, 
where $\hat{y}_*^u$ could be inferred via the shareable knowledge learned on the pre-training datasets.
In addition, the learned $f$ can act as the initialization to be updated by several steps of gradient descent, minimizing the empirical risk over the target dataset $\gD_u$. 
This {\em fine-tuning} strategy learns the task-specific property by fitting to $\gD_u$. 
Due to the heterogeneity of datasets, additional parameters, such as the feature tokenizer, are often introduced to adapt the model to downstream tasks~\citep{zhu2023xtab}, which requires special tuning methods. 

\emph{Our goal is to develop a versatile general model $f$ that can be either directly applied to a target task or fine-tuned without introducing new parameters.}

\section{Proposed Approach}
\begin{figure*}
    \centering
\includegraphics[width=0.9\textwidth]{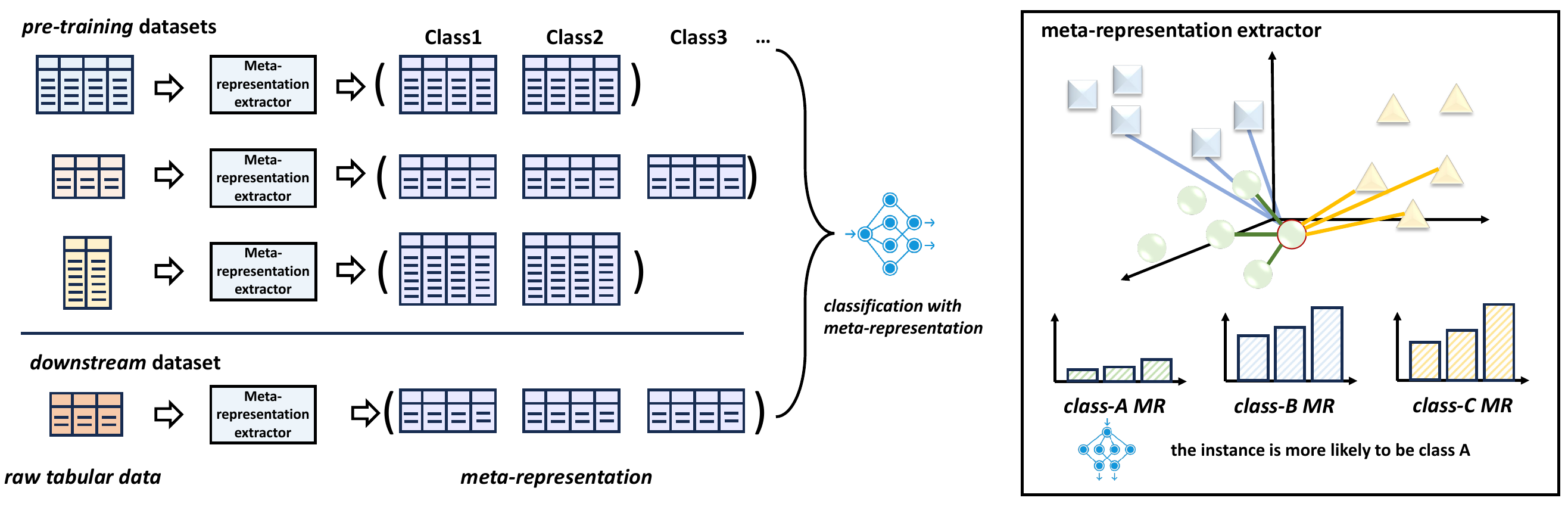}
    \caption{\textbf{An illustration of the Meta-Representation (MR) for classification.} MR transforms heterogeneous tabular datasets with different dimensions into a homogeneous form. 
    A dataset has a set of $K$-dimension MRs, one for each class, which can be used to derive the prediction scores for different classes. We pre-train a joint model on the MRs of multiple datasets and extend its generalization ability to downstream datasets.
    The right figure shows the MR of an instance (a row in the original tabular format), encoding distances from the instance to its nearest neighbors of each class, characterizing the class membership patterns.} 
    \label{fig:meta_representation}
\end{figure*}

Given the heterogeneity in attribute and label spaces across tabular datasets, the core idea behind the proposed \textbf{Tab}ular data \textbf{P}re-\textbf{T}raining via \textbf{M}eta-representation (\ours) is to \emph{pre-unify} them, enabling the training and application of a joint deep neural network.
We draw inspiration from neighborhood-based methods and introduce the concept of meta-representation, which serves as the foundation for {\ours}'s pre-training strategy. 
\ours applies to both classification and regression tasks, as illustrated in~\autoref{fig:meta_representation}.

\subsection{Motivation from Neighborhood Embedding}\label{sec:motivation}
Accurate estimation of the posterior $\Pr(y \mid \vx, \gD)$ is crucial for all classification and regression tasks~\citep{murphy2022probabilistic}. 
While most recent works are parametric methods (\eg, learning a neural network classifier), many traditional methods were non-parametric, such as kernel density estimation.

To elucidate, let us use the non-parametric balloon kernel density estimator with a uniform class prior \citep{terrell1992variable}, whose posterior density is defined as
\begin{equation}
    \Pr(y=c \mid \vx, \gD) = \frac{1}{K} \sum_{j\in \gN_K(\vx, \gD)} \mathbb{I}[y_j = c]\;,\label{eq:kde_estimator}
\end{equation}
where $\mathbb{I}[\cdot]$ is the indicator function and  $\gN(\vx; \gD)$ is the $K$ nearest neighbors of $\vx$ based on distance $\operatorname{dist}(\cdot, \cdot)$. Namely, \emph{the prediction depends on the neighborhood of $\vx$ in $\gD$.} 

Instead of treating all instances in $\gN(\vx; \gD)$ equally, one popular variant is to allocate larger weights to closer neighbors~\citep{bishop2006pattern,RasmussenW06}:
\begin{align}
&\Pr(y \mid \vx, \gD) = \label{eq:kde_vector}\\
&\left(\operatorname{softmax}\left(\left[-\operatorname{dist}(\vx, \vx_1);\ldots;-\operatorname{dist}(\vx, \vx_{K})\right]\right)\right)^\top Y_K\;, \nonumber
\end{align}
where $Y_K\in\{0,1\}^{K\times C}$ is the label matrix of $\mathcal{N}_K(\vx;\mathcal{D})$, with each row corresponding to the one-hot label $\vy_j^\top$ of neighbor $\vx_j$. This formulation also extends to regression, as detailed in~\autoref{sec:kNN_view}.

\noindent\textbf{Remark.}
\autoref{eq:kde_vector} indicates that an instance's relationship to a label $c$ can be inferred from its relationship to the nearest neighbors of that label, \emph{regardless of the dimensionality and attribute meanings.} This provides a {\em unifying} way to tackle heterogeneous datasets, in which the relationships among nearest instances serve as a shared vocabulary across datasets. Our meta-representation is inspired by this idea.

\subsection{Meta-Representation of an Instance}\label{sec:meta_Rep}

\noindent{\bf Abstract meta-representation.} The proposed meta-presentation is to encode the neighborhood information for each instance. We first introduce the abstract form for a {\em classification} dataset $\gD$ with $C$ classes. Let us denote by $\gD_{c} = \{(\vx_j, y_j)\;|\;y_j=c\}\;, \forall c=1,.\ldots, C$ the subset of $\gD$ corresponding to class $c$. For an instance $\vx_i\in\gD$, we calculate its distance to all the $|\gD_{c}|$ instances in $\gD_{c}$ and sort them in an {\em ascending} order: 
\begin{align}
    &\{\operatorname{dist}(\vx_i, \vx_1),\ldots,\operatorname{dist}(\vx_i, \vx_j),\ldots,\operatorname{dist}(\vx_i, \vx_{|\gD_{c}|})\}\notag\\
    {\rm s.t.}\;
    &\operatorname{dist}(\vx_i, \vx_1)\le\ldots\le\operatorname{dist}(\vx_i, \vx_{|\gD_{c}|})\;.\label{eq:calculate_distance}
\end{align}

We then select the $K$ nearest neighbors to construct the class-$c$ local context for $\vx_i$, and define a mapping from $\vx_i$ to its class-$c$ meta-representation by:
\begin{equation}
    \phi_c(\vx_i) = [\operatorname{dist}(\vx_i, \vx_1), \tilde{y}_1,\ldots \operatorname{dist}(\vx_i, \vx_{K}), \tilde{y}_K]\;\in\R^{2K},\label{eq:meta_representation}
\end{equation}
where $\tilde{y}_j$ is set as $1$ in the main paper, and we investigate other ways to provide auxiliary information in~\autoref{appendix_sec:ablation}. 
\emph{$\phi_c(\vx_i)$ captures the class-$c$ neighborhood information and indicates the likelihood of $\vx_i$ belonging to class $c$.
If $\vx_i$ resides in a high-density region of class $c$, the values in $\phi_c(\vx_i)$ would typically be small.} We use $\Phi(\cdot)=\{\phi_c(\cdot)\}_{c=1}^C$ to denote the meta-representation integrating all classes.

With the meta-representation, heterogeneous classification tasks become homogeneous prediction tasks based on local contexts. Regardless of the original dimensionality, $\phi_c(\cdot)$ outputs a fixed-dimensional representation based on $K$, facilitating pre-training over diverse datasets.

In {\em regression} tasks, labels in $\gD$ are scalars. We thus retrieve $K$ nearest neighbors from the whole dataset and slightly modify \autoref{eq:meta_representation} to obtain the meta-representation:
\begin{equation}
    \Phi(\vx_i) = \big[\operatorname{dist}(\vx_i, \vx_1), y_1,\ldots,\operatorname{dist}(\vx_i, \vx_{K}), y_K\big]\;.\label{eq:meta_representation_reg}
\end{equation}

\noindent{\bf Metric-based meta-representation.} 
The distance measure $\operatorname{dist}(\cdot,\cdot)$ in~\autoref{eq:kde_vector} and~\autoref{eq:calculate_distance} plays a crucial role in neighborhood-based methods~\citep{ScholkopfHS01, ScholkopfS02}. While the Euclidean distance is widely used in learning from a single dataset, learning from multiple heterogeneous datasets necessitates a more sophisticated, adaptive metric design. 

To this end, we adopt the weighted Minkowski distance (see~\autoref{appendix_sec:exp_setup} for other metrics):
\begin{equation}
    \operatorname{dist}(\vx_i, \vx_j) = \left(\sum_{l=1}^d w_l \cdot |\vx_{il} - \vx_{jl}|^p\right)^{\frac{1}{p}}\;,\label{eq:weighted_dist}
\end{equation}
where $\vx_{il}$ denote the $l$-th dimension of $\vx_{i}$.
We set $p\in\{1,2\}$ and $w_l > 0$, which weighs each dimension. When $w_l = 1$, \autoref{eq:weighted_dist} degenerates to Euclidean distance ($p=2$) or Manhattan distance ($p=1$). As a dataset may contain noisy or irrelevant attributes, we calculate adaptive weights based on mutual information. Given a training set $\gD=\{\mX, \mY\}$, where $\mX$ and $\mY$ denote the instance matrix and label vector, respectively, we derive the attribute weights by
\begin{equation}
    w_l = \operatorname{normalize}\left(\operatorname{MI}(\mX_{:,l},\; \mY)\right)\;, \label{eq:MIMI}
\end{equation}
where $\mX_{:,l}$ is the $l$-th column of $\mX$, \ie, the vector containing the $l$-th attribute values of all instances. $\operatorname{MI}(\cdot, \cdot)$ computes the mutual information between two sets, measuring the dependency between an attribute and the labels~\citep{Brown2012Conditional}. 
The larger the mutual information is, the more important an attribute is. The $\operatorname{normalize}(\cdot)$ operator divides input values by their cumulative sum for normalization. 

We note that there are many ways to set $\operatorname{dist}(\cdot,\cdot)$. One may even consider learning it by metric learning~\cite{kulis2013metric}. Empirically, we find 
that integrating \autoref{eq:weighted_dist} and \autoref{eq:MIMI} is sufficient to enhance the model's generalizability significantly and robustly.

\noindent{\bf Meta-representation in the few-shot scenario.}
The aforementioned meta-representation assumes there are at least $K$ neighbors in the set $\gD_{c}$. However, in few-shot classification tasks or scenarios with minority classes, $\gD_{c}$ may contain fewer than $K$ instances.
To address this issue, we pad the meta-representation with its last value (the largest distance), inspired by \citep{Yang2012Multilabel}. 

\begin{figure}[t] 
    \centering
    \begin{minipage}{0.315\linewidth}
    \includegraphics[width=\textwidth]{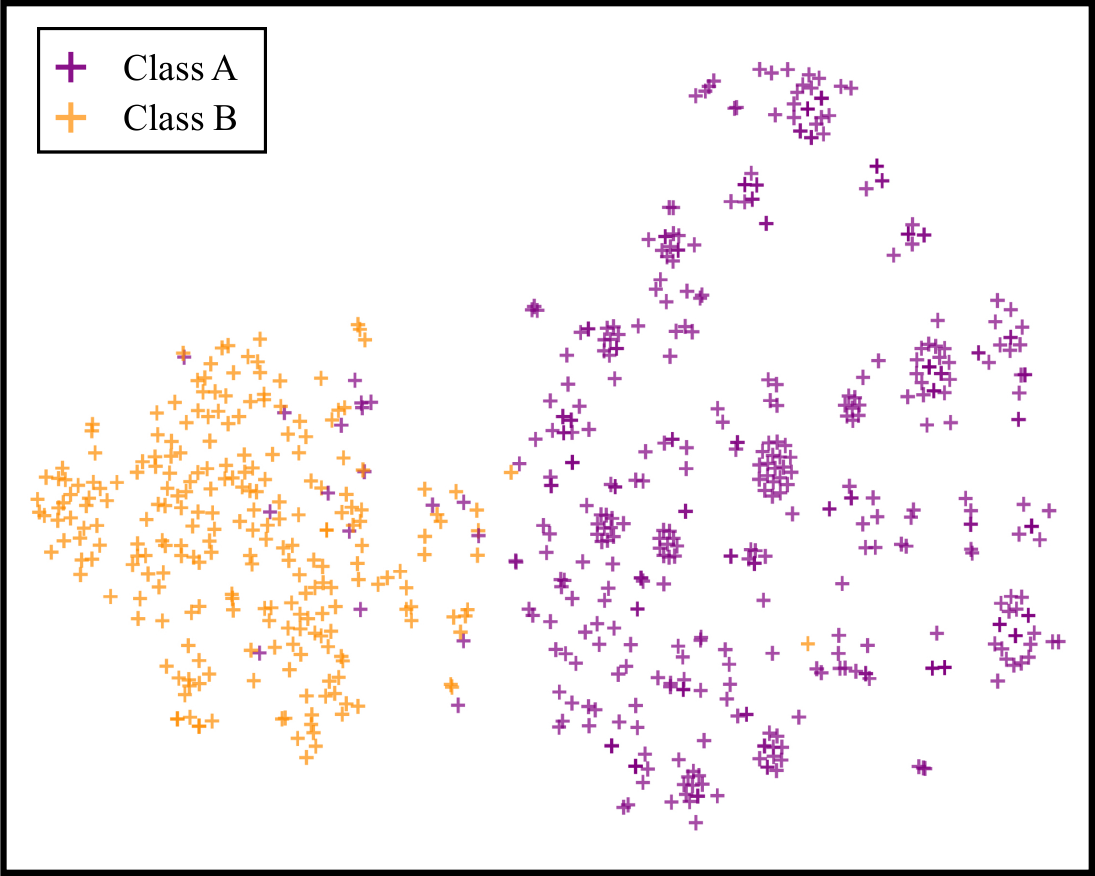}
    \centering
    {\small \mbox{(a) {breast-cancer-wisc}}}
    \end{minipage}
    \begin{minipage}{0.315\linewidth}
    \includegraphics[width=\textwidth]{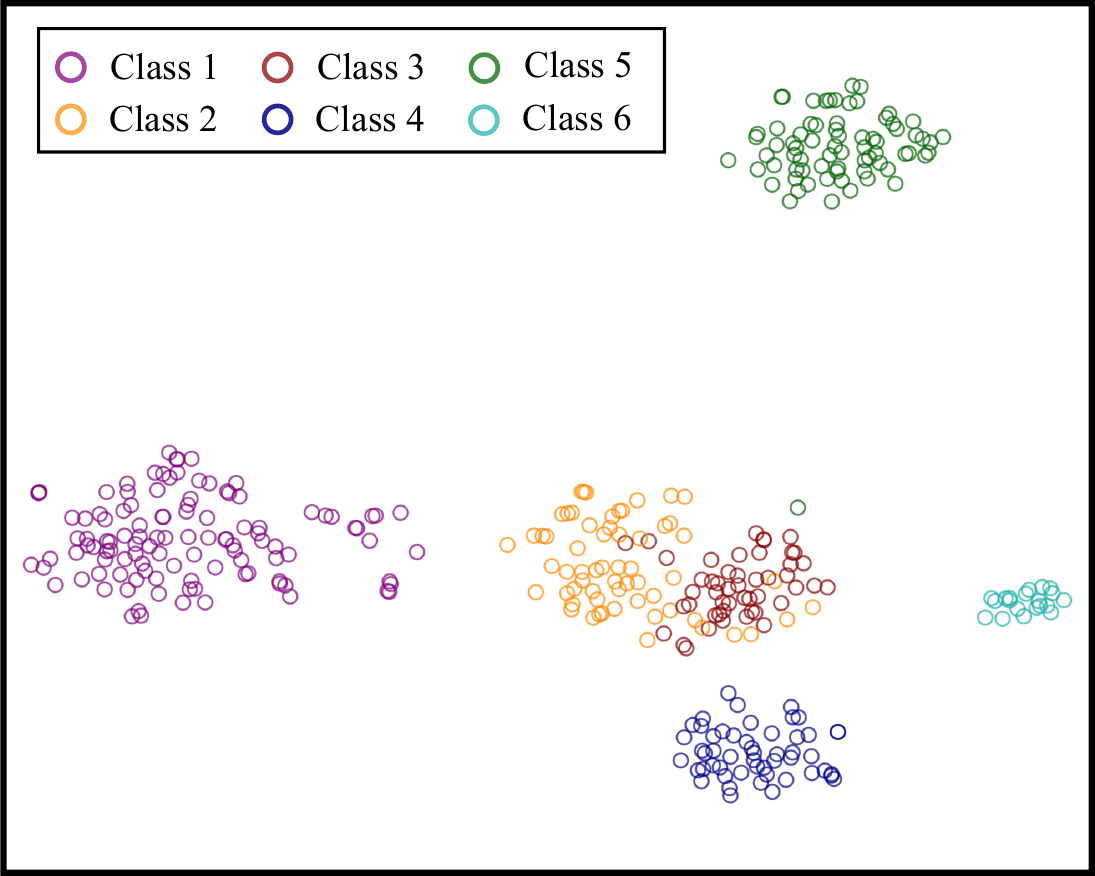}
    \centering
    {\small \mbox{(b) {dermatology}}}
    \end{minipage}
    \begin{minipage}{0.32\linewidth}
    \includegraphics[width=\textwidth]{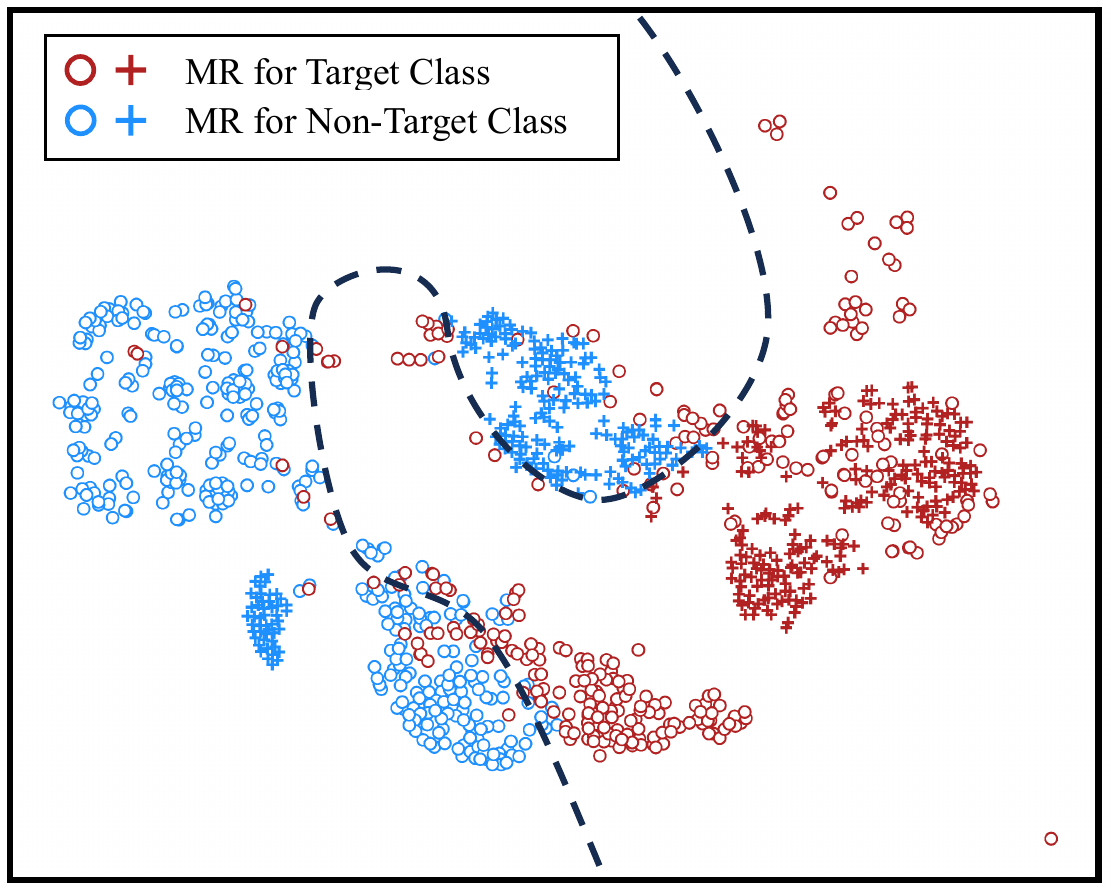}
    \centering
    {\small \mbox{(c) {Joint Space w/ MR.}}}
    \end{minipage}
    \caption{\textbf{Pilot study of Meta-Representation (MR)} over two datasets ``breast-cancer-wisc'' (binary, denoted by ``+'') and ``dermatology'' (multi-class, denoted by ``$\circ$''), using t-SNE~\citep{van2008visualizing}. We use colors to distinguish classes. (a) and (b) display the characteristics of each dataset. In (c), we homogenize the datasets using MR, with red indicating the MR for the target class ($\phi_{y_i}(\vx_i)$) and blue for non-target classes ($\phi_{c \neq y_i}(\vx_i)$). MR effectively unifies datasets into a joint representation space where their classifications can be implemented by the dotted line. 
    }
    \vspace{-4mm}
    \label{fig:pilot_study}
\end{figure}

\subsection{A Pilot Study on Meta-Representation}
We use classification to illustrate how meta-representation unifies two heterogeneous datasets. We consider two datasets, ``breast-cancer-wisc'' (binary) and ``dermatology'' (multi-class). We set $K=8$ and calculate the meta-representation $\Phi(\vx_i)$ for all classes in each dataset. We then \emph{partition the meta-representation into two sets}: the target class ($\phi_{y_i}(\vx_i)$) and the non-target classes ($\phi_{c\neq y_i}(\vx_i)$). 

\autoref{fig:pilot_study} visualizes the results. We use different shapes to differentiate datasets and red/blue to denote the two sets of meta-representations, respectively. We have several observations. First, meta-representations bring different datasets into a shared feature space. Specifically, those for the target and non-target classes are separated (by the dotted line). This indicates the potential to learn a shared, general tabular model on top of meta-representations. Second, meta-representations \emph{do not unify the distributions} of different datasets. For example, the target-class meta-representations denoted by red ``+'' are clustered while those denoted by red ``$\circ$'' are not. Thus, joint training on multiple datasets is necessary for building the general model.

\subsection{Making Predictions via Meta-Representation}
\noindent{\bf Classification}. Given a dataset $\gD$, we represent an instance $\vx_i$ by the meta-representation $\Phi(\vx_i)=\{\phi_c(\vx_i)\}_{c=1}^C$. We then compute the prediction score for each class by 
\begin{equation}
[s(\vx_i)_1,\ldots,s(\vx_i)_C] = {\bf T}_{\Theta}\left([\phi_1(\vx_i), \ldots, \phi_C(\vx_i)]\right),\label{eq:transformation}
\end{equation}
where ${\bf T}_{\Theta}$ is a model that captures the local class membership patterns from the meta-representation and outputs the corresponding scores; ${\Theta}$ denotes the learnable parameters in ${\bf T}$. Based on the scores, the predicted class is 
\begin{equation}
    \hat{y}_i = \argmax_c\; \left\{s(\vx_i)_1,\ldots,s(\vx_i)_C\right\}\;.\label{eq:prediction}
\end{equation}
In this paper, we implement ${\bf T}$ using Multi-Layer Perceptrons (MLPs) inspired by~\cite{jabri2016revisiting},
\begin{equation}
    s(\vx_i)_c = {\bf MLP}(\phi_c(\vx_i)),\;\forall c=1,\ldots,C\;. \label{eq:mlp_mapping}
\end{equation}
The parameters of MLP are shared across all classes, and we describe the detailed architecture in~\autoref{appendix_sec_method}.
When multiple types of distances are used, we concatenate them together at first and then use \autoref{eq:mlp_mapping} to map the concatenated meta-representation vectors to a scalar, reminiscent of multiple kernel learning~\cite{bach2004multiple}. 

The classification strategy based on meta-representation applies to heterogeneous tasks with different attributes and class spaces.
As a result, it enables the use of a joint model ${\bf T}_{\Theta}$ over heterogeneous tasks.

\noindent{\bf Regression}. Meta-representation also applies to tabular regression tasks (see \autoref{eq:meta_representation_reg}). Since the label is a continuous value, we map the meta-representation to a scalar
\begin{equation}
    s(\vx_i) = {\bf T}_{\Theta}\left(\Phi(\vx_i)\right)\;.\label{eq:transformation_reg}
\end{equation}
${\bf T}_{\Theta}$ captures the continuous value distribution in the neighborhood and accordingly predicts the label of $\vx_i$, ideally as a weighted combination of its neighbors' labels. 
We also implement ${\bf T}$ with MLP with learnable parameter ${\Theta}$.
\subsection{Pre-training with Meta-Representation}

We pre-train a joint model ${\bf T}_{\Theta}$ from $\sD$ (see~\autoref{sec:prelim}) by
\begin{equation}
\label{eq:objective}
\min_{\Theta}\;\sum_{t=1}^T\;\sum_{i=1}^{N_t}\ell\left({\bf T}_{\Theta}(\Phi(\vx_i^t)),\; y^t_i\right)\;.
\end{equation}
The loss $\ell(\cdot,\cdot)$ measures the discrepancy between the prediction and true label, and we use cross-entropy for classification and mean square error for regression.

Given a downstream dataset $\gD_u$, we first obtain the meta-representation for each instance. Then the learned ${\bf T}_{\Theta}$ can be applied directly without further training or acts as an initialization for fine-tuning without additional parameters. 
\autoref{appendix_sec_method} provides more details of \ours, including the pre-training and deployment workflows of \ours in Algorithm~\autoref{alg:pretrain} and Algorithm~\autoref{alg:downstream}, respectively. 

\noindent\textbf{Summary and remarks.} \ours treats the relationships among nearest instances as the shared vocabulary, unifying heterogeneous tabular tasks into homogeneous local prediction tasks in a common representation space. This facilitates learning transferable knowledge from multiple datasets and directly applying the learned model to downstream tasks without further training. 

We note that while representing each sample by its local neighborhood may lose some information, it has been proven effective in many machine learning methods (\eg, nearest-neighbor-based approaches), particularly when the data lies on a low-dimensional manifold. The pairwise distance computation can be largely accelerated with approximation techniques~\cite{li2019approximate,andoni2018approximate}.

\section{Experiments}
We conduct extensive experiments to validate \ours. Specifically, we focus on \textbf{few-shot} downstream tasks, where pre-training should thrive, and then extend the evaluation on \textbf{full-shot} tabular tasks. We analyze \ours as well as meta-representation in~\autoref{sec:ablation_main_paper} and~\autoref{appendix_sec:ablation}. 

\subsection{Setups}
\noindent{\bf Datasets}. 
We randomly selected 72 datasets from UCI and OpenML, the two largest repositories of tabular datasets, covering diverse domains such as healthcare and software engineering. These include 36 classification and 36 regression datasets, which were split into two partitions: one for pre-training and the other for downstream evaluation. The main paper presents results for one partition, while results from swapping the partitions are in~\autoref{appendix_sec:ablation}.

To further evaluate the few-shot capabilities of \ours, we incorporate TinyBenchmark from~\cite{ye2024closer} as additional downstream datasets, which do not overlap with the original 36 classification datasets. TinyBenchmark consists of 29 classification datasets that provide consistent average ranking results with evaluations over 300 datasets.

\begin{figure}[t]
    \begin{minipage}{0.95\linewidth}
    \includegraphics[width=\textwidth]{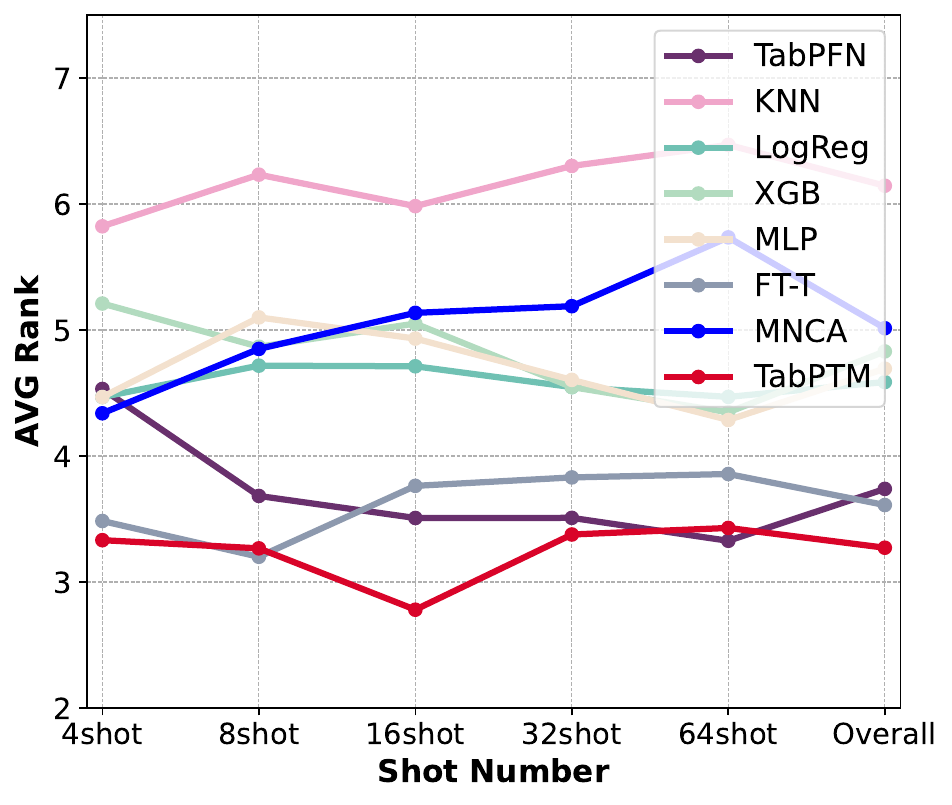}
    \centering
    \end{minipage}
        \vspace{-4mm}
    \caption{Performance of different methods on few-shot tasks across 65 datasets. For each downstream dataset, $\{4,8,16,32,64\}$ examples per class (shot) are randomly sampled as the training set. \ours outperforms other methods overall. The complete results for all few-shot tasks are presented in~\autoref{tab:fewshot_main_whole} and~\autoref{tab:fewshot_tinybench_whole}.
    ``Overall'' averages the ranks among all shots. 
    }
    \vspace{-7mm}
    \label{fig:few-shot_results}
\end{figure}
\noindent{\bf Evaluation criteria}.
After pre-training, performance is evaluated on downstream datasets using accuracy for classification tasks and RMSE for regression tasks. For each dataset, 64\%/16\%/20\% of the instances are randomly split for training/validation/test, respectively. Average results over 15 random seeds are calculated following~\citep{GorishniyRKB21Revisiting}.
To compare methods across datasets, we use average ranking as a measure, following the evaluation protocol in~\cite{DelgadoCBA14hundreds,McElfreshKVCRGW23}. In the few-shot setting, we randomly select \{4, 8, 16, 32, 64\} training instances per class and evaluate the model’s ability to learn with limited data. Each few-shot configuration is repeated 5 times, and the average results are reported.

\noindent{\bf Comparison methods}.
We compare \ours against three types of methods. First, {\em classical models} such as Support Vector Machines (SVM) and XGBoost~\citep{chen2016xgboost} are included as baselines. Second, we evaluate {\em deep tabular models}, including Multi-Layer Perceptron (MLP)\citep{Kadra2021Well}, FT-Transformer (FT-T)\citep{GorishniyRKB21Revisiting}, TabR~\cite{Gorishniy2023TabR}, and ModernNCA~\cite{ye2025modernnca}. Third, we compare against {\em pre-trained tabular models}, such as XTab~\citep{zhu2023xtab}, DEN~\citep{Liu2022Distribution}, and TabPFN~\citep{Hollmann2022TabPFN}.

\noindent{\bf Implementation details}.
For the baseline methods, we strictly follow the training protocol of~\cite{GorishniyRKB21Revisiting,Gorishniy2023TabR,ye2024closer}. In the full-shot setting, hyperparameter tuning is performed using Optuna~\cite{akiba2019optuna} over 100 trials, with early stopping based on validation performance. In the few-shot setting, default hyperparameters are used due to the limited training data available. More details are in~\autoref{appendix_sec:exp_setup}.


\ours is implemented using a three-layer MLP (${\bf T}_{\Theta}$). During pre-training, the learning rate is set to 0.001 and batchsize is 1024. \ours sets $K = 16$ for regression and $K = 128$ for classification. 
In few-shot learning, the pre-trained \ours is applied directly without additional fine-tuning. In full-shot settings, the model is fine-tuned by updating only the top-layer module while keeping pre-trained weights as initialization. No additional learnable parameters are introduced. Fine-tuning is performed with a learning rate of 0.01 for 30 epochs, and the model from the final epoch is used for evaluation. Unlike most tabular learning methods that require extensive hyperparameter tuning for each dataset-method pair, \ours employs default hyperparameters and a limited number of training epochs, making it computationally efficient.

\subsection{Generalization Ability of the Pre-Trained Model}\label{sec:ablation_main_paper}
\noindent{\bf Results on few-shot tasks}. 
The few-shot evaluation initially involves 36 datasets split into two parts: 18 datasets for pre-training and the remaining 18 as downstream tasks for evaluation. To ensure a comprehensive assessment, we swap the splits, pre-training on the second set of 18 datasets and evaluating on the first set.
Additionally, we incorporate 29 classification datasets from TinyBenchmark into the first pre-training phase. This inclusion allows us to assess how well \ours generalizes when trained on a broader and more diverse set of datasets. The final results are computed as the average rank over all 65 datasets across different shot numbers (ranging from 4 to 64).

As shown in~\autoref{fig:few-shot_results}, the rankings of various methods fluctuate across shot numbers. However, \ours consistently maintains strong competitiveness, achieving the best overall performance across all datasets. \ours not only outperforms tree-based models like XGBoost and deep tabular baselines like MLP but also surpasses the neighbor-based MNCA in few-shot scenarios, demonstrating the advantages of its neighbor embedding.
Compared to the pre-trained model TabPFN, \ours retains an advantage across all shot configurations, highlighting its robustness and effectiveness in few-shot learning. The inclusion of diverse datasets from TinyBenchmark in the pre-training phase further emphasizes \ours's ability to generalize across heterogeneous tabular data.

\begin{figure*}[t]
\centering
    \vspace{-2mm}
    \begin{minipage}{0.45\linewidth}
    \includegraphics[width=\textwidth]{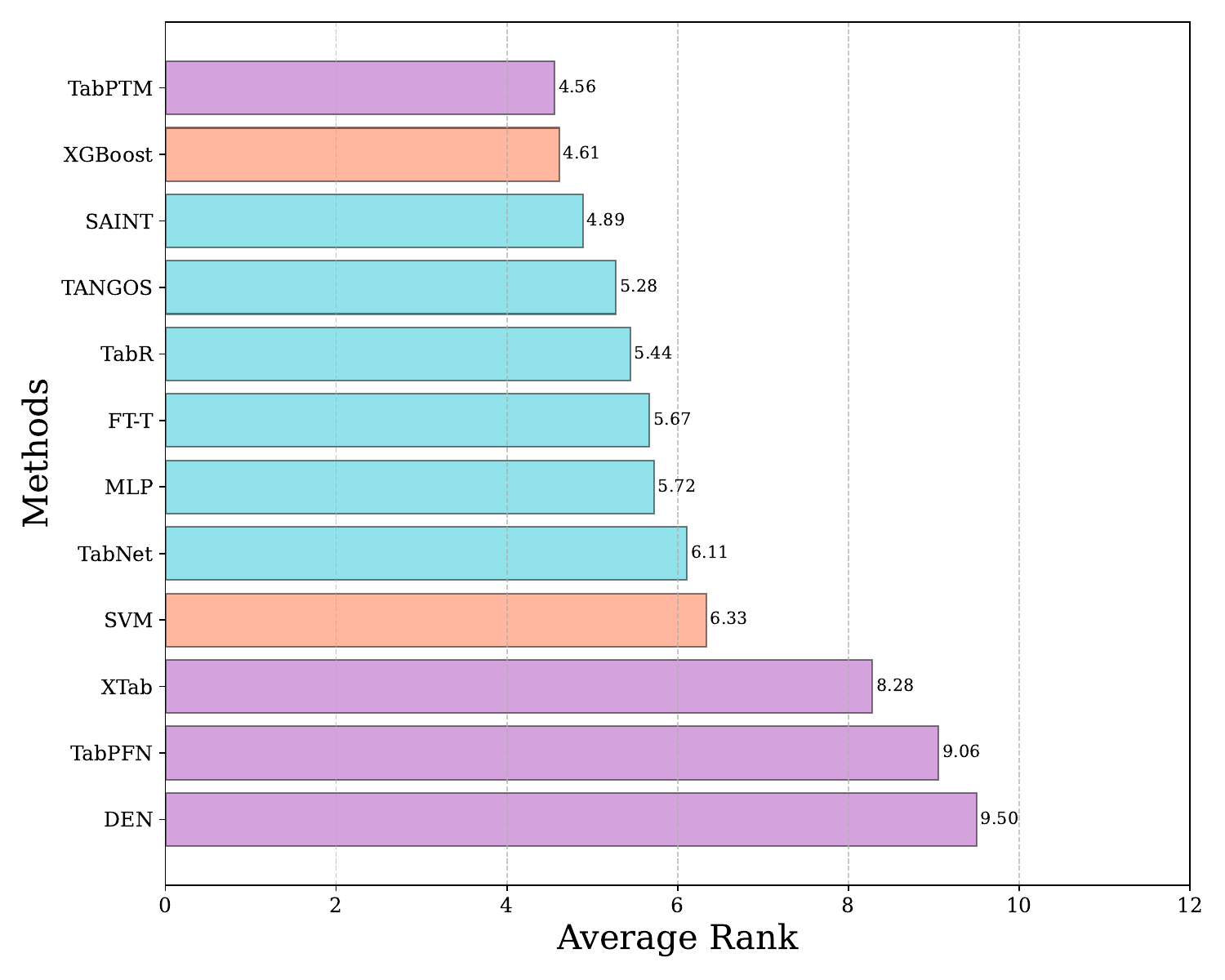}
    \centering
    {\small \mbox{(a) {Classification.}}}
    \end{minipage}
    \begin{minipage}{0.45\linewidth}
    \includegraphics[width=\textwidth]{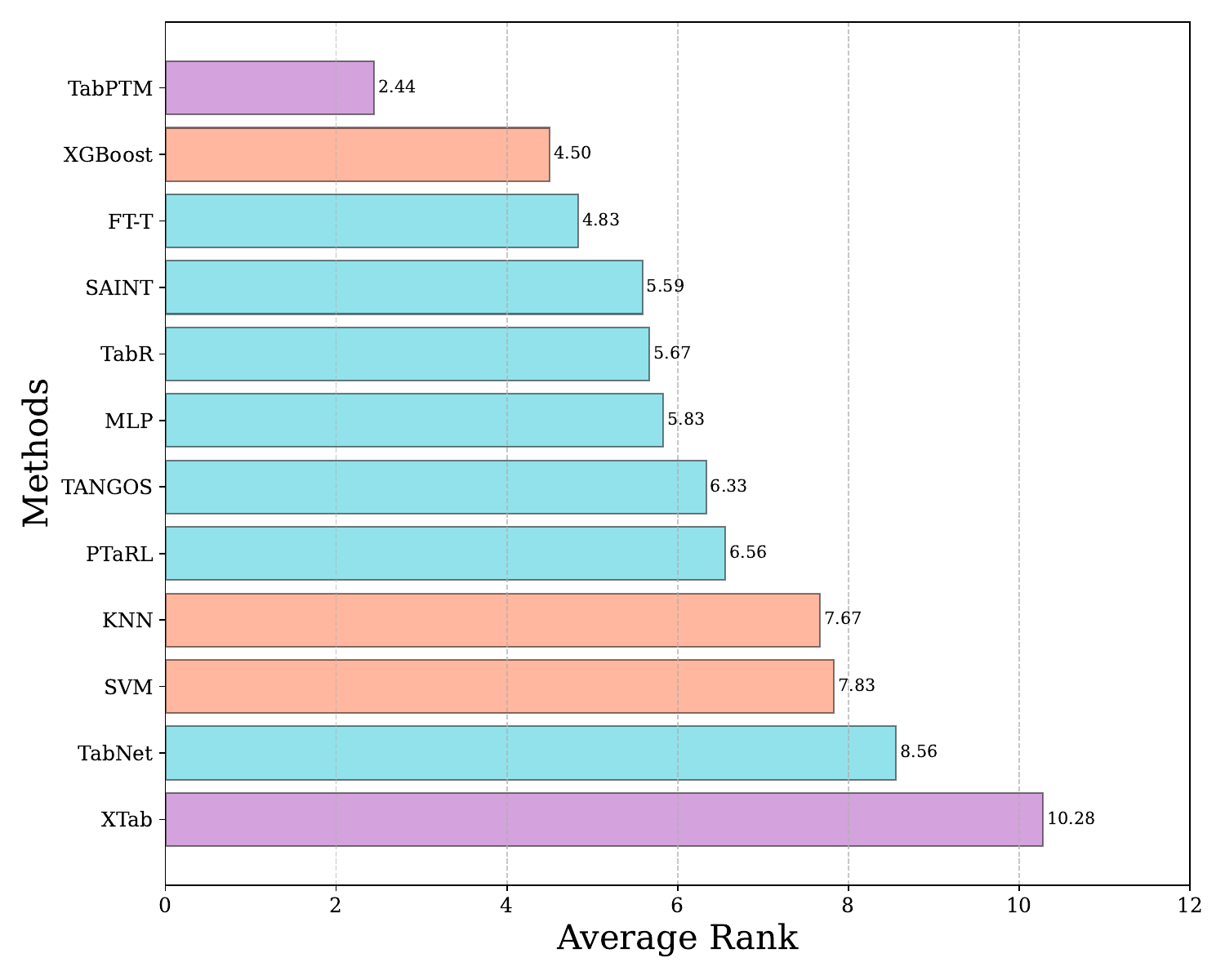}
    \centering
    {\small \mbox{(b) {Regression.}}}
    \end{minipage}
        \vspace{-3mm}
    \caption{Performance comparison across 18 classification and 18 regression datasets in the full-shot setting, where the entire training set is used. The pre-trained \ours, trained on 18 non-overlapping datasets, outperforms all other methods and achieves the highest average rank. The complete results are reported in~\autoref{tab:reg_main} and~\autoref{tab:cls_main}.}
    \vspace{-7mm}
    \label{fig:full-shot_results}
\end{figure*}

\noindent{\bf Results on full-shot tasks}. 
We further evaluate \ours in full-data downstream settings. Unlike other methods that require extensive hyperparameter tuning, we fine-tune \ours using a fixed hyperparameter configuration to assess whether pre-training simplifies the fine-tuning process.

As shown in~\autoref{fig:full-shot_results}, \ours achieves the best average rank across both classification and regression tasks, validating its effectiveness. The pre-trained \ours not only surpasses deep tabular models like TabR but also outperforms the boosting-based XGBoost.
While the MLP baseline is trained on raw features for each dataset separately, \ours, despite using the same MLP architecture, demonstrates superior generalization by leveraging shared knowledge from pre-training. Additionally, XTab pre-trains a joint transformer module and fine-tunes task-specific parameters for each downstream dataset. In contrast, \ours consistently outperforms XTab on most datasets without introducing additional learnable parameters.

\subsection{Ablation Studies}

\begin{table}

\setlength\tabcolsep{2pt}
    \caption{Time statistics on few-shot evaluations in~\autoref{fig:few-shot_results}. The runtime (in seconds) was measured on a system equipped with an Intel(R) Xeon(R) Gold 5218R CPU @ 2.10GHz and an NVIDIA RTX 6000 Ada Generation GPU.}
    \vspace{-0mm}
    \centering
    \tabcolsep 2pt
\begin{tabular}{lcccccc}
\addlinespace
\toprule
Time & TabPFN & XGB & MLP & FT-T & MNCA & \ours \\
\midrule
Min. & 1.276 & {\bf 0.109} & 1.502 & 1.781 & 1.828 & 1.024 \\
Max. & 3.688 & 7.657 & 7.551 & 56.25 & 6.751 & {\bf 3.056} \\
Avg. & 2.185 & 1.494 & 4.304 & 10.02 & 4.855 & {\bf 1.354} \\
\bottomrule
\end{tabular}
    \label{tab:time}
        \vspace{-7mm}
\end{table}

\noindent{\bf Computational cost of \ours}. 
For full-shot tasks, \ours reduces computational overhead by eliminating extensive hyperparameter tuning and requiring only a few fine-tuning iterations. In few-shot tasks, \ours further minimizes costs without additional training. The primary computational challenge lies in the initial nearest neighbor search.
To assess the total computational cost, we analyzed the time required for all few-shot tasks in~\autoref{fig:few-shot_results}, with results summarized in~\autoref{tab:time}. The reported runtime includes data preprocessing, model training, and evaluation. Compared to other deep learning models, \ours significantly reduces computation time while maintaining strong generalization. Although its runtime is comparable to XGBoost, \ours achieves superior performance in few-shot settings, demonstrating both efficiency and robustness.

\noindent{\bf The role of the shared top-layer model}.
To examine whether the MLP learned during pre-training effectively captures shareable knowledge and benefits downstream prediction, we conduct two controlled comparisons. First, we train XGBoost on the meta-representation of each dataset, denoted as XGBoost$_{\rm MR}$. Second, we train only the top-layer MLP in \ours without leveraging the pre-trained model, referred to as {\ours}$_{\rm S}$.

Results in~\autoref{tab:role} indicate that XGBoost$_{\rm MR}$ outperforms standard XGBoost, demonstrating that the proposed meta-representation enhances both classification and regression tasks. Furthermore, \ours surpasses {\ours}$_{\rm S}$, confirming that the pre-trained top-layer MLP indeed encodes transferable knowledge that benefits downstream tasks. The consistent superiority of \ours over XGBoost$_{\rm MR}$ further highlights its strong generalization ability.
Additional ablation studies in~\autoref{sec:pre-train-size} further support that the pre-trained model effectively captures shareable knowledge, significantly improving performance on diverse datasets.

\begin{table}

\setlength\tabcolsep{2pt}
    \caption{Comparison between XGBoost variants and \ours on three regression (STO, ADO, and CAC, evaluated via RMSE) and three classification (BAA, BLO, and ECH, evaluated via accuracy) datasets. XGBoost$_{\rm MR}$ refers to training XGBoost on the meta-representation, while {\ours}$_{\rm S}$ represents training only the top-layer model in \ours without utilizing the pre-trained model.}
    \centering
    \tabcolsep 2pt
\begin{tabular}{ccccc}
    \addlinespace
    \toprule
          & XGBoost & XGBoost$_{\rm MR}$ & {\sc TabPTM}$_{\rm S}$ & \ours \\
    \midrule
    STO $\downarrow$   & 0.8910 & 0.7809 & 0.7692 & 0.7355 \\
    ADO $\downarrow$   & 775.0   & 749.1 & 758.3 & 737.4 \\
    CAC $\downarrow$ & 0.143 & 0.141 & 0.137 & 0.132 \\
    \midrule
    BAA $\uparrow$  & 55.40  & 57.91  & 56.82  & 59.40 \\
    BLO $\uparrow$   & 76.93  & 77.71  & 76.41  & 78.03 \\
    ECH $\uparrow$   & 75.64  & 79.72  & 80.89  & 81.23 \\
    \bottomrule
    \end{tabular}   
    \label{tab:role}
\end{table}

\vspace{-2mm}
\section{Conclusion}
Considering the large amount of heterogeneous tabular datasets from many fields, we explore a way to pre-train a general model and extend its classification or regression ability to downstream datasets. We address the challenge of disparate attribute and label spaces across datasets via meta-representation. Our pre-trained \ours can be directly applied to unseen datasets or be fine-tuned without modifying the architecture. It achieves competitive performance in various scenarios, validating its effectiveness.

\bibliography{main}
\bibliographystyle{icml2025}

\newpage
\appendix
\onecolumn
There are five parts in the appendix:
\begin{itemize}
    \item \autoref{sec:related_appendix}: Additional related methods;
    \item \autoref{sec:kNN_view}: Motivation of meta-representation from nearest neighbor model;
    \item \autoref{appendix_sec_method}: More details and discussions of our \ours approach;
    \item \autoref{appendix_sec:exp_setup}: Details of the experimental setups;
    \item \autoref{appendix_sec:ablation}: Analysis of meta representation and additional ablation studies on \ours.
    \item \autoref{appendix_sec:whole_results}: The complete experimental results in the main text;
\end{itemize}
\begin{appendices}
\section{Additional Related Methods}\label{sec:related_appendix}
\noindent{\bf Learning with tabular data}. Tabular data is one of the most common data forms in many fields~\citep{richardson2007predicting,vanschoren2014openml,superconductivty}, and a lot of classical machine learning methods have been developed for tabular data, such as XGBoost~\citep{chen2016xgboost}, LightGBM~\citep{ke2017lightgbm}, and CatBoost~\citep{Prokhorenkova2018Catboost}.
Recently, researchers have tried to extend the success of deep neural networks such as multi-layer perceptron~\citep{GorishniyRKB21Revisiting}, Transformer~\citep{Huang2020TabTransformer}, and diffusion models~\citep{Kotelnikov2022TabDDPM} from visual and textual domains to the tabular fields~\citep{Borisov2022Deep}. 
Attribute embeddings~\citep{SongS0DX0T19AutoInt} and deep architectures have been designed~\citep{GuoTYLH17,Katzir2021Net,Chen2023Excelformer} for tabular data, and some simple baselines can achieve competitive results as the classical methods after carefully tuned~\citep{Kadra2021Well,Jeffares2023TANGOS}. Deep tabular models have the flexibility for various scenarios and can be incorporated well with the classical methods~\citep{Cheng2016Wide,WangFFW17DCN,ZivA22Tabular,Grinsztajn2022Why,ye2025modernnca}.

\noindent{\bf Reuse a heterogeneous tabular model}.
Instead of training a tabular model from scratch on a new task, reusing a pre-trained model from a related task becomes a useful choice, especially when efficiency is emphasized~\citep{Tommasi2014Learning,Kuzborskij2017Fast,Aghbalou2023Hypothesis}. 
In addition to the distribution shift between the pre-trained and the target tasks, the changes in their attribute spaces as well as the label spaces make the transfer of a tabular model challenging~\citep{HouZ18One,Ye2021Heterogeneous}. 
Tabular model reuse across heterogeneous datasets usually relies on some assumptions, \eg, the existence of a set of overlapped attributes between two datasets~\citep{Hou2022Prediction,levin2022transfer,Onishi2023TabRet,zhou2023unlocking} and the column meanings (the textual names of all attributes)~\citep{Wang2022TransTab}.

\noindent{\bf Learning with multiple tabular datasets}.
One more step to take advantage of the ability of deep neural networks on tabular fields is to pre-train a discriminative model on a large number of tabular datasets and extend its ability to downstream tasks. 
The in-distribution generalization ability of a pre-trained tabular model has been validated in multi-task learning~\citep{Argyriou2006Multitask,Zhang2022Multitask,Rubachev2022revisiting,Luetto2023One} and self-supervised learning~\citep{UcarHE21SubTab,BahriJTM22Scarf}, where all datasets are collected in a homogeneous form. In multi-view learning, a model is required to capture the consistent nature among heterogeneous views, but those multiple views of an instance are paired and share the same label space~\citep{Xu2013Multi}.
Some recent approaches utilize the deep neural network to pre-train a more generalizable tabular model, taking the difference in attributes and labels into account. 
One representative kind of approach assumes the existence of attribute names along with a dataset so that each instance could be transformed into a text, then a large language model could be applied to generalize the classification ability~\citep{Liu2022PTab,Hegselmann2022TabLLM,Zhang2023Generative,Wang2023AnyPredict}. Another thread of method learns shared components such as attribute-agnostic transformation across datasets, which provides a good model initialization for partial parameters given a downstream task~\citep{Iwata2020Meta,Liu2022Distribution,Zhang2023Meta,shen2023cross,zhu2023xtab,liu2025tabpfn}. 
We propose TabPTM to transform all datasets into a uniform form with meta-representation to enable the pre-training. Then, the pre-trained model could be applied directly to a downstream dataset, or fine-tuned without introducing additional parameters.

\noindent{\bf Meta-Representation}. The notion of meta-representation has been previously leveraged in multi-label learning to express the relationship between an instance and a specific label, which facilitates decoupling the label correlations~\citep{Yang2012Multilabel,Zhang2015Lift}. The meta-representation is also used together with the raw tabular features to assist text classification~\citep{Canuto2014On,Canuto2018Evaluation}.
In this paper, we provide a more general form of meta-representation, which contains both the distance values as well as the label information, working in both classification and regression tasks.
Moreover, we employ meta-representation as a pivotal tool to construct a pre-trained model that can effectively operate across multiple heterogeneous tabular datasets.
We also emphasize the metric-based variant and the strategies to deal with few-shot scenarios when a deep neural network is pre-trained over the meta-representations. 

\section{Meta-Representation from the Nearest Neighbor Perspective}\label{sec:kNN_view}
As mentioned in~\autoref{eq:kde_estimator} in~\autoref{sec:motivation}, we can formulate the problem of estimating the posterior density $\Pr(y_i \mid \vx_i, \gD)$ in the nearest neighbor form, and then represent the estimator with meta-representation.

Given an instance $\vx_i$, $K$NN calculates the distance between $\vx_i$ and other instances in $\gD$, assume the $K$ nearest neighbors (those $K$ nearest instances to $\vx_i$ based on a particular distance measure $\operatorname{dist}(\cdot,\cdot)$) are $\gN(\vx_i; \gD) = \{(\vx_1, y_1), \ldots, (\vx_K, y_K)\}$.
Then, the label $y_i$ of $\vx_i$ is predicted based on those labels in the neighbor set $\gN(\vx_i; \gD)$. For classification task with $C$ classes, we assume the one-hot form of a certain label $y_k$ in $\gN(\vx_i; \gD)$ is $\vy_k\in\{0,1\}^C$, then we have
\begin{equation}
    \hat{y}_i = \argmax_c \left(\sum_{(\vx_k,\vy_k)\in\gN(\vx_i; \gD)} \vy_k\right)_c\;,\label{eq:nn_cls}
\end{equation}
where the confidence of $C$ classes is calculated by voting from the neighbors, and the label is predicted by the elements with the largest confidence. While for the regression case, $y_k\in\mathbb{R}$, we have 
\begin{equation}
    \hat{y}_i = \frac{1}{K}\sum_{(\vx_k, y_k)\in\gN(\vx_i; \gD)} y_k\;.\label{eq:nn_reg}
\end{equation}
The prediction is the average of the labels of the neighbors. Thus, in both classification and regression cases, the nearest neighbor model transforms the prediction task over the whole dataset $\gD$ into a prediction task based on the local context $\mathcal{N}(\vx_i; \gD)$. The neighborhoods of different instances reveal the property of an instance, and vary significantly.

Instead of considering all instances in the neighborhood equally, a variant of the nearest neighbor model re-weights the influence of the neighbors, \ie, making the nearer neighbors have more weight in the prediction~\cite{bishop2006pattern,RasmussenW06,Vinyals2016Matching}. Given a vector whose elements are the normalized distance between $\vx_i$ and its neighbors:
\begin{equation}
    \hat{\phi}(\vx_i) = \operatorname{softmax}\left(\left[-\operatorname{dist}(\vx_i, \vx_1),\ldots,-\operatorname{dist}(\vx_i, \vx_{K})\right]\right)\;,\label{eq:distance_vec}
\end{equation}
with $\hat{\phi}(\vx_i)\in\mathbb{R}_+^K$, where the softmax operator $\operatorname{softmax}(\cdot)$ makes the sum of weights equals one. Thus, the prediction for classification in~\autoref{eq:nn_cls} could be updated by
\begin{equation}
    \hat{y}_i = \argmax_c \;\left(\hat{\phi}(\vx_i)^\top Y_K\right)_c\;,
    \label{eq:nn_weihgt_cls}
\end{equation}
where we define $Y_K\in\{0,1\}^{K\times C}$ as the set of one-hot labels in $\mathcal{N}(\vx_i;\mathcal{D})$, and each row of $Y_K\in\{0,1\}$ is the one-hot label of an instance. 
In addition, define the label vector $[y_1, \ldots, y_K]\in\mathbb{R}^K$, the weighted version of regression prediction in~\autoref{eq:nn_reg} becomes
\begin{equation}
    \hat{y}_i = \;\hat{\phi}(\vx_i)^\top [y_1, \ldots, y_K]\;.\label{eq:nn_weight_reg}
\end{equation}

Based on the prediction of $K$NN in~\autoref{eq:nn_weihgt_cls} and~\autoref{eq:nn_weight_reg}, the label of an instance is determined by both the transformed distance vector $\hat{\phi}(\vx_i)$ to the neighbors and their corresponding labels. 
Therefore, we set our meta-presentation as the concatenation of both the distance values and labels, which captures instance and label distribution in a local context. We then learn a transformation ${\bf T}_{\Theta}$ in~\autoref{eq:transformation_reg} to map the meta-representation to the label and optimize the joint model over multiple heterogeneous tabular datasets.

The distance metric in $K$NN determines the set of neighbors and influences the discriminative ability~\cite{XingNJR02,WeinbergerS09}. By learning a metric that pulls similar instances together and pushes dissimilar ones away from each other, the classification and regression ability of nearest neighbor models is improved~\cite{Kulis13,2015BelletHS}. 
The learned distance metric is directly related to the number of attributes (dimension), and several special strategies, such as semantic mapping, are required to apply a learned metric from one dataset to another~\cite{ZhangY10,KulisSD11,YangHL11}. 
Instead of learning a separate distance metric for each tabular task, we use mutual information to select related features given a tabular task automatically in~\autoref{eq:weighted_dist}, which acts as an adaptive metric in \ours. The learned weights of \ours are shared among different heterogeneous tasks.

\section{Details and Discussions of \ours}\label{appendix_sec_method}
In this section, we describe the detailed architecture of \ours as well as the whole training flow. Finally, we discuss the relationship between \ours and some related methods.

\subsection{Details of the Score Transformation}
In~\autoref{eq:transformation}, we utilize a transformation ${\bf T}_{\Theta}$ to map the meta-representation to the prediction score of an instance, for all $C$ classes or the regression score. The transformation could be implemented via various kinds of deep neural networks. 

We describe the detailed architectures of the deep model following~\citep{GorishniyRKB21Revisiting}. Multi-Layer Perceptron (MLP) contains several layers of non-linear blocks
\begin{align}
{\bf MLP}(\vx) & =\operatorname{Linear}(\operatorname{MLPBlock}(\ldots(\operatorname{MLPBlock}(\vx)))) \\
\operatorname{MLPBlock}(\vx) & =\operatorname{Dropout}(\operatorname{ReLU}(\operatorname{Linear}(\vx)))\;.
\end{align}
The $\operatorname{Linear}$ block means a fully connected layer with linear projection.
In the classification scenario, MLP maps the class-wise meta-representation $\phi_c(\vx_i)$ to the prediction score $s(\vx_i)_c$ (a scalar):
\begin{equation}
    s(\vx_i)_c = {\bf MLP}(\phi_c(\vx_i)),\;\forall c=1,\ldots,C\;. \label{eq:mlp_mapping2}
\end{equation}
Although this is a one-to-one mapping from the class-specific meta-representation to the confidence score, we validate its effectiveness in our experiments. In the regression scenario, a single MLP is applied to all the inputs.

The mapping ${\bf T}$ could also be implemented with Residual Network (ResNet)~\citep{He2016ResNet} and Transformer~\citep{vaswani2017attention}. For example, ResNet has the following architecture: 
\begin{align}
{\bf ResNet}(\vx) & =\operatorname{Prediction}\left(\operatorname{ResNetBlock}\left(\ldots\left(\operatorname{ResNetBlock}\left(\operatorname{Linear}\left(\vx\right)\right)\right)\right)\right) \\
\operatorname{ResNetBlock}(\vx) & =\vx+\operatorname{Dropout}(\operatorname{Linear}(\operatorname{Dropout}(\operatorname{ReLU}(\operatorname{Linear}(\operatorname{BatchNorm}(\vx)))))) \\
\operatorname{ Prediction }(\vx) & =\operatorname{ Linear }(\operatorname{ReLU}(\operatorname{BatchNorm}(\vx)))\;.
\end{align}
Different from MLP, ResNet has a residual link from its input to the output, and Batch Normalization~\citep{Ioffe2015Batch} is introduced in the building block of ResNet. 

We investigate several choices of ${\bf T}$ in our experiments. We find the simple MLP (\eg, with three layers) is competitive in most cases. Applying more complicated ResNet and Transformer cannot obtain obvious improvements. Therefore, we set ${\bf T}$ with MLP, and better architectures can be explored in further research. 

\subsection{The Pre-training and Downstream Application Workflow}
\ours is a classification/regression model based on the meta-representation. The meta-representation transforms an instance into a $K$-dimensional form no matter how many dimensions it originally had, and the meta-representation extractor does not contain any learnable parameters. 

The pre-training objective in~\autoref{eq:objective} is constructed based on the loss function of various pre-training tabular datasets (seen datasets). 
The indexes of nearest neighbors for each instance $\vx_i$ are calculated at the beginning of the optimization and recorded for later use.  
We optimize~\autoref{eq:objective} in a stochastic manner. 
In particular, we randomly select a seen tabular dataset and randomly sample a mini-batch from the dataset in each iteration. 
For each sampled instance, we calculate the metric-based meta-representation based on its neighborhood. For classification, we make predictions with~\autoref{eq:prediction}, while for regression, we obtain the predicted label directly from the mapped results ${\bf T}_\Theta(\Phi(\vx_i))$.
The pre-training phase of \ours on multiple heterogeneous tabular datasets is summarized in Algorithm~\autoref{alg:pretrain}.

The learned \ours could be deployed with the following two options.

{\bf Direct Generalization}. Since all tabular instances could be transformed into the homogeneous $\Phi(\vx_i)$ regardless of its original dimension, the learned \ours can be applied to a downstream tabular dataset directly. 
We use classification on a downstream dataset with unseen classes and attributes as an example, and the results are also reported in our experiments.
Given a new dataset $\gD_u$ and a corresponding test instance $\vx_*^u$, we calculate the meta-representation of $\vx_*^u$ w.r.t. each one of the $C_u$ unseen classes. The distances between $\vx_*^u$ to the nearest neighbors in $\gD_u$ and the labels of those neighbors are utilized to obtain the meta-representation. 
We obtain $C_u$ meta-representations with $K$-dimension, one for each class. Then we apply the learned model ${\bf T}_\Theta$ over them to obtain the $C_u$-dimension class confidence vector without additional training. The instance is classified as one of the unseen classes by selecting the index with the maximum confidence.
We summarize this workflow in Algorithm~\autoref{alg:downstream}. The direct deployment requires the search of nearest neighbors in the downstream dataset, which is fast and could be accelerated by some off-the-shelf methods.

{\bf Fine-Tuned Generealization}. Another choice is to use the pre-trained \ours as initialization and fine-tune the weights over the downstream task $\gD_u$. In other words, the following objective is minimized with gradient descent for several steps
\begin{equation}
    \min_{\Theta}\; \sum_{(\vx_i^u, y_i^u)\sim\gD_u} \ell({\bf T}_\Theta(\vx_i^u), y_i^u)\;.
\end{equation}
The optimization starts from the learned weights $\Theta$ from the pre-training stage. 
Additional learnable parameters are usually added when fine-tuning a pre-trained tabular model~\cite{zhu2023xtab}, and the size of parameters is related to the dimension and class number of a downstream tabular data.
In \ours, it is notable that no additional learnable parameters are required to be added in this deployment stage, which makes the fine-tuning efficient and avoids using special hyper-parameters for different sets of parameters.

\begin{algorithm}[t]
	\caption{Pre-training on multiple heterogeneous tabular datasets.}
    \label{alg:pretrain}
	\begin{algorithmic}[1]{
			\REQUIRE {$T$ tabular training set $\sD=\{\gD_1, \ldots, \gD_T\}$, the initialized model ${\bf T}_{\Theta}$
            }
			\FORALL{iteration = 1,...} 
			\STATE {Sample a dataset $\gD_t$ from $\sD$}
			\STATE {Sample a mini-batch with $B$ instances $\{\vx^t_{i},y^t_{i}\}_{i=1}^B$}			\FORALL{$(\vx^t_{i},y^t_{i})$}
			\STATE {Get metric-based meta-representation $\Phi(\vx^t_{i})$ based on nearest neighbors in $\gD_t$}
			\STATE {Obtain the prediction score ${\bf T}_{\Theta}(\Phi(\vx^t_{i}))$ }
            \IF{classification}
			\STATE {Predict via $\hat{y}_i^t = \argmax_c\;{\bf T}_{\Theta}(\Phi(\vx^t_{i}))=\left\{s(\vx^t_{i})_1,\ldots,s(\vx^t_{i})_{C_t}\right\}$}
            \ELSE
			\STATE {Predict via $\hat{y}_i^t = {\bf T}_{\Theta}(\Phi(\vx^t_{i}))$ for regression}
            \ENDIF
            \STATE {Compute loss $\ell(\hat{y}_i^t, y_i^t)$}
			\ENDFOR
			\STATE {Accumulate $B$ losses as Eq.~\ref{eq:objective}}
			\STATE {Update $\Theta$ with SGD}
			\ENDFOR
   \STATE {\bf Return: }{Pre-trained model $\Theta$}
		}
	\end{algorithmic}
\end{algorithm}
\begin{algorithm}[t]
	\caption{Apply the pre-trained model to the downstream tabular dataset.}
	\label{alg:downstream}
	\begin{algorithmic}[1]{
			\REQUIRE {Tabular training set $\gD_u$, test instance $\vx_*^u$, the learned model ${\bf T}_{\Theta}$}
			\STATE {Compute the metric-based meta-representation $\Phi(\vx_*^u)$ for $\vx_*^u$}
            \IF{classification}
            \STATE {
            The meta-representation $\Phi(\vx_*^u)=\{(\phi_c(\vx_*^u))\}_{c=1}^{C_u}$
            }
			\STATE {Obtain the prediction score $\{{\bf T}_{\Theta}(\phi_c(\vx_*^u))\}_{c=1}^{C_u}$ }
			\STATE {Predict via $\hat{y}_*^u = \argmax_c\left\{s(\vx_*^u)_1,\ldots,s(\vx_*^u)_{C_u}\right\}$}
            \ELSE
			\STATE {Predict via $\hat{y}_*^u = {\bf T}_{\Theta}(\phi(\vx_*^u))$}
            \ENDIF
			\STATE {\bf Return: } {The predicted label of $\vx_*^u$}
		}
	\end{algorithmic}
\end{algorithm}

\section{Details of Experimental Setups}\label{appendix_sec:exp_setup}

\begin{table}[t]
  \tabcolsep 3pt
  \centering
  \caption{The detailed statistics of all tabular datasets. ``Abbr.'' means the abbreviation of the name of the tabular dataset. The $C$ and $N$ denote the class number and the instance number of the datasets. There are two types of attributes with numerical and categorical values, and we denote their numbers as ``Num.'' and ``Cat.'', respectively.}
  \begin{adjustbox}{width=1.0\textwidth,center}
   \begin{tabular}{lcccccc|lcccccc}
    \addlinespace
    \toprule
    {\bf Name} & {\bf Abbr.} & {\bf Task type} & {\bf C} & {\bf N} & {\bf Num.} & {\bf Cat.} & {\bf Name} & {\bf Abbr.} & {\bf Task type} & {\bf C} & {\bf N} & {\bf Num.} & {\bf Cat.} \\
    \midrule
    Another-Dataset-on-used-Fiat-500-(1538-rows)   & ADO            & regression          & 1 & 1538       & 6             & 0      & Bank\_Customer\_Churn\_Dataset & BCC   & binclass & 2     & 10000 & 6     & 4 \\
    archive2                                       & ARC            & regression          & 1 & 1143       & 11            & 1        & CDC\_Diabetes\_Health\_Indicators & CDH   & binclass & 2     & 253680 & 7     & 14 \\
    archive\_r56\_Maths                            & ARM            & regression          & 1 & 397        & 1             & 29  & E-CommereShippingData & ECS   & binclass & 2     & 10999 & 6     & 4 \\
    archive\_r56\_Portuguese                       & ARP            & regression          & 1 & 651        & 1             & 29    & Fitness\_Club\_c & FCC   & binclass & 2     & 11000 & 10    & 5 \\
    airfoil\_self\_noise                           & ASN            & regression          & 1 & 1503       & 5             & 0   & banknote\_authentication & BAA   & binclass & 2     & 1382  & 4     & 0 \\
    analcatdata\_supreme                           & ASU            & regression          & 1 & 4052       & 7             & 0  & kc2   & KC2   & binclass & 2     & 522   & 21    & 0 \\
   auction\_verification                          & AVE            & regression          & 1 & 2043       & 6             & 1       & maternal\_health\_risk & MHR   & multiclass & 3     & 1014  & 6     & 0 \\
   Bias\_correction\_r                            & BCR            & regression          & 1 & 7725       & 21            & 0    & seismic+bumps & SEB   & binclass & 2     & 2584  & 14    & 5 \\
  communities\_and\_crime                        & CAC            & regression          & 1 & 1994       & 102           & 0   & sports\_articles\_for\_objectivity\_analysis & SAF   & binclass & 2     & 800   & 57    & 2 \\
   combined\_cycle\_power\_plant                  & CCP            & regression          & 1 & 9568       & 4             & 0    & turiye\_student\_evaluation & TSE   & multiclass & 5     & 5820  & 30    & 2 \\
   concrete\_compressive\_strength                & CCS            & regression          & 1 & 1030       & 8             & 0      & water\_quality & WAQ   & binclass & 2     & 7996  & 20    & 0 \\
   1000-Cameras-Dataset                           & CDA            & regression          & 1 & 1038       & 10            & 0   & blood & BLO   & binclass & 2     & 748   & 4     & 0 \\
    Contaminant-detection                          & CDI            & regression          & 1 & 2400       & 30            & 0    & breast-cancer & BRC   & binclass & 2     & 286   & 9     & 0 \\
   dataset\_sales                                 & DAT            & regression          & 1 & 10738      & 10            & 0   & breast-cancer-wisc & BCW   & binclass & 2     & 699   & 9     & 0 \\
   debutanizer                                    & DBT            & regression          & 1 & 2394       & 7             & 0    & breast-cancer-wisc-prog & BCP   & binclass & 2     & 198   & 33    & 0 \\
   3D\_Estimation\_using\_RSSI\_of\_WLAN\_dataset & EUR            & regression          & 1 & 5760       & 6             & 0    & dermatology & DER   & multiclass & 6     & 366   & 34    & 0 \\
   Goodreads-Computer-Books                       & GCB            & regression          & 1 & 1234       & 5             & 0       & echocardiogram & ECH   & binclass & 2     & 131   & 10    & 0 \\
    housing\_price\_prediction                     & HPP            & regression          & 1 & 545        & 5             & 7          & heart-cleveland & HEC   & multiclass & 5     & 303   & 13    & 0 \\
   Is-this-a-good-customer                        & ITA            & regression          & 1 & 1723       & 9             & 4        & Basketball\_c & BAS   & binclass & 2     & 1340  & 11    & 0 \\
  Parkinson\_Multiple\_Sound\_Recording          & PMS            & regression          & 1 & 1040       & 26            & 0     & diabetes & DIA   & binclass & 2     & 202944 & 3     & 18 \\
   puma8NH                                        & PNH            & regression          & 1 & 8192       & 8             & 0 & mice\_protein\_expression & MIC   & multi-class & 8     & 1080  & 77    & 0 \\
   puma32H                                        & PUH            & regression          & 1 & 8192       & 32            & 0     & Wilt  & WIL   & binclass & 2     & 4821  & 5     & 0 \\
    Student\_Alcohol\_Consumption                  & SAC            & regression          & 1 & 395        & 13            & 17      & bank\_marketing & BAN   & binclass & 2     & 45211 & 7     & 7 \\
   Shop\_Customer\_Data                           & SCA            & regression          & 1 & 2000       & 4             & 2      & statlog\_german\_credit\_data & STA   & binclass & 2     & 1000  & 7     & 13 \\
   stock\_fardamento02                            & SFA            & regression          & 1 & 6277       & 5             & 1       & company\_bankruptcy\_prediction & COM   & binclass & 2     & 6819  & 93    & 2 \\
   sulfur                                         & SFU            & regression          & 1 & 10081      & 6             & 0   & drug\_consumption & DRU   & multiclass & 7     & 1884  & 12    & 0 \\
    Shipping                                       & SHP            & regression          & 1 & 10999      & 5             & 4  & dry\_bean\_dataset & DRY   & multiclass & 7     & 13611 & 16    & 0 \\
    shrutime                                       & SHR            & regression          & 1 & 10000      & 4             & 6      & internet\_firewall & INT   & multiclass & 4     & 65532 & 7     & 0 \\
    satellite\_image                               & SIM            & regression          & 1 & 6435       & 36            & 0     & heart-hungarian & HEA   & binclass & 2     & 294   & 12    & 0 \\
   Student\_Performance\_Portuguese               & SPP            & regression          & 1 & 397        & 15            & 17   & heart-va & HEV   & multiclass & 5     & 200   & 12    & 0 \\
    stock                                          & STO            & regression          & 1 & 950        & 9             & 0        & breast-cancer-wisc-diag & BRE   & binclass & 2     & 569   & 30    & 0 \\
   svmguide3                                      & SVM            & regression          & 1 & 1243       & 22            & 0   & mammographic & MAM   & binclass & 2     & 961   & 5     & 0 \\
    VulNoneVul                                     & VNV            & regression          & 1 & 5692       & 16            & 0     & parkinsons & PAR   & binclass & 2     & 195   & 22    & 0 \\
    wine+quality                                   & WQA            & regression          & 1 & 6497       & 11            & 0    & post-operative & POS   & multiclass & 3     & 90    & 8     & 0 \\
    Wine\_Quality\_white                           & WQW            & regression          & 1 & 4898       & 11            & 0              & primary-tumor & PRI   & multiclass & 15    & 330   & 17    & 0 \\
    Waterstress                                    & WTS            & regression          & 1 & 1188       & 22            & 0   & spect & SPE   & binclass & 2     & 265   & 22    & 0 \\
    \bottomrule
    \end{tabular}
    \end{adjustbox}
  \label{tab:dataset}
\end{table}

\subsection{Datasets}
We experiment with 72 tabular datasets, which contain 36 datasets for classification and 36 datasets for regression~\footnote{We do not consider datasets and temporal splits in TabReD~\cite{rubachev2025tabred} in our experiments, as it requires specialized robust learning techniques to bridge discrepancies between training and test sets, which is beyond the scope of this study.}. The statistics of all the datasets are listed in~\autoref{tab:dataset}. 
We use $C$ and $N$ to denote the class number and the instance number of the datasets. For a regression task, we set $C=1$. There are two types of attributes with numerical and categorical values, and we denote their numbers as ``Num.'' and ``Cat.'', respectively. The datasets are collected from UCI Machine Learning Repository~\footnote{https://archive.ics.uci.edu/} or OpenML~\footnote{https://www.openml.org/}.

We use the same protocol for classification and regression. Given 36 datasets, we randomly split them into two sets, each with 18 datasets. The experiments in the main paper utilize the first 18 datasets to pre-train \ours, and set the remaining ones as downstream datasets. In other words, we pre-train the model using 18 heterogeneous tabular datasets, and evaluate its generalization ability on 18 datasets with different sizes. We also evaluate other configurations of the pre-training and downstream split. For example, we use the 18 datasets in the second part to pre-train the model and the first 18 ones as the downstream datasets. The additional results are reported in~\autoref{tab:other_pretrain_reg} and~\autoref{tab:other_pretrain_cls}.

To further assess the few-shot capabilities of \ours, we incorporate TinyBenchmark from~\cite{ye2024closer} as an additional set of downstream datasets, which do not overlap with the original 36 classification datasets. TinyBenchmark comprises 29 classification datasets and has been shown to produce consistent average ranking results, aligning with evaluations conducted over 300 datasets.

For each dataset, we define five distinct shot settings: 4, 8, 16, 32, and 64. For example, in the 64-shot setting, we randomly sample 64 instances per class from the original training set, repeating this process five times to generate five unique 64-shot tasks. In total, our few-shot experiments cover $36 + 29 = 65$ datasets, leading to:
\[
65 \text{ (datasets)} \times 5 \text{ (shot settings)} \times 5 \text{ (random sampling repetitions)} = 1625 \text{ tasks}
\]

\subsection{Additional Implementation Details}
We compare our \ours with different types of methods and describe the detailed way we tune their hyper-parameters in this subsection. We mainly follow the setups in ~\citep{GorishniyRKB21Revisiting} to determine the hyper-parameters.

\noindent{\bf Classical methods and deep tabular methods}.
Both classical tabular methods (SVM, XGBoost) and standard deep methods (MLP) are trained for each dataset separately. We use the official hyper-parameter search spaces for deep tabular methods (FT-T, TabCaps, and TabR). We tune their hyper-parameters and carry out early stopping on the corresponding validation set of a given dataset. All hyper-parameters are selected by Optuna library\footnote{https://optuna.org/} with Bayesian optimization over 30 trials~\cite{GorishniyRKB21Revisiting}. The best hyper-parameters are used and the average accuracy/RMSE over 10 different random seeds is calculated. 

\noindent{\bf Pre-training and fine-tuning approaches}.
For TabPFN, we utilize the best official checkpoint, then we apply TabPFN on a downstream dataset with its in-context learning ability.
For XTab, We reuse the checkpoint with the highest number of training epochs from the official implementation, then we perform evaluations on the target datasets using XTab's light fine-tuning approach. 
For DEN, we divide all pre-training datasets into binary and multiclass groups. Each group is then used to train models on their corresponding downstream unseen datasets. We set the learning rate as 0.001 and fine-tune the transform block on the downstream tasks. 

\noindent{\bf \ours}.
We implement our \ours with a three-layer MLP. 
The combination of three distances, namely Euclidean distance, Manhattan distance, and 
Bray-Curtis distance, are utilized. The influence of distances are investigated in~\autoref{appendix_sec:ablation}.
During the pre-training, we randomly sample 1024 examples from a seen dataset in each iteration. 
When we fine-tune \ours in~\autoref{tab:reg_main} and~\autoref{tab:cls_main}, we set the learning rate as 0.01 and fine-tune the whole model for 30 epochs. The model in the last epoch is saved for evaluation.

\section{Additional Experiments and Analyses}\label{appendix_sec:ablation}
We analyze the properties of \ours from the following aspects.

\subsection{Direct Generalization of \ours}
The learned \ours has the ability to make predictions directly given a downstream dataset. In few-shot tasks, \ours does not undergo fine-tuning; instead, it constructs a meta-representation and directly makes predictions using the pretrained model. In full-shot tasks, the results of this variant, denoted by {\sc TabPTM}$_{\rm D}$ are included in~\autoref{tab:reg_main_whole} and \autoref{tab:cls_main_whole}.
We find that the direct generalization of the pre-trained \ours is competitive in some cases (\eg, on ``ADO'', ``DER'', and ``ECH''), but fine-tuning \ours is still necessary on most full-shot tasks.

The inconsistency between \ours' training-free capability in few-shot tasks and its performance in full-shot tasks is primarily due to two factors. In few-shot scenarios, training-free inference helps mitigate overfitting, which could otherwise occur when fine-tuning with very limited data. The pre-trained \ours serves as a strong inductive bias, ensuring robustness and preventing the model from being overly sensitive to small sample variations. However, in full-shot settings, where each class may contain thousands of samples, a significantly larger amount of task-specific information becomes available. These datasets often exhibit complex intra-class variations that require more fine-grained decision boundaries. Without fine-tuning, the model may struggle to capture these nuances, leading to suboptimal performance.

\subsection{Analysis of Meta Representation}
\noindent{\bf Richer supervision helps in classification}. 
As we mentioned in~\autoref{eq:meta_representation}, an auxiliary label $\tilde{y}_j$ is appended after the distance value to provide the label information in the meta-representation. We set $\tilde{y}_j$ following a one-vs.-rest manner in our implementation, which differentiates those instances in the neighborhood belonging to a certain class. Here, we investigate whether richer supervision in $\tilde{y}_j$ helps the classification task.

Given a binary downstream dataset $\gD_u$, we first train XGBoost with default hyper-parameters (denoted as ``teacher''), whose predictions (scalars) on the training set instances are used as a kind of richer supervision to set $\tilde{y}_j$. 
We train \ours from scratch on the downstream dataset, denoted by {\sc TabPTM}$_{\rm S}$. We also train \ours with the supervision-enriched meta-representation from scratch on the downstream dataset, denoted by {\sc TabPTM}$_{\rm rich}$. Finally, we include the results of the fine-tuned \ours pre-trained from various heterogeneous datasets. The results are reported in~\autoref{tab:distill}.

\begin{table}[tbp]
  \centering
  \caption{Average accuracy on four classification datasets in~\autoref{tab:cls_main}. We use XGBoost with default hyper-parameters as the ``teacher'', whose predictions on an instance are appended to the meta-representation as enriched supervision. \ours variants trained on a given dataset from scratch with vanilla and enriched supervision are denoted as {\sc TabPTM}$_{\rm S}$ and {\sc TabPTM}$_{\rm rich}$. 
  }
    \begin{tabular}{cccccc}
    \addlinespace
    \toprule
    Dataset & \ours & XGBoost & Teacher & {\sc TabPTM}$_{\rm S}$ & {\sc TabPTM}$_{\rm rich}$ \\
    \midrule
    ECS & 67.40 & {67.80} & 66.90 & 64.00 & 67.20 \\
    BAA & {59.40} & 55.40 & 53.80 & 45.50 & 54.20 \\
    SEB & {92.80} & 92.70 & 92.60 & {92.80} & 92.70 \\
    SAF & 81.90 & {83.20} & 81.00 & 78.00 & 80.20 \\
    \bottomrule
\end{tabular}
  \label{tab:distill}
\end{table}

XGBoost with default hyper-parameters performs better than the vanilla trained {\sc TabPTM}$_{\rm S}$ in most cases, so we set it as the teacher. With supervision-enriched meta-representation, we find {\sc TabPTM}$_{\rm rich}$ outperforms {\sc TabPTM}$_{\rm S}$, which validates our assumptions that we can improve the discerning ability of the model by introducing richer supervision into the meta-representation. However, {\sc TabPTM}$_{\rm S}$ cannot achieve as good results as \ours, so by training a joint model on heterogeneous tabular datasets, it indeed learns shareable experience to achieve more discriminative models. 

\noindent{\bf Possible explanations of the better regression results}.
The a bit improvement of regression results to the classification ones may result from the form of meta-representation and the sharing of prediction experience in \ours.

Regression tasks may benefit more from the local context. The core idea of meta-representation is to make predictions in a local context (based on the neighborhood of an instance). Then, the model's ability depends on the ``locality'' of the true hypothesis, \ie, whether the decision function of an instance depends on its neighborhood or not. Since we may provide estimates of the regression function or conditional expectation by specifying the nature of the local neighborhood~\cite{OrmoneitH99,HastieTF09,KpotufeG13}, it is more probable to predict the continuous label of a center instance based on the weighted labels of its neighbors. Therefore, making predictions via meta-representation may help regression tasks.

More shareable knowledge in regression. As depicted in~\autoref{fig:pilot_study}, the strategy for inferring an instance's label from the distribution of distances to its neighbors is applicable across heterogeneous tabular tasks. For classification tasks, if an instance resides within a high-density region of a class (akin to being near the class center), the majority of values in the meta-representation would typically be small, indicating close proximity to neighboring instances of that class. Conversely, if only a few values in the meta-representation are small, while most are large, it indicates that the instance is likely located at the boundary among classes. Such a kind of discerning strategy is more shareable in regression tasks, which works in a way that determines the label via the weighted combination of its neighbors. Therefore, using a pre-trained model (the top-layer MLP in \ours) can be generalized to downstream tasks.

Rich label information in regression. In addition to the distance values between a given instance and its nearest neighbors, we append the labels of the neighbors into the meta-representation. The labels in regression tasks are continuous, and even if two instances are close to each other, they may have different label values, which introduces rich information. In contrast, since the labels in classification tasks are discrete, two close instances may have the same label and only differ in their distance values in meta-representation. In summary, the labels in regression tasks may introduce richer supervision. We validate this assumption in the previous experiments. We use the prediction of a well-trained XGBoost as the label of an instance in meta-representation, which transforms the labels in classification tasks into continuous ones. The experiments validate that the \ours equipped with such continuous labels outperforms the original one, which explains the different results in regression and classification to some extent.

\noindent{\bf The influence of the metric on meta-representation}.
We compare the results when we pre-train \ours over the vanilla meta-representation and its metric-based variant for regression tasks. Since different regression tasks have diverse label ranges, we compare the change of average rank with and without the metric. \ours gets an average rank of 1.941 over the 18 datasets in~\autoref{tab:reg_main}, compared with 12 methods in total. If the metric is not used, the average rank increases to 2.765 (the lower, the better). The results clearly indicate that the distance metric filters out redundant and noisy attributes, which is necessary to improve the generalization ability of \ours.

\subsection{The Influence of the Dimension of the Meta-representation}
\begin{figure}[t]
    \centering
    \begin{minipage}{0.4\linewidth}
    \includegraphics[width=\textwidth]{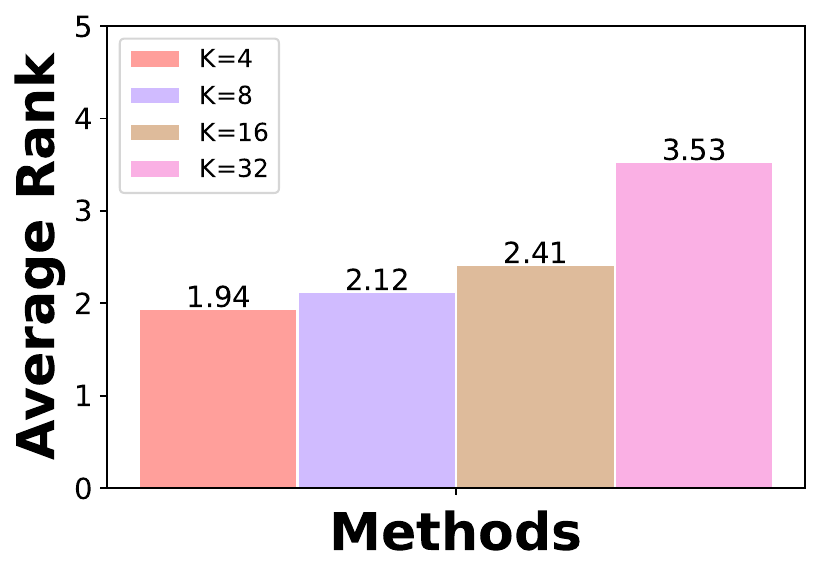}
    \centering
    {\mbox{(a) {Regression}}}
    \end{minipage}
    \begin{minipage}{0.4\linewidth}
    \includegraphics[width=\textwidth]{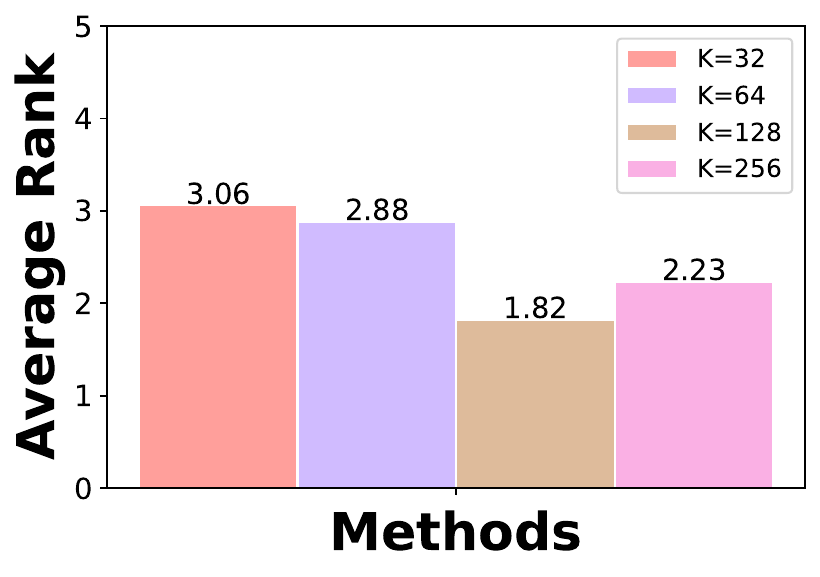}
    \centering
    {\mbox{(b) {Classification}}}
    \end{minipage}
    \caption{Average rank (the lower, the better) on 18 downstream datasets as in~\autoref{tab:reg_main} and~\autoref{tab:cls_main}. Different dimension values $K$ of meta-representation are used to pre-train \ours.}
    \label{fig:few-dimension}
\end{figure}

In~\autoref{eq:meta_representation}, we consider the nearest $K$ neighbors in the training set of each class in classification and $K$ neighbors in the whole training set for regression. Then the size of the meta-representation is related to $K$. In full-shot tasks, we set $K=16$ for regression and $K=128$ for classification by default in previous experiments.
We pre-train \ours over meta-representations with different dimensions $K$ in~\autoref{fig:few-dimension}, where we show the change of average rank on 18 downstream datasets when comparing 4 values of $K$. We find different types of downstream datasets may prefer various dimension values.

\subsection{The Influence of the Distances in Meta-representation}
\begin{figure}[t]
    \centering
    \begin{minipage}{0.4\linewidth}
    \includegraphics[width=\textwidth]{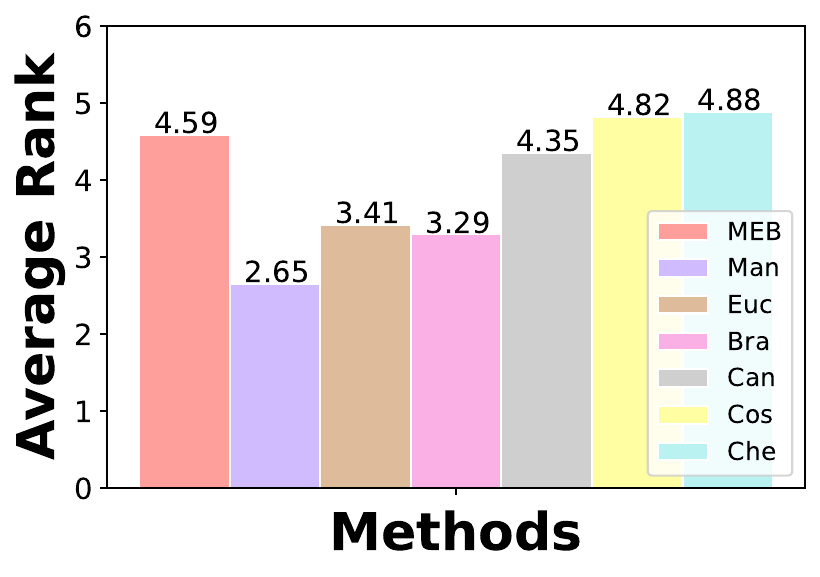}
    \centering
    {\mbox{(a) {Regression}}}
    \end{minipage}
    \begin{minipage}{0.4\linewidth}
    \includegraphics[width=\textwidth]{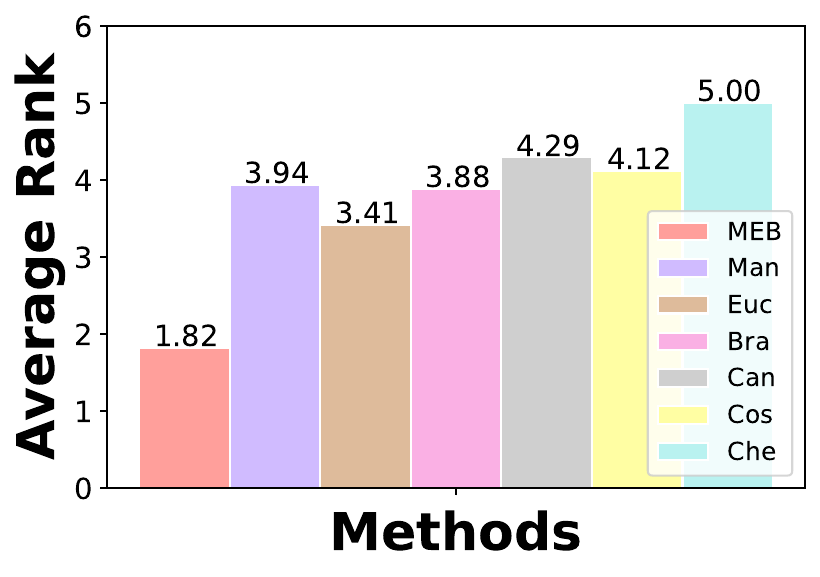}
    \centering
    {\mbox{(b) {Classification}}}
    \end{minipage}
    \caption{Average rank (the lower, the better) on 18 downstream datasets as in~\autoref{tab:reg_main} and~\autoref{tab:cls_main}. Different distances or their combinations are used to obtain the meta-representation. ``Man'' denotes Manhattan distance, ``Euc'' denotes the Euclidean distance, ``Bra'' denotes Bray-Curtis distance, ``CAN'' denotes Canberra distance, ``COS'' denotes cosine distance,  and ``CHE'' denotes Chebyshev distance. ``MEB'' denotes the combination of Manhattan, Euclidean, and Bray-Curtis distances.
}
    \label{fig:distance_choice}
\end{figure}

In~\autoref{eq:meta_representation}, a certain distance is utilized to obtain the nearest neighbors in the training set to construct the meta-representation. The indexes of instances, as well as the distance values, depend on the choice of distances. 
We consider several distance measures (equipped with the adaptive metric in~\autoref{eq:weighted_dist}. The average rank on 18 downstream datasets as in~\autoref{tab:reg_main} and~\autoref{tab:cls_main} based on different distances as well as their combinations are investigated. ``Man'' denotes Manhattan distance, ``Euc'' denotes the Euclidean distance, ``Bra'' denotes Bray-Curtis distance, ``CAN'' denotes Canberra distance, ``COS'' denotes cosine distance,  and ``CHE'' denotes Chebyshev distance. ``MEB'' denotes the combination of Manhattan, Euclidean, and Bray-Curtis distances.
The results in~\autoref{fig:distance_choice} show the change of average rank on 18 downstream datasets when comparing seven distance choices. The Manhattan distance works the best for regression, and the combination of three distances is the best choice for classification. We set the latter one as the default distance choice in \ours.

\subsection{The influence of pre-training Size}\label{sec:pre-train-size}
To investigate the impact of varying pre-training data size on the performance of \ours, we split the 18 pre-training datasets into two groups: one containing 6 datasets (HEA, HEV, BRE, MAM, PAR, POS) and another containing 12 datasets, which included the initial 6 plus PRI, SPE, DIA, MIC, WIL, and BAN. All other experimental settings remained consistent with those in \autoref{tab:cls_main}. 
We also compared the performance of models trained directly on individual downstream datasets using the same meta-representation. Hyper-parameters were optimized using the Optuna library, with 30 trials and a maximum of 200 training epochs. This approach is referred to as ``{\sc TabPTM}$_{\rm S}$''.

We randomly selected six datasets as downstream tasks, and their accuracy, along with the average performance rank across the four configurations over 18 downstream datasets, is presented in \autoref{tab:pre-training-size}. The results indicate that pre-training on a larger and more diverse set of datasets enhances \ours's ability to learn transferable knowledge for tabular predictions.
Even with only 30 epochs of fine-tuning, \ours significantly outperformed models that were fully trained for 200 epochs on downstream datasets.

\begin{table}[tbp]
\centering
\tabcolsep 2pt
\caption{The change of classification accuracy when we increase the size of the pre-training datasets. 
In addition to three sizes of the pre-training datasets, we also list the results training \ours directly on a downstream dataset, corresponding to {\sc TabPTM}$_{\rm S}$. 
We list the accuracy of six randomly selected datasets and the average rank of four variants over the 18 downstream datasets. 
By enlarging the pre-training size, we find \ours has enhanced performance for \ours.}
\begin{tabular}{lcccc}
\addlinespace
\toprule
Dataset & 6 datasets & 12 datasets & 18 datasets & {\sc TabPTM}$_{\rm S}$ \\
\midrule
BAA & 51.10 & 52.50 & {\bf 59.40} & 55.40 \\
BCP & 71.80 & 73.80 & {\bf 82.00} & 77.00 \\
BLO & 71.80 & 75.10 & {\bf 78.00} & 76.90 \\
ECH & 72.20 & 73.30 & {\bf 81.20} & 72.60 \\
ECS & 65.70 & 66.10 & {\bf 67.40} & 67.30 \\
KC2 & 79.00 & 78.70 & {\bf 83.10} & 78.70 \\
\midrule
avg. rank & 3.667 & 2.833 & 1.000 & 2.333 \\
\bottomrule
\end{tabular}
  \label{tab:pre-training-size}
  \vspace{-2mm}

\end{table}

\subsection{Results with Other Pre-training Datasets}
Recall that there are 36 tabular datasets for both classification and regression. We randomly select 18 of them as the pre-training datasets and 18 of them as the downstream ones. We exchange the pre-training and downstream datasets in this subsection. 
The results are shown in~\autoref{tab:other_pretrain_reg} and~\autoref{tab:other_pretrain_cls}. The results validate the regression and classification ability of \ours variants.

\begin{table*}[tbp]
  \centering
  \caption{Average RMSE on 18 downstream regression datasets. The best results are in bold, and the second-best results are underlined. The value beside the dataset name denotes the scale of the results. For example, $\times 10$ means all results should be multiplied by $10$. ``OOM'' means out-of-memory.}
  \vspace{-1mm}
  \tabcolsep 1.2pt
    \begin{tabular}{lcccccccccccc}
\addlinespace
\toprule
$\downarrow$ & {\small \ours} & {\small SVM} & {\small $K$NN} & {\small XGBoost} & {\small MLP} & {\small FT-T} & {\small TabR} & {\small SAINT} & {\small TANGOS} & {\small TabNet} & {\small PTaRL} & {\small XTab} \\
\midrule
SPP{\tiny $\times 10$} & .1648 & .2080 & .3240 & \underline{.1550} & .1960 & {\bf .1430} & .1790 & .1970 & .2040 & .4670 & .3610 & .4740 \\
HPP{\tiny $\times 10^{7}$} & .1074 & .1070 & .1130 & \underline{.1060} & .1110 & {\bf .1050} & .1100 & \underline{.1060} & .1130 & .1400 & .1100 & .1580 \\
STO & {\bf .7355} & .9510 & .7950 & .8910 & .9410 & .9880 & .9660 & \underline{.7810} & .8030 & 1.060 & 1.180 & 4.920 \\        
CDA{\tiny $\times 10^{3}$} & \underline{.4956} & .6980 & .5490 & {\bf .4910} & .5910 & .6610 & .5970 & .5380 & .5890 & .6350 & .5560 & .7440 \\
PMS{\tiny $\times 10^{2}$} & \underline{.1534} & .1600 & .1600 & .1630 & {\bf .1520} & .1630 & .1730 & .1720 & .1540 & .1670 & .1630 & .1720 \\
GCB & \underline{.4828} & .4880 & .5040 & {\bf .4440} & .4870 & .4840 & .4840 & .4860 & .4840 & .4880 & .4840 & .5020 \\        
ADO{\tiny $\times 10^{3}$} & {\bf .7374} & .7960 & .7880 & .7750 & .7720 & .7710 & .7730 & .7730 & .7810 & .8200 & \underline{.7670} & 1.300 \\
CAC & {\bf .1322} & .1350 & .1470 & .1430 & \underline{.1340} & .1350 & .1450 & OOM & .1390 & .1590 & .1380 & .1780 \\
AVE{\tiny $\times 10^{3}$} & \underline{3.080} & 6.390 & 4.680 & {\bf .6980} & 9.770 & 9.940 & 9.930 & 9.920 & 9.810 & 3.310 & 3.480 & 9.980 \\
DBT{\tiny $\times 10^{-1}$} & .6617 & 1.110 & .7070 & .7840 & .8530 & .7360 & {\bf .6230} & \underline{.6390} & .6880 & 1.030 & .9600 & 1.450 \\
ASU{\tiny $\times 10^{-1}$} & \underline{.9036} & 1.860 & 1.080 & {\bf .8420} & 1.010 & .9380 & .9040 & 1.010 & 1.070 & 1.110 & 1.290 & 3.020 \\
VNV{\tiny $\times 10^{-1}$} & 1.015 & 1.020 & 1.070 & 1.030 & {\bf .9880} & 1.000 & 1.030 & 1.020 & 1.020 & 1.010 & \underline{.9910} & 1.010 \\
SFA{\tiny $\times 10^{2}$} & .1830 & .1890 & .2010 & {\bf .1670} & .2020 & \underline{.1810} & .1840 & .2000 & .2000 & .2310 & .2190 & .1870 \\
PNH{\tiny $\times 10$} & .3318 & .3380 & .3860 & .3400 & .3320 & {\bf .3270} & \underline{.3300} & .3340 & .3370 & .3330 & .3350 & .3830 \\
PUH{\tiny $\times 10^{-2}$} & .7804 & 2.560 & 2.780 & .9380 & 1.300 & \underline{.7760} & {\bf .6070} & .8810 & .8920 & .9190 & 1.610 & 2.860 \\
SFU{\tiny $\times 10^{-1}$} & \underline{.2032} & .3480 & .2210 & .2540 & .2390 & .2550 & .2310 & {\bf .1890} & .2220 & .2980 & .3080 & .3690 \\
SHP & \underline{.4318} & .4930 & .4840 & {\bf .4290} & .4460 & .4390 & .4320 & .4530 & .4510 & .4340 & .4340 & .4400 \\
SAC{\tiny $\times 10$} & {\bf .3983} & \underline{.4010} & .4170 & .4330 & .4470 & .4490 & .4510 & .4510 & .4510 & .4920 & .4320 & .4380 \\
\midrule
rank &  {\bf 2.444} & 7.833 & 7.667 & \underline{4.500} & 5.833 & 4.833 & 5.667 & 5.588 & 6.333 & 8.556 & 6.556 & 10.278 \\
\bottomrule
\end{tabular}
      \vspace{-4mm}
  \label{tab:reg_main}
\end{table*}

\begin{table*}[tbp]
  \centering
    \caption{Average accuracy on 18 downstream classification datasets. The best results are shown in bold, and the second-best results are underlined.}
  \vspace{-1mm}
  \tabcolsep 2.2pt
\begin{tabular}{lcccccccccccc}
\addlinespace
\toprule
$\uparrow$ & {\small \ours} & {\small SVM} & {\small XGBoost} & {\small MLP} & {\small FT-T} & {\small TabR} & {\small SAINT} & {\small TANGOS} & {\small TabNet} & {\small TabPFN} & {\small XTab} & {\small DEN} \\
\midrule
BAA & {\bf 59.40} & 53.10 & 55.40 & 54.10 & 53.80 & \underline{55.60} & 55.50 & 54.70 & 54.60 & 54.60 & 49.30 & 45.90 \\        
BCC & 85.30 & 86.50 & \underline{86.80} & 85.70 & 86.50 & 85.80 & {\bf 87.10} & 86.40 & 86.00 & 78.90 & 85.00 & 80.10 \\        
BCP & 82.00 & 75.00 & \underline{84.70} & 79.00 & 83.20 & {\bf 87.30} & 83.00 & 76.30 & 82.30 & 74.00 & 77.70 & 75.00 \\        
BCW & {\bf 97.50} & 97.10 & 97.10 & 96.60 & 95.80 & 97.10 & 95.90 & 88.40 & 96.90 & 95.10 & \underline{97.30} & 97.10 \\        
BLO & \underline{78.00} & 74.70 & 76.90 & 77.70 & 77.90 & 77.80 & 75.40 & {\bf 78.20} & 77.20 & 76.40 & 75.70 & 71.20 \\        
BRC & 67.10 & 55.20 & {\bf 69.00} & 62.60 & 65.90 & 66.80 & \underline{67.50} & 67.00 & 67.00 & 66.70 & 65.60 & 63.40 \\        
CDH & 75.20 & 75.10 & 75.10 & 75.20 & \underline{75.30} & 74.20 & 75.00 & 75.20 & 75.00 & 75.00 & {\bf 86.70} & 70.20 \\        
DER & \underline{99.20} & 98.60 & 99.00 & 98.50 & {\bf 99.50} & 98.60 & 98.40 & 98.80 & 98.90 & 89.50 & 80.00 & 97.70 \\        
ECH & \underline{81.20} & 77.80 & 75.60 & 79.80 & 74.10 & 78.00 & 80.20 & 70.40 & 76.50 & 69.90 & {\bf 84.40} & 80.30 \\        
ECS & 67.40 & {\bf 75.10} & 67.80 & 67.20 & 67.80 & 66.20 & 67.70 & 67.80 & 67.10 & 67.20 & 66.30 & \underline{72.90} \\        
FCC & 77.30 & 76.70 & {\bf 79.20} & \underline{78.80} & 78.00 & 77.90 & 77.60 & 77.60 & 78.60 & 77.00 & 72.80 & 71.00 \\        
HEC & 52.30 & {\bf 55.70} & 50.60 & 52.20 & 51.90 & 53.80 & \underline{53.90} & 51.90 & 52.00 & 49.80 & 51.30 & 47.50 \\        
KC2 & 83.10 & 81.90 & 79.60 & {\bf 84.50} & 79.10 & 82.00 & 77.80 & 81.60 & 83.10 & 79.70 & \underline{84.40} & 82.60 \\        
MHR & 68.70 & 67.00 & {\bf 81.90} & 73.20 & 69.60 & 73.80 & \underline{78.10} & 74.10 & 71.40 & 58.80 & 54.30 & 56.80 \\        
SAF & 81.90 & 83.50 & 83.20 & 81.80 & 80.60 & {\bf 84.60} & \underline{83.60} & 82.90 & 80.30 & 79.70 & 68.80 & 78.30 \\        
SEB & 92.80 & 92.80 & 92.70 & \underline{92.90} & \underline{92.90} & 92.80 & {\bf 93.00} & \underline{92.90} & 92.80 & \underline{92.90} & 92.80 & 92.10 \\
TSE & 48.30 & 49.20 & {\bf 51.90} & 49.70 & \underline{51.20} & 49.20 & 50.20 & 50.40 & 49.30 & 49.20 & 46.70 & 31.50 \\        
WAQ & {\bf 91.10} & \underline{89.80} & 89.40 & 89.40 & 88.70 & 88.80 & 89.60 & 89.60 & 89.20 & 88.40 & 88.00 & 88.20 \\        
\midrule
rank & \bf 4.556 & 6.333 & \underline{4.611} & 5.722 & 5.667 & 5.444 & 4.889 & 5.278 & 6.111 & 9.056 & 8.278 & 9.500 \\
\bottomrule
\end{tabular}
      \vspace{-6mm}
  \label{tab:cls_main}
\end{table*}
\

\begin{table}[ht]
  \centering
  \caption{Average RMSE on 18 downstream datasets (the lower, the better). 
  We use {\em another set of 18 datasets as the pre-training datasets} w.r.t.~\autoref{tab:reg_main}.
  The best results are shown in bold, and the second-best results are underlined.
  The value beside the dataset name denotes the scale of the results.}
  \tabcolsep 2pt

    \begin{tabular}{lcccccccccccc}
\addlinespace
\toprule
$\downarrow$ & \ours & SVM & $K$NN & XGBoost & MLP & FT-T & TabR & SAINT & TANGOS & TabNet & PTaRL & XTab \\
\midrule
ARM{\tiny $\times 10$} & .4414 & {\bf .4000} & .4480 & \underline{.4290} & .4300 & .4390 & .4650 & .4360 & .4440 & .4910 & .4420 & .4480 \\
ARP{\tiny $\times 10$} & .3038 & .2930 & .3130 & .2980 & .2930 & {\bf .2860} & .3120 & \underline{.2870} & .2960 & .3450 & .3040 & .3530 \\
CCS{\tiny $\times 10$} & .8217 & \underline{.6990} & .9110 & {\bf .5470} & 1.610 & 1.630 & 1.590 & 1.630 & 1.620 & .7400 & .7070 & 1.720 \\
ARC{\tiny $\times 10^{3}$} & .3733 & .3790 & .3710 & \underline{.3670} & {\bf .3530} & .4010 & .3880 & .3950 & .3680 & .4220 & .3700 & .4790 \\
WTS & .4649 & .4970 & .4680 & .4750 & {\bf .4430} & .4470 & .4710 & .4690 & {\bf .4430} & .4750 & .4500 & .5030 \\
SVM & .7189 & .7610 & .7760 & .7220 & {\bf .6990} & .7200 & .7240 & .8200 & {\bf .6990} & .7970 & .7030 & .8610 \\
ASN{\tiny $\times 10$} & .2321 & .3200 & .2790 & .1830 & .2670 & .1830 & \underline{.1510} & {\bf .1460} & .2040 & .2600 & .3230 & .6420 \\
ITA & .3152 & .3320 & .3420 & {\bf .3110} & \underline{.3130} & .3200 & .3160 & .3160 & .3180 & .3230 & .3200 & .3210 \\        
SCA{\tiny $\times 10^{2}$} & \underline{.2854} & .2960 & .3140 & .2870 & .2860 & .2860 & .2870 & .2870 & .2860 & .2910 & {\bf .2850} & .2890 \\
CDI & .2842 & .3070 & .3290 & .2910 & \underline{.2480} & .2690 & .2680 & .2800 & {\bf .2230} & .3660 & .2810 & .4430 \\        
WQW & .6879 & .6940 & {\bf .6430} & \underline{.6670} & .6880 & .7060 & .6860 & .7080 & .6850 & .7450 & .6980 & .7930 \\        
EUR{\tiny $\times 10^{-2}$} & .8943 & 43.70 & 6.760 & {\bf .1380} & 26.40 & 8.040 & \underline{.7840} & 4.290 & 11.40 & 17.80 & 33.90 & 53.30 \\
SIM & .6974 & .8010 & .7190 & .7570 & .7440 & .7700 & .7700 & {\bf .6660} & \underline{.6830} & .8280 & .7640 & 1.140 \\        
WQA & .7172 & .7290 & \underline{.7020} & {\bf .7010} & .7200 & .7240 & .7080 & .7420 & .7230 & .7510 & .7280 & .7930 \\        
BCR & .8353 & 1.070 & 1.170 & .9280 & .8560 & .8200 & {\bf .5610} & \underline{.7580} & .7790 & 1.180 & 1.080 & 1.730 \\        
CCP{\tiny $\times 10$} & \underline{.3606} & .4110 & .3850 & {\bf .3450} & 1.680 & 1.690 & 1.600 & 1.700 & 1.700 & .4140 & .4250 & 1.690 \\
SHR & .3243 & .3420 & .3480 & .3210 & .3220 & {\bf .3170} & {\bf .3170} & .3260 & .3240 & .3260 & .3190 & .3400 \\
DAT{\tiny $\times 10$} & .3993 & .4410 & .4590 & .4020 & {\bf .3960} & .4200 & .4200 & .4180 & {\bf .3960} & .4020 & .3990 & .4300 \\
\midrule
avg. rank & 4.667 & 8.000 & 7.611 & 3.944 & 4.278 & 5.722 & 5.500 & 6.222 & 4.500 & 9.167 & 6.000 & 11.111 \\
\bottomrule
\end{tabular}

  \label{tab:other_pretrain_reg}
\end{table}

\begin{table*}[t]
  \centering
  \caption{Average accuracy on 18 downstream datasets (the higher, the better). 
  We use {\em another set of 18 datasets as the pre-training datasets} w.r.t.~\autoref{tab:cls_main}.
  The best results are shown in bold, and the second-best results are underlined. ``OOM'' means out-of-memory.}

  \tabcolsep 2.5pt

\begin{tabular}{lcccccccccccc}
\addlinespace
\toprule
$\uparrow$ & \ours & SVM & XGBoost & MLP & FT-T & TabR & SAINT & TANGOS & TabNet & TabPFN & XTab & DEN \\
\midrule
BAN & 86.20 & \underline{88.00} & \underline{88.00} & \underline{88.00} & \underline{88.00} & \underline{88.00} & \underline{88.00} & {\bf 96.20} & \underline{88.00} & \underline{88.00} & \underline{88.00} & 84.80 \\
BAS & 67.30 & {\bf 70.40} & 68.00 & 65.80 & 69.60 & 68.40 & \underline{70.20} & 69.80 & 65.70 & 62.20 & 66.40 & 66.50 \\        
BRE & \underline{97.50} & 96.50 & 97.20 & 96.30 & 95.60 & {\bf 98.20} & 96.10 & 96.20 & 96.80 & 94.10 & 93.00 & 95.70 \\        
COM & \underline{96.50} & \underline{96.50} & {\bf 96.60} & 96.20 & 96.30 & 96.40 & 96.20 & OOM & 96.30 & 96.30 & 96.40 & 96.10 \\ 
DIA & 76.20 & {\bf 77.60} & 75.50 & 74.50 & \underline{77.20} & 75.90 & 76.00 & 75.90 & 73.90 & 69.60 & 70.10 & 67.90 \\        
DRU & 40.20 & {\bf 41.40} & 40.30 & 39.90 & 40.80 & {\bf 41.40} & 39.50 & 40.60 & 40.20 & 40.30 & 40.40 & 40.10 \\
DRY & 92.90 & 92.90 & {\bf 93.10} & {\bf 93.10} & 93.00 & 92.70 & {\bf 93.10} & {\bf 93.10} & {\bf 93.10} & 92.50 & 92.30 & 91.80 \\
HEA & 81.70 & {\bf 84.70} & 77.60 & 82.10 & 80.60 & 80.90 & \underline{83.30} & 77.40 & 81.90 & 68.00 & 79.30 & 78.30 \\        
HEV & \underline{34.50} & 30.00 & {\bf 38.70} & 30.00 & 27.20 & 30.70 & 24.50 & 28.50 & 29.50 & 24.30 & 33.00 & 32.30 \\        
INT & \underline{91.50} & 74.90 & {\bf 93.50} & 79.80 & 79.80 & 80.10 & 91.00 & 80.30 & 79.80 & 61.50 & 56.40 & 63.80 \\        
MAM & 83.40 & 82.90 & {\bf 84.60} & 81.30 & 82.70 & {\bf 84.60} & 83.20 & 84.10 & 82.00 & 81.40 & 80.30 & 78.10 \\
MIC & 98.90 & \underline{99.10} & 96.90 & \underline{99.10} & 98.10 & 98.90 & {\bf 99.50} & \underline{99.10} & 98.70 & 85.90 & 43.10 & 83.00 \\
PAR & \underline{94.50} & 92.30 & 94.40 & 90.30 & 90.80 & {\bf 94.90} & 93.20 & 89.90 & 90.40 & 86.80 & 87.70 & 87.50 \\        
POS & {\bf 84.80} & 55.60 & \underline{83.30} & 78.90 & 59.30 & \underline{83.30} & 70.40 & 71.50 & 75.20 & 43.30 & 74.40 & 73.40 \\
PRI & 50.90 & 48.50 & {\bf 54.60} & 47.10 & 44.10 & 51.20 & \underline{54.40} & 48.00 & 47.30 & 34.60 & 30.00 & 22.20 \\        
SPE & \underline{72.60} & {\bf 73.60} & 71.40 & 68.70 & 70.80 & 71.30 & 71.60 & 69.40 & 68.10 & 59.60 & 70.70 & 68.90 \\        
STA & 70.40 & {\bf 70.50} & 70.40 & 68.00 & 70.40 & {\bf 70.50} & 69.80 & 69.60 & 69.60 & 69.60 & 65.30 & 66.90 \\
WIL & 88.40 & 84.50 & 77.40 & 87.60 & {\bf 89.30} & \underline{89.00} &  OOM  & 87.90 & 88.40 & 82.00 & 62.90 & 71.20 \\
\midrule
avg. rank & 4.056 & 4.056 & 3.889 & 6.667 & 5.889 & 3.500 & 4.941 & 5.471 & 6.333 & 9.611 & 8.667 & 10.000 \\
\bottomrule
\end{tabular}

  \label{tab:other_pretrain_cls}
\end{table*}

\section{Whole Experimental Results}\label{appendix_sec:whole_results}
\autoref{tab:reg_main} and~\autoref{tab:cls_main} include the whole results of~\autoref{fig:full-shot_results}. The standard deviation of~\autoref{tab:reg_main} and~\autoref{tab:cls_main} are listed in~\autoref{tab:reg_main_whole} and~\autoref{tab:cls_main_whole}, respectively. The standard deviation of \ours also comes from the estimation of mutual information when constructing the metric-based meta-representation.~\autoref{tab:fewshot_main_whole} and~\autoref{tab:fewshot_tinybench_whole} are the complete results of few-shot experiments.

\begin{table*}[tbp]
\caption{The whole results of~\autoref{tab:reg_main}: RMSE with standard deviation on 18 downstream datasets (the lower, the better). The best results are shown in bold, and the second-best results are underlined. The value beside the dataset name denotes the scale of the results. {\sc TabPTM}$_{\rm D}$ predicts directly without additional training, while \ours is fine-tuned on the downstream dataset. ``OOM'' means out-of-memory.}
\centering
\begin{adjustbox}{width=1.0\textwidth,center}
\begin{tabular}{lccccccccccccc}
\addlinespace
\toprule
 & {\sc TabPTM}$_{\rm D}$ & \ours & SVM & XGBoost  & MLP & FT-T & TabR & SAINT & TANGOS & TabNet & TabPFN & XTab & DEN \\
\midrule
SPP{\tiny $\times 10$} & .2525 & .1648 & .2079 & .3239 & .1551 & .1963 & {\bf .1430} & .1786 & .1972 & .2038 & .4671 & .3615 & .4736 \\
 & \quad{\scriptsize$\pm$  0.00} & \quad{\scriptsize$\pm$  .001} & \quad{\scriptsize$\pm$  0.00} & \quad{\scriptsize$\pm$  0.00} & \quad{\scriptsize$\pm$  .007} & \quad{\scriptsize$\pm$  .013} & \quad{\scriptsize$\pm$  .009} & \quad{\scriptsize$\pm$  .005} & \quad{\scriptsize$\pm$  .012} & \quad{\scriptsize$\pm$  .005} & \quad{\scriptsize$\pm$  .068} & \quad{\scriptsize$\pm$  .135} & \quad{\scriptsize$\pm$  .018} \\
HPP{\tiny $\times 10^{7}$} & .1210 & .1074 & .1075 & .1132 & .1064 & .1107 & {\bf .1054} & .1103 & .1064 & .1134 & .1400 & .1104 & .1585 \\
 & \quad{\scriptsize$\pm$  .001} & \quad{\scriptsize$\pm$  .002} & \quad{\scriptsize$\pm$  0.00} & \quad{\scriptsize$\pm$  0.00} & \quad{\scriptsize$\pm$  .004} & \quad{\scriptsize$\pm$  .002} & \quad{\scriptsize$\pm$  .001} & \quad{\scriptsize$\pm$  .002} & \quad{\scriptsize$\pm$  .001} & \quad{\scriptsize$\pm$  .002} & \quad{\scriptsize$\pm$  .016} & \quad{\scriptsize$\pm$  0.00} & \quad{\scriptsize$\pm$  .012} \\
STO & 2.974 & {\bf .7355} & .9513 & .7948 & .8907 & .9408 & .9883 & .9659 & .7809 & .8032 & 1.057 & 1.184 & 4.922 \\
 & \quad{\scriptsize$\pm$  .001} & \quad{\scriptsize$\pm$  .002} & \quad{\scriptsize$\pm$  0.00} & \quad{\scriptsize$\pm$  0.00} & \quad{\scriptsize$\pm$  .047} & \quad{\scriptsize$\pm$  .014} & \quad{\scriptsize$\pm$  .020} & \quad{\scriptsize$\pm$  .161} & \quad{\scriptsize$\pm$  .015} & \quad{\scriptsize$\pm$  .014} & \quad{\scriptsize$\pm$  .063} & \quad{\scriptsize$\pm$  .015} & \quad{\scriptsize$\pm$  .352} \\
CDA{\tiny $\times 10^{3}$} & .5567 & .4956 & .6976 & .5489 & {\bf .4914} & .5908 & .6613 & .5966 & .5381 & .5893 & .6351 & .5561 & .7441 \\
 & \quad{\scriptsize$\pm$  .014} & \quad{\scriptsize$\pm$  .015} & \quad{\scriptsize$\pm$  0.00} & \quad{\scriptsize$\pm$  0.00} & \quad{\scriptsize$\pm$  .053} & \quad{\scriptsize$\pm$  .005} & \quad{\scriptsize$\pm$  .125} & \quad{\scriptsize$\pm$  .058} & \quad{\scriptsize$\pm$  .076} & \quad{\scriptsize$\pm$  .022} & \quad{\scriptsize$\pm$  .051} & \quad{\scriptsize$\pm$  .019} & \quad{\scriptsize$\pm$  .051} \\
PMS{\tiny $\times 10^{2}$} & .1601 & .1534 & .1603 & .1603 & .1629 & {\bf .1524} & .1633 & .1727 & .1720 & .1543 & .1673 & .1630 & .1717 \\
 & \quad{\scriptsize$\pm$  .001} & \quad{\scriptsize$\pm$  .001} & \quad{\scriptsize$\pm$  0.00} & \quad{\scriptsize$\pm$  0.00} & \quad{\scriptsize$\pm$  .002} & \quad{\scriptsize$\pm$  .001} & \quad{\scriptsize$\pm$  .007} & \quad{\scriptsize$\pm$  .006} & \quad{\scriptsize$\pm$  .009} & \quad{\scriptsize$\pm$  .004} & \quad{\scriptsize$\pm$  .005} & \quad{\scriptsize$\pm$  .002} & \quad{\scriptsize$\pm$  .002} \\
GCB & .5087 & .4828 & .4884 & .5038 & {\bf .4437} & .4866 & .4838 & .4844 & .4855 & .4843 & .4876 & .4841 & .5015 \\
 & \quad{\scriptsize$\pm$  .002} & \quad{\scriptsize$\pm$  .001} & \quad{\scriptsize$\pm$  0.00} & \quad{\scriptsize$\pm$  0.00} & \quad{\scriptsize$\pm$  .016} & \quad{\scriptsize$\pm$  .001} & \quad{\scriptsize$\pm$  0.00} & \quad{\scriptsize$\pm$  .001} & \quad{\scriptsize$\pm$  .001} & \quad{\scriptsize$\pm$  .001} & \quad{\scriptsize$\pm$  .007} & \quad{\scriptsize$\pm$  0.00} & \quad{\scriptsize$\pm$  .016} \\
ADO{\tiny $\times 10^{3}$} & {\bf .7184} & .7374 & .7956 & .7880 & .7752 & .7725 & .7710 & .7726 & .7731 & .7810 & .8197 & .7668 & 1.295 \\
 & \quad{\scriptsize$\pm$  .002} & \quad{\scriptsize$\pm$  .002} & \quad{\scriptsize$\pm$  0.00} & \quad{\scriptsize$\pm$  0.00} & \quad{\scriptsize$\pm$  .010} & \quad{\scriptsize$\pm$  .004} & \quad{\scriptsize$\pm$  .007} & \quad{\scriptsize$\pm$  .001} & \quad{\scriptsize$\pm$  .003} & \quad{\scriptsize$\pm$  .008} & \quad{\scriptsize$\pm$  .025} & \quad{\scriptsize$\pm$  .006} & \quad{\scriptsize$\pm$  .294} \\
CAC & .1341 & {\bf .1322} & .1348 & .1467 & .1433 & .1337 & .1350 & .1452 & OOM & .1393 & .1591 & .1384 & .1781 \\
 & \quad{\scriptsize$\pm$  0.00} & \quad{\scriptsize$\pm$  .001} & \quad{\scriptsize$\pm$  0.00} & \quad{\scriptsize$\pm$  0.00} & \quad{\scriptsize$\pm$  .002} & \quad{\scriptsize$\pm$  .001} & \quad{\scriptsize$\pm$  .001} & \quad{\scriptsize$\pm$  .006} &   & \quad{\scriptsize$\pm$  .002} & \quad{\scriptsize$\pm$  .010} & \quad{\scriptsize$\pm$  .001} & \quad{\scriptsize$\pm$  .019} \\
AVE{\tiny $\times 10^{3}$} & 3.816 & 3.080 & 6.394 & 4.678 & {\bf .6983} & 9.765 & 9.942 & 9.933 & 9.920 & 9.811 & 3.314 & 3.484 & 9.982 \\
 & \quad{\scriptsize$\pm$  .007} & \quad{\scriptsize$\pm$  .007} & \quad{\scriptsize$\pm$  0.00} & \quad{\scriptsize$\pm$  0.00} & \quad{\scriptsize$\pm$  .124} & \quad{\scriptsize$\pm$  .027} & \quad{\scriptsize$\pm$  .146} & \quad{\scriptsize$\pm$  .594} & \quad{\scriptsize$\pm$  .070} & \quad{\scriptsize$\pm$  .036} & \quad{\scriptsize$\pm$  .236} & \quad{\scriptsize$\pm$  .134} & \quad{\scriptsize$\pm$  .184} \\
DBT{\tiny $\times 10^{-1}$} & .9609 & .6617 & 1.107 & .7065 & .7841 & .8534 & .7356 & {\bf .6233} & .6385 & .6884 & 1.034 & .9602 & 1.446 \\
 & \quad{\scriptsize$\pm$  34.5} & \quad{\scriptsize$\pm$  34.6} & \quad{\scriptsize$\pm$  0.00} & \quad{\scriptsize$\pm$  0.00} & \quad{\scriptsize$\pm$  .029} & \quad{\scriptsize$\pm$  .005} & \quad{\scriptsize$\pm$  .018} & \quad{\scriptsize$\pm$  .044} & \quad{\scriptsize$\pm$  .017} & \quad{\scriptsize$\pm$  .016} & \quad{\scriptsize$\pm$  .037} & \quad{\scriptsize$\pm$  .010} & \quad{\scriptsize$\pm$  .035} \\
ASU{\tiny $\times 10^{-1}$} & 2.452 & .9036 & 1.862 & 1.078 & {\bf .8418} & 1.013 & .9379 & .9037 & 1.010 & 1.071 & 1.109 & 1.293 & 3.016 \\
 & \quad{\scriptsize$\pm$  .012} & \quad{\scriptsize$\pm$  .032} & \quad{\scriptsize$\pm$  0.00} & \quad{\scriptsize$\pm$  0.00} & \quad{\scriptsize$\pm$  .038} & \quad{\scriptsize$\pm$  .025} & \quad{\scriptsize$\pm$  .015} & \quad{\scriptsize$\pm$  .029} & \quad{\scriptsize$\pm$  .042} & \quad{\scriptsize$\pm$  .046} & \quad{\scriptsize$\pm$  .085} & \quad{\scriptsize$\pm$  .040} & \quad{\scriptsize$\pm$  .905} \\
VNV{\tiny $\times 10^{-1}$} & 1.071 & 1.015 & 1.020 & 1.071 & 1.032 & {\bf .9879} & 1.004 & 1.031 & 1.020 & 1.019 & 1.007 & .9910 & 1.007 \\
 & \quad{\scriptsize$\pm$  .016} & \quad{\scriptsize$\pm$  .016} & \quad{\scriptsize$\pm$  0.00} & \quad{\scriptsize$\pm$  0.00} & \quad{\scriptsize$\pm$  .012} & \quad{\scriptsize$\pm$  .003} & \quad{\scriptsize$\pm$  .010} & \quad{\scriptsize$\pm$  .032} & \quad{\scriptsize$\pm$  .008} & \quad{\scriptsize$\pm$  .016} & \quad{\scriptsize$\pm$  .011} & \quad{\scriptsize$\pm$  .002} & \quad{\scriptsize$\pm$  .004} \\
SFA{\tiny $\times 10^{2}$} & .2505 & .1830 & .1888 & .2012 & {\bf .1672} & .2024 & .1806 & .1843 & .2003 & .2001 & .2312 & .2192 & .1872 \\
 & \quad{\scriptsize$\pm$  .013} & \quad{\scriptsize$\pm$  .013} & \quad{\scriptsize$\pm$  0.00} & \quad{\scriptsize$\pm$  0.00} & \quad{\scriptsize$\pm$  .016} & \quad{\scriptsize$\pm$  .005} & \quad{\scriptsize$\pm$  0.00} & \quad{\scriptsize$\pm$  .005} & \quad{\scriptsize$\pm$  .015} & \quad{\scriptsize$\pm$  .002} & \quad{\scriptsize$\pm$  .027} & \quad{\scriptsize$\pm$  .002} & \quad{\scriptsize$\pm$  .003} \\
PNH{\tiny $\times 10$} & .3893 & .3318 & .3375 & .3862 & .3404 & .3324 & {\bf .3265} & .3296 & .3344 & .3366 & .3326 & .3346 & .3832 \\
 & \quad{\scriptsize$\pm$  0.00} & \quad{\scriptsize$\pm$  0.00} & \quad{\scriptsize$\pm$  0.00} & \quad{\scriptsize$\pm$  0.00} & \quad{\scriptsize$\pm$  .002} & \quad{\scriptsize$\pm$  .001} & \quad{\scriptsize$\pm$  0.00} & \quad{\scriptsize$\pm$  0.00} & \quad{\scriptsize$\pm$  .003} & \quad{\scriptsize$\pm$  .001} & \quad{\scriptsize$\pm$  .004} & \quad{\scriptsize$\pm$  .001} & \quad{\scriptsize$\pm$  .009} \\
PUH{\tiny $\times 10^{-2}$} & 1.515 & .7804 & 2.557 & 2.776 & .9376 & 1.299 & .7762 & {\bf .6072} & .8815 & .8925 & .9189 & 1.612 & 2.855 \\
 & \quad{\scriptsize$\pm$  266.} & \quad{\scriptsize$\pm$  266.} & \quad{\scriptsize$\pm$  0.00} & \quad{\scriptsize$\pm$  0.00} & \quad{\scriptsize$\pm$  .070} & \quad{\scriptsize$\pm$  .658} & \quad{\scriptsize$\pm$  .003} & \quad{\scriptsize$\pm$  .008} & \quad{\scriptsize$\pm$  .020} & \quad{\scriptsize$\pm$  .012} & \quad{\scriptsize$\pm$  .068} & \quad{\scriptsize$\pm$  .031} & \quad{\scriptsize$\pm$  .100} \\
SFU{\tiny $\times 10^{-1}$} & .3174 & .2032 & .3479 & .2206 & .2545 & .2395 & .2555 & .2307 & {\bf .1891} & .2218 & .2976 & .3084 & .3686 \\
 & \quad{\scriptsize$\pm$  21.5} & \quad{\scriptsize$\pm$  21.5} & \quad{\scriptsize$\pm$  0.00} & \quad{\scriptsize$\pm$  0.00} & \quad{\scriptsize$\pm$  .021} & \quad{\scriptsize$\pm$  .007} & \quad{\scriptsize$\pm$  .065} & \quad{\scriptsize$\pm$  .009} & \quad{\scriptsize$\pm$  .032} & \quad{\scriptsize$\pm$  .033} & \quad{\scriptsize$\pm$  .019} & \quad{\scriptsize$\pm$  .003} & \quad{\scriptsize$\pm$  .006} \\
SHP & .4391 & .4318 & .4930 & .4841 & {\bf .4292} & .4464 & .4393 & .4318 & OOM  & .4514 & .4341 & .4341 & .4402 \\
 & \quad{\scriptsize$\pm$  .007} & \quad{\scriptsize$\pm$  .001} & \quad{\scriptsize$\pm$  0.00} & \quad{\scriptsize$\pm$  0.00} & \quad{\scriptsize$\pm$  .001} & \quad{\scriptsize$\pm$  .001} & \quad{\scriptsize$\pm$  .014} & \quad{\scriptsize$\pm$  .005} &  & \quad{\scriptsize$\pm$  .007} & \quad{\scriptsize$\pm$  .003} & \quad{\scriptsize$\pm$  .001} & \quad{\scriptsize$\pm$  .011} \\
SAC{\tiny $\times 10$} & .4007 & {\bf .3983} & .4015 & .4169 & .4329 & .4472 & .4494 & .4511 & .4508 & .4508 & .4923 & .4324 & .4379 \\
 & \quad{\scriptsize$\pm$  .010} & \quad{\scriptsize$\pm$  .009} & \quad{\scriptsize$\pm$  0.00} & \quad{\scriptsize$\pm$  0.00} & \quad{\scriptsize$\pm$  .014} & \quad{\scriptsize$\pm$  .002} & \quad{\scriptsize$\pm$  .002} & \quad{\scriptsize$\pm$  .005} & \quad{\scriptsize$\pm$  .001} & \quad{\scriptsize$\pm$  .004} & \quad{\scriptsize$\pm$  .028} & \quad{\scriptsize$\pm$  .016} & \quad{\scriptsize$\pm$  .009} \\
\midrule
avg. rank & 8.500 & {\bf 2.444} & 8.611 & 8.056 & \underline{ 4.778}& 6.167 & 5.389 & 6.278 & 5.875 & 6.889 & 9.000 & 7.111 & 11.111 \\
\bottomrule
\end{tabular}
\end{adjustbox}
\label{tab:reg_main_whole}
\end{table*}

\begin{table*}[tbp]
\caption{The whole results of~\autoref{tab:cls_main}: Average accuracy with standard deviation on 18 downstream datasets. The best results are shown in bold, and the second-best results are underlined. {\sc TabPTM}$_{\rm D}$ predicts directly without additional training, while \ours is fine-tuned on the downstream dataset.}
\centering
\begin{adjustbox}{width=1.0\textwidth,center}
\begin{tabular}{lccccccccccccc}
\addlinespace
\toprule
 & {\sc TabPTM}$_{\rm D}$ & \ours & SVM & XGBoost  & MLP & FT-T & TabR & SAINT & TANGOS & TabNet & TabPFN & XTab & DEN \\
\midrule
BCC & 85.90 & 85.30 & 86.50 & 86.80 & 85.70 & 86.50 & 85.80 & {\bf 87.10} & 86.40 & 86.00 & 78.90 & 85.00 & 80.10 \\
 & \quad{\scriptsize$\pm$  0.30} & \quad{\scriptsize$\pm$  0.80} & \quad{\scriptsize$\pm$  0.01} & \quad{\scriptsize$\pm$  0.28} & \quad{\scriptsize$\pm$  0.01} & \quad{\scriptsize$\pm$  0.17} & \quad{\scriptsize$\pm$  0.26} & \quad{\scriptsize$\pm$  0.34} & \quad{\scriptsize$\pm$  0.30} & \quad{\scriptsize$\pm$  1.50} & \quad{\scriptsize$\pm$  1.60} & \quad{\scriptsize$\pm$  0.60} & \quad{\scriptsize$\pm$  2.30} \\
CDH & 72.60 & 75.20 & 75.10 & 75.10 & 75.20 & 75.30 & 74.20 & 75.00 & 75.20 & 75.00 & 75.00 & {\bf 86.70} & 70.20 \\
 & \quad{\scriptsize$\pm$  0.80} & \quad{\scriptsize$\pm$  0.00} & \quad{\scriptsize$\pm$  0.01} & \quad{\scriptsize$\pm$  0.09} & \quad{\scriptsize$\pm$  0.10} & \quad{\scriptsize$\pm$  0.12} & \quad{\scriptsize$\pm$  0.14} & \quad{\scriptsize$\pm$  0.14} & \quad{\scriptsize$\pm$  0.13} & \quad{\scriptsize$\pm$  0.13} & \quad{\scriptsize$\pm$  0.26} & \quad{\scriptsize$\pm$  0.61} & \quad{\scriptsize$\pm$  1.20} \\
ECS & 67.40 & 67.40 & {\bf 75.10} & 67.80 & 67.20 & 67.80 & 66.20 & 67.70 & 67.80 & 67.10 & 67.20 & 66.30 & 72.90 \\
 & \quad{\scriptsize$\pm$  1.30} & \quad{\scriptsize$\pm$  0.80} & \quad{\scriptsize$\pm$  0.01} & \quad{\scriptsize$\pm$  0.29} & \quad{\scriptsize$\pm$  0.23} & \quad{\scriptsize$\pm$  0.43} & \quad{\scriptsize$\pm$  0.52} & \quad{\scriptsize$\pm$  0.34} & \quad{\scriptsize$\pm$  0.45} & \quad{\scriptsize$\pm$  0.46} & \quad{\scriptsize$\pm$  0.70} & \quad{\scriptsize$\pm$  1.30} & \quad{\scriptsize$\pm$  0.68} \\
FCC & 76.30 & 77.30 & 76.70 & {\bf 79.20} & 78.80 & 78.00 & 77.90 & 77.60 & 77.60 & 78.60 & 77.00 & 72.80 & 71.00 \\
 & \quad{\scriptsize$\pm$  0.75} & \quad{\scriptsize$\pm$  1.40} & \quad{\scriptsize$\pm$  0.01} & \quad{\scriptsize$\pm$  1.20} & \quad{\scriptsize$\pm$  2.00} & \quad{\scriptsize$\pm$  1.70} & \quad{\scriptsize$\pm$  0.74} & \quad{\scriptsize$\pm$  1.60} & \quad{\scriptsize$\pm$  2.10} & \quad{\scriptsize$\pm$  1.50} & \quad{\scriptsize$\pm$  2.50} & \quad{\scriptsize$\pm$  1.80} & \quad{\scriptsize$\pm$  1.00} \\
BAA & 48.40 & {\bf 59.40} & 53.10 & 55.40 & 54.10 & 53.80 & 55.60 & 55.50 & 54.70 & 54.60 & 54.60 & 49.30 & 45.90 \\
 & \quad{\scriptsize$\pm$  1.20} & \quad{\scriptsize$\pm$  1.40} & \quad{\scriptsize$\pm$  0.01} & \quad{\scriptsize$\pm$  0.73} & \quad{\scriptsize$\pm$  1.40} & \quad{\scriptsize$\pm$  1.20} & \quad{\scriptsize$\pm$  0.01} & \quad{\scriptsize$\pm$  1.20} & \quad{\scriptsize$\pm$  1.50} & \quad{\scriptsize$\pm$  0.92} & \quad{\scriptsize$\pm$  1.70} & \quad{\scriptsize$\pm$  1.30} & \quad{\scriptsize$\pm$  2.00} \\
KC2 & 84.10 & 83.10 & 81.90 & 79.60 & {\bf 84.50} & 79.10 & 82.00 & 77.80 & 81.60 & 83.10 & 79.70 & 84.40 & 82.60 \\
 & \quad{\scriptsize$\pm$  1.50} & \quad{\scriptsize$\pm$  1.60} & \quad{\scriptsize$\pm$  0.01} & \quad{\scriptsize$\pm$  1.40} & \quad{\scriptsize$\pm$  1.70} & \quad{\scriptsize$\pm$  2.40} & \quad{\scriptsize$\pm$  0.65} & \quad{\scriptsize$\pm$  2.20} & \quad{\scriptsize$\pm$  1.00} & \quad{\scriptsize$\pm$  2.30} & \quad{\scriptsize$\pm$  2.30} & \quad{\scriptsize$\pm$  2.70} & \quad{\scriptsize$\pm$  2.60} \\
MHR & 64.40 & 68.70 & 67.00 & {\bf 81.90} & 73.20 & 69.60 & 73.80 & 78.10 & 74.10 & 71.40 & 58.80 & 54.30 & 56.80 \\
 & \quad{\scriptsize$\pm$  1.30} & \quad{\scriptsize$\pm$  1.10} & \quad{\scriptsize$\pm$  0.01} & \quad{\scriptsize$\pm$  0.63} & \quad{\scriptsize$\pm$  1.20} & \quad{\scriptsize$\pm$  1.40} & \quad{\scriptsize$\pm$  0.72} & \quad{\scriptsize$\pm$  2.20} & \quad{\scriptsize$\pm$  1.80} & \quad{\scriptsize$\pm$  1.70} & \quad{\scriptsize$\pm$  5.00} & \quad{\scriptsize$\pm$  1.60} & \quad{\scriptsize$\pm$  1.50} \\
SEB & 92.60 & 92.80 & 92.80 & 92.70 & 92.90 & 92.90 & 92.80 & {\bf 93.00} & 92.90 & 92.80 & 92.90 & 92.80 & 92.10 \\
 & \quad{\scriptsize$\pm$  1.80} & \quad{\scriptsize$\pm$  2.80} & \quad{\scriptsize$\pm$  0.01} & \quad{\scriptsize$\pm$  0.01} & \quad{\scriptsize$\pm$  1.00} & \quad{\scriptsize$\pm$  1.20} & \quad{\scriptsize$\pm$  0.05} & \quad{\scriptsize$\pm$  0.21} & \quad{\scriptsize$\pm$  0.17} & \quad{\scriptsize$\pm$  0.13} & \quad{\scriptsize$\pm$  2.90} & \quad{\scriptsize$\pm$  1.50} & \quad{\scriptsize$\pm$  1.00} \\
SAF & 81.40 & 81.90 & 83.50 & 83.20 & 81.80 & 80.60 & {\bf 84.60} & 83.60 & 82.90 & 80.30 & 79.70 & 68.80 & 78.30 \\
 & \quad{\scriptsize$\pm$  0.98} & \quad{\scriptsize$\pm$  1.30} & \quad{\scriptsize$\pm$  0.01} & \quad{\scriptsize$\pm$  0.98} & \quad{\scriptsize$\pm$  1.10} & \quad{\scriptsize$\pm$  1.50} & \quad{\scriptsize$\pm$  0.73} & \quad{\scriptsize$\pm$  1.10} & \quad{\scriptsize$\pm$  0.96} & \quad{\scriptsize$\pm$  1.60} & \quad{\scriptsize$\pm$  1.30} & \quad{\scriptsize$\pm$  2.00} & \quad{\scriptsize$\pm$  0.03} \\
TSE & 44.50 & 48.30 & 49.20 & {\bf 51.90} & 49.70 & 51.20 & 49.20 & 50.20 & 50.40 & 49.30 & 49.20 & 46.70 & 31.50 \\
 & \quad{\scriptsize$\pm$  0.60} & \quad{\scriptsize$\pm$  0.80} & \quad{\scriptsize$\pm$  0.01} & \quad{\scriptsize$\pm$  0.15} & \quad{\scriptsize$\pm$  0.80} & \quad{\scriptsize$\pm$  0.79} & \quad{\scriptsize$\pm$  0.55} & \quad{\scriptsize$\pm$  0.58} & \quad{\scriptsize$\pm$  0.64} & \quad{\scriptsize$\pm$  0.79} & \quad{\scriptsize$\pm$  1.10} & \quad{\scriptsize$\pm$  0.51} & \quad{\scriptsize$\pm$  1.40} \\
WAQ & 88.90 & {\bf 91.10} & 89.80 & 89.40 & 89.40 & 88.70 & 88.80 & 89.60 & 89.60 & 89.20 & 88.40 & 88.00 & 88.20 \\
 & \quad{\scriptsize$\pm$  0.70} & \quad{\scriptsize$\pm$  1.10} & \quad{\scriptsize$\pm$  0.01} & \quad{\scriptsize$\pm$  0.20} & \quad{\scriptsize$\pm$  0.20} & \quad{\scriptsize$\pm$  0.41} & \quad{\scriptsize$\pm$  0.18} & \quad{\scriptsize$\pm$  0.19} & \quad{\scriptsize$\pm$  0.37} & \quad{\scriptsize$\pm$  0.27} & \quad{\scriptsize$\pm$  0.64} & \quad{\scriptsize$\pm$  2.20} & \quad{\scriptsize$\pm$  2.10} \\
BLO & 73.60 & 78.00 & 74.70 & 76.90 & 77.70 & 77.90 & 77.80 & 75.40 & {\bf 78.20} & 77.20 & 76.40 & 75.70 & 71.20 \\
 & \quad{\scriptsize$\pm$  0.45} & \quad{\scriptsize$\pm$  1.90} & \quad{\scriptsize$\pm$  0.01} & \quad{\scriptsize$\pm$  1.90} & \quad{\scriptsize$\pm$  1.90} & \quad{\scriptsize$\pm$  0.41} & \quad{\scriptsize$\pm$  0.38} & \quad{\scriptsize$\pm$  2.50} & \quad{\scriptsize$\pm$  1.00} & \quad{\scriptsize$\pm$  0.13} & \quad{\scriptsize$\pm$  2.60} & \quad{\scriptsize$\pm$  1.20} & \quad{\scriptsize$\pm$  0.20} \\
BRC & 66.00 & 67.10 & 55.20 & {\bf 69.00} & 62.60 & 65.90 & 66.80 & 67.50 & 67.00 & 67.00 & 66.70 & 65.60 & 63.40 \\
 & \quad{\scriptsize$\pm$  0.85} & \quad{\scriptsize$\pm$  1.00} & \quad{\scriptsize$\pm$  0.01} & \quad{\scriptsize$\pm$  0.42} & \quad{\scriptsize$\pm$  1.80} & \quad{\scriptsize$\pm$  2.90} & \quad{\scriptsize$\pm$  1.60} & \quad{\scriptsize$\pm$  1.40} & \quad{\scriptsize$\pm$  1.70} & \quad{\scriptsize$\pm$  1.70} & \quad{\scriptsize$\pm$  1.60} & \quad{\scriptsize$\pm$  0.34} & \quad{\scriptsize$\pm$  1.70} \\
BCW & 96.60 & {\bf 97.50} & 97.10 & 97.10 & 96.60 & 95.80 & 97.10 & 95.90 & 88.40 & 96.90 & 95.10 & 97.30 & 97.10 \\
 & \quad{\scriptsize$\pm$  1.30} & \quad{\scriptsize$\pm$  1.70} & \quad{\scriptsize$\pm$  0.01} & \quad{\scriptsize$\pm$  1.30} & \quad{\scriptsize$\pm$  0.20} & \quad{\scriptsize$\pm$  1.10} & \quad{\scriptsize$\pm$  1.30} & \quad{\scriptsize$\pm$  0.46} & \quad{\scriptsize$\pm$  2.30} & \quad{\scriptsize$\pm$  0.53} & \quad{\scriptsize$\pm$  1.80} & \quad{\scriptsize$\pm$  2.40} & \quad{\scriptsize$\pm$  1.70} \\
BCP & 82.00 & 82.00 & 75.00 & 84.70 & 79.00 & 83.20 & {\bf 87.30} & 83.00 & 76.30 & 82.30 & 74.00 & 77.70 & 75.00 \\
 & \quad{\scriptsize$\pm$  1.70} & \quad{\scriptsize$\pm$  2.00} & \quad{\scriptsize$\pm$  0.01} & \quad{\scriptsize$\pm$  2.70} & \quad{\scriptsize$\pm$  1.40} & \quad{\scriptsize$\pm$  2.00} & \quad{\scriptsize$\pm$  1.00} & \quad{\scriptsize$\pm$  1.70} & \quad{\scriptsize$\pm$  0.38} & \quad{\scriptsize$\pm$  0.12} & \quad{\scriptsize$\pm$  1.20} & \quad{\scriptsize$\pm$  1.80} & \quad{\scriptsize$\pm$  2.00} \\
DER & {\bf 99.80} & 99.20 & 98.60 & 99.00 & 98.50 & 99.50 & 98.60 & 98.40 & 98.80 & 98.90 & 89.50 & 80.00 & 97.70 \\
 & \quad{\scriptsize$\pm$  0.30} & \quad{\scriptsize$\pm$  0.50} & \quad{\scriptsize$\pm$  0.01} & \quad{\scriptsize$\pm$  0.80} & \quad{\scriptsize$\pm$  0.30} & \quad{\scriptsize$\pm$  0.39} & \quad{\scriptsize$\pm$  0.30} & \quad{\scriptsize$\pm$  0.90} & \quad{\scriptsize$\pm$  0.10} & \quad{\scriptsize$\pm$  0.34} & \quad{\scriptsize$\pm$  0.00} & \quad{\scriptsize$\pm$  0.50} & \quad{\scriptsize$\pm$  0.50} \\
ECH & {\bf 84.70} & 81.20 & 77.80 & 75.60 & 79.80 & 74.10 & 78.00 & 80.20 & 70.40 & 76.50 & 69.90 & 84.40 & 80.30 \\
 & \quad{\scriptsize$\pm$  1.90} & \quad{\scriptsize$\pm$  1.80} & \quad{\scriptsize$\pm$  0.01} & \quad{\scriptsize$\pm$  1.60} & \quad{\scriptsize$\pm$  0.33} & \quad{\scriptsize$\pm$  1.60} & \quad{\scriptsize$\pm$  2.50} & \quad{\scriptsize$\pm$  1.60} & \quad{\scriptsize$\pm$  1.90} & \quad{\scriptsize$\pm$  1.60} & \quad{\scriptsize$\pm$  1.60} & \quad{\scriptsize$\pm$  1.50} & \quad{\scriptsize$\pm$  1.80} \\
HEC & 52.70 & 52.30 & {\bf 55.70} & 50.60 & 52.20 & 51.90 & 53.80 & 53.90 & 51.90 & 52.00 & 49.80 & 51.30 & 47.50 \\
 & \quad{\scriptsize$\pm$  1.00} & \quad{\scriptsize$\pm$  2.50} & \quad{\scriptsize$\pm$  0.01} & \quad{\scriptsize$\pm$  2.30} & \quad{\scriptsize$\pm$  2.60} & \quad{\scriptsize$\pm$  1.80} & \quad{\scriptsize$\pm$  1.90} & \quad{\scriptsize$\pm$  2.20} & \quad{\scriptsize$\pm$  1.60} & \quad{\scriptsize$\pm$  1.20} & \quad{\scriptsize$\pm$  1.60} & \quad{\scriptsize$\pm$  1.10} & \quad{\scriptsize$\pm$  2.90} \\
\midrule
avg. rank & 7.944 & \bf 4.833 & 6.833 & \underline{5.056} & 6.389 & 6.167 & 6.056 & 5.167 & 5.778 & 6.722 & 9.944 & 8.889 & 10.556 \\
\bottomrule
\end{tabular}
\end{adjustbox}
\label{tab:cls_main_whole}
\end{table*}

\begin{longtable}{lccccccccc}
\caption{The complete results for the few-shot tasks on the 36 main datasets (\autoref{fig:few-shot_results}). The absence of certain shot configurations in some datasets is attributed to the lack of sufficient training data for specific classes. ``NCE'' indicates that the number of classes exceeds the predefined class limit supported by TabPFN.} \\
\toprule
Dataset & Shot & TabPFN & $K$NN & LogReg & XGB & MLP & FT-T & MNCA & \ours \\
\toprule
\endfirsthead

\toprule
Dataset & Shot & TabPFN & $K$NN & LogReg & XGB & MLP & FT-T & MNCA & \ours \\
\toprule
\endhead

\toprule
\endfoot
heart-hungarian & 4 & .7085 & \underline{.7220} & \underline{.7220} & .5831 & .7119 & .6712 & .7017 & {\bf .7627} \\
heart-hungarian & 8 & .6644 & \underline{.7322} & .6339 & .6373 & .6746 & .6915 & .6542 & {\bf .7492} \\
heart-hungarian & 16 & .8102 & {\bf .8305} & .7932 & .6746 & .7559 & .7559 & .7729 & \underline{.8237} \\
heart-hungarian & 32 & .7966 & {\bf .8644} & .8136 & .7017 & .7831 & .7661 & .7729 & \underline{.8441} \\
heart-hungarian & 64 & .8339 & {\bf .8949} & .8169 & .6881 & .8169 & .7797 & .7661 & \underline{.8407} \\
heart-va & 4 & .1450 & .1700 & {\bf .2200} & \underline{.2100} & .2050 & \underline{.2100} & .2000 & .1950 \\
breast-cancer-wisc-diag & 4 & \underline{.9193} & .8579 & {\bf .9211} & .8404 & .9140 & .9053 & .8947 & .9018 \\
breast-cancer-wisc-diag & 8 & {\bf .9333} & .9158 & .9211 & .8702 & \underline{.9246} & .9193 & .9123 & .9123 \\
breast-cancer-wisc-diag & 16 & .9158 & .8895 & .9123 & .8825 & .9053 & {\bf .9263} & .9105 & \underline{.9158} \\
breast-cancer-wisc-diag & 32 & {\bf .9526} & .9105 & {\bf .9526} & .9368 & .9421 & .9333 & .9263 & .9158 \\
breast-cancer-wisc-diag & 64 & {\bf .9737} & .8982 & {\bf .9737} & .9491 & .9456 & .9439 & .9263 & .9140 \\
mammographic & 4 & .7959 & \underline{.8114} & .8083 & .7420 & .7979 & .8052 & .8062 & {\bf .8394} \\
mammographic & 8 & .8228 & .8207 & .8197 & .8093 & .8249 & \underline{.8363} & .8228 & {\bf .8415} \\
mammographic & 16 & .7917 & .7627 & \underline{.8135} & .7492 & .7793 & .7865 & .7813 & {\bf .8466} \\
mammographic & 32 & \underline{.8228} & .7876 & .8124 & .8010 & .8052 & .7969 & .7907 & {\bf .8352} \\
mammographic & 64 & {\bf .8363} & .7824 & .8321 & .8021 & .8311 & \underline{.8342} & .8114 & .8311 \\
parkinsons & 4 & .7333 & .6718 & .7128 & .7333 & {\bf .7385} & \underline{.7333} & .7026 & .6821 \\
parkinsons & 8 & \underline{.7590} & .7026 & .7538 & .6821 & .7333 & {\bf .7641} & .6872 & .7179 \\
parkinsons & 16 & .7487 & .7077 & .7077 & .7641 & .7641 & \underline{.7692} & {\bf .7897} & .6769 \\
primary-tumor & 4 & NCE & \underline{.3121} & {\bf .3152} & .2515 & .2879 & .2879 & .1727 & .2333 \\
spect & 4 & .5887 & \underline{.6528} & .5887 & .6264 & .6000 & .6226 & .6264 & {\bf .6566} \\
spect & 8 & .6189 & .5962 & .6226 & \underline{.6566} & .6340 & .6302 & .6377 & {\bf .6792} \\
spect & 16 & .6642 & .6302 & .6566 & .6302 & .6453 & .6679 & \underline{.6906} & {\bf .7019} \\
spect & 32 & .6151 & .5811 & \underline{.6264} & .5283 & .5698 & .6189 & {\bf .6491} & .6113 \\
spect & 64 & .6792 & .6302 & .6868 & .6491 & {\bf .6943} & .6830 & \underline{.6906} & .6717 \\
diabetes & 4 & .6247 & .5753 & .6351 & .5987 & {\bf .6519} & \underline{.6481} & .6065 & .6442 \\
diabetes & 8 & .6312 & .6519 & .6468 & .6584 & .6182 & \underline{.6610} & .6247 & {\bf .7013} \\
diabetes & 16 & {\bf .7052} & .6610 & .6922 & .6766 & .6675 & .6857 & .6805 & \underline{.6922} \\
diabetes & 32 & .7286 & .6870 & {\bf .7325} & .6818 & .7130 & .7000 & .7013 & \underline{.7312} \\
diabetes & 64 & \underline{.7260} & .6844 & .7247 & .7013 & .7078 & .6961 & .7156 & {\bf .7338} \\
mice\_protein\_expression & 4 & \underline{.5620} & .4481 & {\bf .5907} & .4611 & .5481 & .5120 & .5519 & .3917 \\
mice\_protein\_expression & 8 & {\bf .7704} & .5639 & \underline{.7667} & .6204 & .7398 & .6417 & .6935 & .5111 \\
mice\_protein\_expression & 16 & {\bf .8935} & .7028 & .8602 & .7806 & \underline{.8611} & .5722 & .8185 & .6472 \\
mice\_protein\_expression & 32 & {\bf .9759} & .8398 & .9472 & .8787 & \underline{.9472} & .5898 & .9287 & .8009 \\
mice\_protein\_expression & 64 & {\bf .9944} & .9481 & .9685 & .9620 & .9787 & .7676 & \underline{.9898} & .9333 \\
Wilt & 4 & {\bf .6242} & .3277 & .5328 & \underline{.6166} & .5782 & .5442 & .5992 & .4547 \\
Wilt & 8 & {\bf .7745} & .5894 & .7032 & .5985 & .7076 & \underline{.7631} & .6773 & .6350 \\
Wilt & 16 & {\bf .9080} & .6027 & .8087 & .7720 & \underline{.8371} & .8046 & .8191 & .6104 \\
Wilt & 32 & {\bf .9629} & .6305 & .8099 & .8825 & .9270 & .9217 & \underline{.9345} & .7277 \\
Wilt & 64 & {\bf .9612} & .6752 & .8114 & .9130 & \underline{.9445} & .9318 & .9198 & .7351 \\
bank\_marketing & 4 & .5314 & \underline{.5668} & .5315 & .4687 & .5382 & .5324 & .5457 & {\bf .6180} \\
bank\_marketing & 8 & .4696 & .5296 & .4992 & .5216 & .4855 & .5409 & \underline{.5501} & {\bf .6252} \\
bank\_marketing & 16 & .5132 & .4002 & \underline{.5262} & .5120 & {\bf .5481} & .4992 & .4955 & .5246 \\
bank\_marketing & 32 & .4671 & .4660 & .4939 & .5084 & .4892 & .4856 & \underline{.5227} & {\bf .5845} \\
bank\_marketing & 64 & .3862 & .4776 & .4927 & .4934 & {\bf .5104} & .4604 & .4913 & \underline{.5059} \\
statlog\_german\_credit\_data & 4 & .5110 & .5260 & .5450 & .5120 & {\bf .5560} & .5190 & .5290 & \underline{.5500} \\
statlog\_german\_credit\_data & 8 & .4240 & .5060 & {\bf .5210} & .4990 & .5080 & \underline{.5120} & .5070 & .4810 \\
statlog\_german\_credit\_data & 16 & \underline{.5370} & .4600 & .5160 & .4900 & .5300 & .5100 & .5300 & {\bf .5550} \\
statlog\_german\_credit\_data & 32 & .5090 & .5360 & .5120 & .4810 & .5210 & \underline{.5390} & {\bf .5430} & .5280 \\
statlog\_german\_credit\_data & 64 & .4280 & .4980 & .5090 & {\bf .5280} & .5030 & .5110 & \underline{.5220} & .4980 \\
company\_bankruptcy\_prediction & 4 & \underline{.8468} & .6799 & .7248 & .7428 & .7342 & .8104 & .7569 & {\bf .8984} \\        
company\_bankruptcy\_prediction & 8 & \underline{.8346} & .7243 & .7400 & .7919 & .7519 & .7758 & .8227 & {\bf .8494} \\        
company\_bankruptcy\_prediction & 16 & .7977 & .8006 & .7299 & \underline{.8213} & .7465 & .7481 & .7689 & {\bf .8802} \\        
company\_bankruptcy\_prediction & 32 & .8229 & .7677 & .7787 & \underline{.8305} & .8018 & .7949 & .7990 & {\bf .8424} \\        
company\_bankruptcy\_prediction & 64 & .8375 & .8085 & .8103 & {\bf .8534} & .8147 & .8254 & .8073 & \underline{.8501} \\        
drug\_consumption & 4 & .1576 & .1289 & .1310 & \underline{.1607} & .1448 & .1533 & .1496 & {\bf .1682} \\
drug\_consumption & 8 & {\bf .1905} & \underline{.1480} & .1310 & .1273 & .1326 & .1310 & .1390 & .1443 \\
drug\_consumption & 16 & {\bf .1926} & .1576 & .1337 & .1512 & .1443 & .1576 & .1432 & \underline{.1724} \\
dry\_bean\_dataset & 4 & .7928 & .7920 & \underline{.8318} & .8122 & .7887 & {\bf .8452} & .7974 & .8208 \\
dry\_bean\_dataset & 8 & .8622 & .8488 & {\bf .8689} & .8320 & .8436 & \underline{.8667} & .8308 & .8540 \\
dry\_bean\_dataset & 16 & .8922 & .8724 & \underline{.8923} & .8765 & .8818 & {\bf .8959} & .8605 & .8801 \\
dry\_bean\_dataset & 32 & \underline{.9043} & .8764 & .9033 & .8923 & .9023 & {\bf .9066} & .8493 & .8914 \\
dry\_bean\_dataset & 64 & .9120 & .8977 & \underline{.9129} & .9069 & {\bf .9132} & .9098 & .8857 & .9088 \\
internet\_firewall & 4 & {\bf .8115} & .7007 & .6677 & \underline{.7788} & .6766 & .7575 & .7029 & .4492 \\
internet\_firewall & 8 & {\bf .8525} & .7968 & .6130 & \underline{.8356} & .5845 & .7540 & .8253 & .2812 \\
internet\_firewall & 16 & {\bf .8591} & .8095 & .6141 & \underline{.8506} & .6085 & .7789 & .8413 & .3870 \\
internet\_firewall & 32 & .8368 & {\bf .8797} & .7212 & \underline{.8724} & .7453 & .7725 & .8537 & .7996 \\
Basketball\_c & 4 & \underline{.6500} & .5784 & .6179 & .6388 & .6269 & .6478 & {\bf .6575} & .5985 \\
Basketball\_c & 8 & {\bf .6455} & .5761 & \underline{.6448} & .6313 & .6224 & .6433 & .6149 & .6216 \\
Basketball\_c & 16 & \underline{.6291} & .5806 & .6224 & .6179 & .5821 & .6037 & .5925 & {\bf .6358} \\
Basketball\_c & 32 & \underline{.6746} & .6425 & .6634 & .6119 & .6351 & .6530 & .6276 & {\bf .6806} \\
Basketball\_c & 64 & {\bf .6851} & .6157 & \underline{.6806} & .6224 & .5933 & .6090 & .6060 & .6724 \\
Bank\_Customer\_Churn\_Dataset & 4 & {\bf .6732} & .5641 & .6438 & .6377 & .6372 & .6427 & \underline{.6549} & .6341 \\
Bank\_Customer\_Churn\_Dataset & 8 & {\bf .6841} & .5434 & .6441 & .6640 & .6450 & .6577 & \underline{.6672} & .6458 \\
Bank\_Customer\_Churn\_Dataset & 16 & .6140 & .5766 & .5991 & .6201 & .5954 & {\bf .6474} & .6346 & \underline{.6449} \\
Bank\_Customer\_Churn\_Dataset & 32 & .7021 & .6442 & .6723 & \underline{.7128} & .6744 & .6992 & .6871 & {\bf .7370} \\
Bank\_Customer\_Churn\_Dataset & 64 & .6909 & .6330 & .6750 & .6697 & .6633 & \underline{.7040} & .6750 & {\bf .7283} \\
CDC\_Diabetes\_Health & 4 & .5901 & .6353 & .5802 & \underline{.6665} & .6291 & .6463 & .6509 & {\bf .7969} \\       
CDC\_Diabetes\_Health & 8 & .6290 & .6339 & .6110 & \underline{.6700} & .6214 & .6359 & .6439 & {\bf .6774} \\       
CDC\_Diabetes\_Health & 16 & .6645 & .6381 & .6280 & .6398 & .6322 & \underline{.6979} & .6409 & {\bf .7376} \\       
CDC\_Diabetes\_Health & 32 & .6927 & .6689 & \underline{.6940} & .6486 & .6657 & .6609 & .6510 & {\bf .7115} \\       
CDC\_Diabetes\_Health & 64 & .6959 & .6876 & \underline{.7024} & .6610 & .6796 & .6667 & .6528 & {\bf .7121} \\       
E-CommereShippingData & 4 & .5863 & .5680 & .5719 & .4852 & .5734 & .6009 & \underline{.6024} & {\bf .6118} \\
E-CommereShippingData & 8 & .6176 & .5451 & .5965 & {\bf .6583} & .5830 & .6183 & .5909 & \underline{.6365} \\
E-CommereShippingData & 16 & .6447 & .5717 & .6137 & \underline{.6472} & .6076 & .6322 & .6380 & {\bf .6544} \\
E-CommereShippingData & 32 & {\bf .6476} & .5593 & .6257 & .6435 & .5977 & \underline{.6471} & .6419 & .6432 \\
E-CommereShippingData & 64 & {\bf .6670} & .5822 & .6487 & .6595 & .6303 & .6569 & .6471 & \underline{.6610} \\
Fitness\_Club\_c & 4 & .6260 & .5873 & .6133 & {\bf .7567} & .6080 & .6733 & .6800 & \underline{.7180} \\
Fitness\_Club\_c & 8 & .6307 & .5180 & .5847 & \underline{.6593} & .5747 & .6333 & .6220 & {\bf .6820} \\
Fitness\_Club\_c & 16 & .6407 & .5793 & .6313 & \underline{.6547} & .6093 & .6460 & .6507 & {\bf .7733} \\
Fitness\_Club\_c & 32 & \underline{.7320} & .5440 & .7227 & .6773 & .6287 & .6647 & .6413 & {\bf .7447} \\
Fitness\_Club\_c & 64 & .7367 & .5893 & \underline{.7387} & .6727 & .6733 & .7173 & .6633 & {\bf .7547} \\
banknote\_authentication & 4 & .5149 & .5135 & .5098 & .5076 & .5149 & \underline{.5309} & .5236 & {\bf .5382} \\
banknote\_authentication & 8 & .4836 & .4771 & .4916 & .4844 & .4967 & .4887 & {\bf .5193} & \underline{.5025} \\
banknote\_authentication & 16 & .4938 & .4924 & .4785 & .4931 & .4975 & .4975 & {\bf .5033} & \underline{.5004} \\
banknote\_authentication & 32 & .4596 & .4931 & .4895 & .4953 & .4807 & .4749 & {\bf .5047} & \underline{.5004} \\
banknote\_authentication & 64 & .4531 & .4924 & .4858 & \underline{.5004} & .4924 & .4982 & .4895 & {\bf .5265} \\
kc2 & 4 & .6324 & \underline{.7505} & .6571 & .7181 & .6571 & .7143 & .6552 & {\bf .7790} \\
kc2 & 8 & .6305 & .7048 & .6743 & \underline{.7295} & .6876 & .6590 & .6838 & {\bf .7695} \\
kc2 & 16 & .6267 & .6476 & .6952 & .5943 & .6514 & {\bf .7067} & .6343 & \underline{.7029} \\
kc2 & 32 & .7067 & .7181 & .7143 & .6876 & .7371 & \underline{.7505} & .7086 & {\bf .7619} \\
kc2 & 64 & .7257 & .7067 & .7333 & .7181 & \underline{.7562} & .7390 & .7257 & {\bf .7752} \\
maternal\_health\_risk & 4 & {\bf .5567} & .5192 & .5350 & .4749 & .5222 & \underline{.5379} & .5271 & .5054 \\
maternal\_health\_risk & 8 & {\bf .5744} & .5488 & .5379 & .5222 & .5448 & \underline{.5685} & .5586 & .5310 \\
maternal\_health\_risk & 16 & {\bf .6158} & \underline{.6059} & .5833 & .5803 & .5744 & .5941 & .5813 & .5980 \\
maternal\_health\_risk & 32 & .6325 & .6266 & .6167 & {\bf .6483} & .6286 & \underline{.6463} & .6296 & .6217 \\
maternal\_health\_risk & 64 & .6562 & .6384 & .6079 & {\bf .6906} & .6335 & .6611 & \underline{.6788} & .6483 \\
seismic+bumps & 4 & .5954 & .5752 & .5838 & .4847 & .5729 & .5954 & \underline{.6723} & {\bf .7656} \\
seismic+bumps & 8 & .7168 & .5903 & .6135 & .6998 & .6816 & \underline{.7335} & .7087 & {\bf .7663} \\
seismic+bumps & 16 & .6662 & .6584 & .6685 & .5872 & .6561 & \underline{.6816} & .6333 & {\bf .7520} \\
seismic+bumps & 32 & \underline{.7435} & .7269 & .7300 & .6186 & .6990 & .7412 & .6925 & {\bf .7961} \\
seismic+bumps & 64 & .7238 & .7203 & \underline{.7516} & .6166 & .6878 & .7373 & .6712 & {\bf .7567} \\
sports\_articles\_for\_objectivity & 4 & {\bf .7590} & .6890 & \underline{.7520} & .7160 & .7510 & .7470 & .7480 & .7360 \\
sports\_articles\_for\_objectivity & 8 & .7320 & \underline{.7640} & .7180 & .6790 & .7360 & .7610 & .7150 & {\bf .7910} \\
sports\_articles\_for\_objectivity & 16 & .7970 & \underline{.8160} & .7780 & .7970 & .7780 & .8060 & .7860 & {\bf .8190} \\
sports\_articles\_for\_objectivity & 32 & {\bf .8410} & \underline{.8220} & .8200 & .8130 & .7970 & .7960 & .8000 & .8200 \\
sports\_articles\_for\_objectivity & 64 & \underline{.8320} & .8180 & .8060 & .7890 & .7810 & .7570 & .7710 & {\bf .8380} \\
turiye\_student\_evaluation & 4 & .2756 & .2426 & .3038 & {\bf .3368} & .2828 & \underline{.3086} & .2682 & .2826 \\
turiye\_student\_evaluation & 8 & .2404 & .2447 & .2825 & {\bf .3168} & .2687 & \underline{.2947} & .2811 & .2900 \\
turiye\_student\_evaluation & 16 & .2952 & .2658 & .2981 & \underline{.3163} & .2938 & .3134 & .2897 & {\bf .3247} \\
turiye\_student\_evaluation & 32 & {\bf .3832} & .2945 & .3593 & .3462 & .3380 & .3668 & .3234 & \underline{.3730} \\
turiye\_student\_evaluation & 64 & \underline{.3921} & .3297 & .3777 & .3514 & .3694 & .3893 & .3498 & {\bf .3959} \\
water\_quality & 4 & .6046 & {\bf .7191} & .6292 & \underline{.6956} & .6199 & .6416 & .6008 & .6699 \\
water\_quality & 8 & .5580 & {\bf .7280} & .5754 & .5920 & .5686 & .5749 & .6146 & \underline{.7164} \\
water\_quality & 16 & .6514 & .6438 & .6633 & .6430 & .6545 & \underline{.6980} & .6412 & {\bf .7060} \\
water\_quality & 32 & .6412 & .6586 & .6531 & .6520 & .6476 & \underline{.6666} & .6601 & {\bf .6857} \\
water\_quality & 64 & .7121 & .7165 & \underline{.7209} & .6807 & .7045 & .6946 & .6757 & {\bf .7396} \\
blood & 4 & .5867 & .5387 & .5653 & .5600 & .5627 & {\bf .5920} & \underline{.5880} & .5600 \\
blood & 8 & .6547 & .6413 & .6387 & .5733 & .6467 & {\bf .7040} & .6280 & \underline{.6600} \\
blood & 16 & .5600 & .5280 & .5467 & .4827 & .5387 & \underline{.5800} & .5080 & {\bf .6147} \\
blood & 32 & .6360 & .6427 & .6133 & .6040 & {\bf .6573} & \underline{.6547} & .6440 & .6387 \\
blood & 64 & \underline{.6733} & .6400 & .5893 & .6200 & {\bf .6880} & .6600 & .6427 & .6613 \\
breast-cancer & 4 & .5517 & .6310 & .5828 & \underline{.6379} & .5276 & .5621 & .5759 & {\bf .6483} \\
breast-cancer & 8 & {\bf .6276} & .5276 & .5379 & .5138 & .5586 & .5931 & .5690 & \underline{.5931} \\
breast-cancer & 16 & .5862 & .5483 & .5483 & .5345 & .5793 & .5690 & \underline{.6000} & {\bf .6069} \\
breast-cancer & 32 & \underline{.6586} & .6552 & .6517 & .5966 & .6552 & .6586 & .6379 & {\bf .6724} \\
breast-cancer-wisc & 4 & \underline{.9700} & .9143 & .9657 & .9586 & .9671 & {\bf .9743} & .9671 & .9371 \\
breast-cancer-wisc & 8 & \underline{.9729} & .9486 & .9671 & .9243 & .9629 & {\bf .9743} & .9443 & .9586 \\
breast-cancer-wisc & 16 & .9686 & .9571 & {\bf .9786} & .9371 & .9686 & \underline{.9700} & .9600 & .9629 \\
breast-cancer-wisc & 32 & .9629 & .9600 & {\bf .9686} & .9543 & .9629 & \underline{.9657} & .9571 & .9629 \\
breast-cancer-wisc & 64 & .9714 & .9586 & {\bf .9729} & \underline{.9714} & .9657 & .9714 & .9629 & .9700 \\
breast-cancer-wisc-prog & 4 & .5850 & .4000 & .5500 & {\bf .6750} & .6000 & .6050 & {\bf .6750} & .5700 \\
breast-cancer-wisc-prog & 8 & \underline{.5800} & .4250 & .5700 & .5500 & .5200 & {\bf .5850} & .5550 & .5500 \\
breast-cancer-wisc-prog & 16 & .6300 & .4750 & \underline{.6750} & .5850 & .6150 & .6250 & .6650 & {\bf .7100} \\
dermatology & 4 & \underline{.9514} & .8568 & .9324 & .9054 & .9324 & {\bf .9595} & .9189 & .9189 \\
dermatology & 8 & .9297 & .8757 & .9324 & {\bf .9649} & .9216 & \underline{.9486} & .9189 & .9243 \\
echocardiogram & 4 & \underline{.6593} & .6370 & .6222 & .5778 & .6370 & .6519 & .6296 & {\bf .7778} \\
echocardiogram & 8 & .6889 & .6963 & .6889 & .6148 & .6444 & .7111 & \underline{.7259} & {\bf .7333} \\
echocardiogram & 16 & .6074 & \underline{.7037} & .6222 & .6889 & .6148 & .6519 & .6889 & {\bf .7037} \\
\midrule
avg. rank  &  & 3.658 & 5.879 & 4.470 & 5.215 & 4.899 & 3.765 & 4.772 & {\bf 3.168 }
\label{tab:fewshot_main_whole}
\end{longtable}

\begin{longtable}{lccccccccc}
\caption{The complete results for the few-shot tasks on the 29 benchmark datasets (\autoref{fig:few-shot_results}). The absence of certain shot configurations in some datasets is attributed to the lack of sufficient training data for specific classes.} \\
\toprule
Dataset & Shot & TabPFN & $K$NN & LogReg & XGB & MLP & FT-T & MNCA & \ours \\
\toprule
\endfirsthead

\toprule
Dataset & Shot & TabPFN & $K$NN & LogReg & XGB & MLP & FT-T & MNCA & \ours \\
\toprule
\endhead

\toprule
\endfoot

FOREX\_audchf-day-High & 4 & .4905 & .4975 & .4779 & .4894 & .5041 & .4986 & {\bf .5106} & \underline{.5046} \\
FOREX\_audchf-day-High & 8 & {\bf .5411} & .4916 & .4872 & .4774 & .4877 & \underline{.4986} & .4948 & .4926 \\
FOREX\_audchf-day-High & 16 & {\bf .5662} & .5117 & .5019 & .5046 & .5123 & \underline{.5166} & .5084 & .5101 \\
FOREX\_audchf-day-High & 32 & {\bf .6229} & .5041 & .5335 & .5090 & .5395 & .5166 & \underline{.5460} & .5155 \\
FOREX\_audchf-day-High & 64 & {\bf .6703} & .4986 & \underline{.5929} & .5362 & .5864 & .5575 & .5662 & .5215 \\
taiwanese\_bankruptcy\_prediction & 4 & .7575 & {\bf .8314} & .7301 & .5629 & .7240 & .6730 & .7079 & \underline{.7943} \\
taiwanese\_bankruptcy\_prediction & 8 & \underline{.8114} & .7911 & .7563 & .7044 & .7622 & .7383 & .7606 & {\bf .8315} \\
taiwanese\_bankruptcy\_prediction & 16 & \underline{.7982} & .7630 & .7456 & .7122 & .7352 & .7358 & .7292 & {\bf .8226} \\
taiwanese\_bankruptcy\_prediction & 32 & \underline{.8444} & .8161 & .8029 & .8323 & .8104 & .8157 & .8265 & {\bf .8513} \\
taiwanese\_bankruptcy\_prediction & 64 & \underline{.8340} & .8312 & .7982 & .8304 & .7949 & .8145 & .8245 & {\bf .8402} \\
rl & 4 & .5109 & .5093 & .4972 & {\bf .5360} & .5115 & .5177 & .5139 & \underline{.5272} \\
rl & 8 & .5145 & .5030 & .5109 & \underline{.5378} & .5169 & .5181 & {\bf .5384} & .5290 \\
rl & 16 & .5431 & .5097 & .5332 & {\bf .5602} & .5229 & \underline{.5573} & .5372 & .5429 \\
rl & 32 & .5676 & .5465 & .5825 & {\bf .5893} & .5648 & \underline{.5891} & .5600 & .5873 \\
rl & 64 & .5575 & .5529 & .5785 & {\bf .5960} & .5771 & \underline{.5831} & .5487 & .5807 \\
pc3 & 4 & .6000 & .6249 & .5981 & {\bf .7495} & .6013 & .6134 & .6294 & \underline{.7316} \\
pc3 & 8 & .6281 & .6211 & \underline{.6383} & .6045 & .6211 & .6115 & .5725 & {\bf .7693} \\
pc3 & 16 & {\bf .7610} & .6665 & .7297 & .7214 & .7214 & .7099 & .6716 & \underline{.7399} \\
pc3 & 32 & .7482 & .7259 & .7482 & .7444 & \underline{.7546} & .7246 & .7284 & {\bf .7936} \\
pc3 & 64 & .7585 & .7431 & \underline{.7783} & .7214 & .7457 & .7150 & .7118 & {\bf .7815} \\
eye\_movements\_bin & 4 & .4924 & \underline{.5083} & .4950 & .4971 & .4987 & .5060 & .5034 & {\bf .5162} \\
eye\_movements\_bin & 8 & {\bf .5204} & .5004 & .5158 & .5137 & \underline{.5184} & .5168 & .5120 & .5087 \\
eye\_movements\_bin & 16 & {\bf .5198} & .5025 & .5084 & .5183 & .5176 & \underline{.5188} & .5138 & .5159 \\
eye\_movements\_bin & 32 & .5167 & .5030 & .5205 & {\bf .5213} & .5110 & {\bf .5213} & .5158 & .5172 \\
eye\_movements\_bin & 64 & .5059 & .5089 & .5164 & \underline{.5189} & {\bf .5217} & .5180 & .5133 & .5067 \\
BNG(breast-w) & 4 & {\bf .9632} & .9248 & .9585 & .8297 & \underline{.9592} & .9552 & .9451 & .9177 \\
BNG(breast-w) & 8 & {\bf .9720} & .9573 & .9674 & .8781 & .9668 & \underline{.9702} & .9583 & .9577 \\
BNG(breast-w) & 16 & {\bf .9729} & .9659 & \underline{.9724} & .9411 & .9668 & .9707 & .9653 & .9644 \\
BNG(breast-w) & 32 & \underline{.9705} & .9672 & {\bf .9713} & .9506 & .9643 & .9703 & .9621 & .9653 \\
BNG(breast-w) & 64 & {\bf .9757} & .9702 & \underline{.9748} & .9673 & .9713 & .9743 & .9726 & .9707 \\
FOREX\_cadjpy-hour-High & 4 & .5041 & .5012 & .5088 & \underline{.5145} & .4994 & .5042 & .5089 & {\bf .5183} \\
FOREX\_cadjpy-hour-High & 8 & \underline{.5052} & .4937 & .5022 & .5016 & .5014 & .5022 & .5044 & {\bf .5123} \\
FOREX\_cadjpy-hour-High & 16 & .4920 & .4955 & .4967 & \underline{.5015} & .4975 & .4958 & .4996 & {\bf .5069} \\
FOREX\_cadjpy-hour-High & 32 & .5013 & .5049 & {\bf .5216} & .5061 & \underline{.5104} & .5069 & .5075 & .5069 \\
FOREX\_cadjpy-hour-High & 64 & .4803 & .4995 & {\bf .5266} & .5066 & .5082 & \underline{.5132} & .4981 & .5082 \\
dis & 4 & .5044 & .1955 & .4789 & .4858 & .4901 & {\bf .5701} & \underline{.5229} & .3807 \\
dis & 8 & .5272 & .2011 & .4832 & .5730 & .5097 & \underline{.5894} & {\bf .7420} & .3738 \\
dis & 16 & .6374 & .3560 & .5396 & {\bf .7926} & .5608 & .6747 & \underline{.7870} & .6461 \\
dis & 32 & .7073 & .3857 & .5581 & {\bf .8458} & .6021 & \underline{.8366} & .7799 & .7330 \\
sylvine & 4 & .5940 & .5746 & .5922 & .4911 & .5819 & \underline{.6328} & .6113 & {\bf .7337} \\
sylvine & 8 & .6697 & .6211 & .6771 & \underline{.7883} & .6564 & .7770 & .7282 & {\bf .8361} \\
sylvine & 16 & .7696 & .6630 & .7641 & {\bf .8966} & .7555 & .8603 & .8174 & \underline{.8730} \\
sylvine & 32 & .8353 & .6800 & .8238 & {\bf .9079} & .8094 & .8849 & .8535 & \underline{.8915} \\
sylvine & 64 & .8782 & .7221 & .8714 & \underline{.8972} & .8517 & .8741 & .8527 & {\bf .9069} \\
BNG(tic-tac-toe) & 4 & .4640 & .5287 & .5368 & {\bf .6084} & .5276 & .5473 & .5297 & \underline{.5553} \\
BNG(tic-tac-toe) & 8 & .5380 & .5794 & .5878 & .5891 & .5824 & \underline{.5938} & {\bf .5974} & .5835 \\
BNG(tic-tac-toe) & 16 & .5626 & \underline{.6045} & .5986 & .5874 & .6037 & .5969 & .5918 & {\bf .6205} \\
BNG(tic-tac-toe) & 32 & .5816 & .6118 & {\bf .6281} & .6275 & \underline{.6279} & .6136 & .6065 & .6129 \\
BNG(tic-tac-toe) & 64 & .6067 & .6441 & .6429 & .6564 & {\bf .6653} & \underline{.6638} & .6281 & .6558 \\
online\_shoppers & 4 & .7454 & .6935 & .7797 & .7326 & \underline{.7927} & .7596 & .7461 & {\bf .8185} \\
online\_shoppers & 8 & .6531 & .6394 & .7145 & \underline{.8147} & .7056 & .7309 & .7215 & {\bf .8221} \\
online\_shoppers & 16 & .6103 & .7574 & .7085 & \underline{.7867} & .7015 & .7384 & .7366 & {\bf .8016} \\
online\_shoppers & 32 & .7431 & .6633 & .7065 & .7868 & .6921 & .7740 & \underline{.7927} & {\bf .8493} \\
online\_shoppers & 64 & .7842 & .6664 & .6825 & \underline{.8114} & .6713 & .7481 & .7921 & {\bf .8525} \\
Cardiovascular-Disease-dataset & 4 & .5992 & .5732 & .5938 & .5679 & .5804 & \underline{.6295} & .6182 & {\bf .6353} \\
Cardiovascular-Disease-dataset & 8 & {\bf .6587} & .5821 & \underline{.6574} & .6320 & .6326 & .6522 & .6175 & .6391 \\
Cardiovascular-Disease-dataset & 16 & \underline{.6683} & .6044 & .6542 & .6506 & .6403 & .6536 & .6358 & {\bf .6893} \\
Cardiovascular-Disease-dataset & 32 & {\bf .6763} & .5936 & .6542 & .6471 & .6285 & .6536 & .6231 & \underline{.6739} \\
Cardiovascular-Disease-dataset & 64 & \underline{.6992} & .6186 & .6945 & .6487 & .6592 & .6873 & .6352 & {\bf .7077} \\
credit & 4 & \underline{.6669} & .6385 & .6651 & .5514 & {\bf .6716} & .6523 & .6500 & .6477 \\
credit & 8 & \underline{.6966} & .6239 & .6815 & .6753 & .6772 & {\bf .7161} & .6596 & .6957 \\
credit & 16 & \underline{.6884} & .6624 & .6817 & .5947 & .6588 & .6538 & .6442 & {\bf .7067} \\
credit & 32 & {\bf .7335} & .6853 & \underline{.7298} & .6815 & .7167 & .7135 & .6841 & .7100 \\
credit & 64 & {\bf .7530} & .6788 & \underline{.7301} & .7005 & .7051 & .7140 & .6912 & .7250 \\
FOREX\_audsgd-hour-High & 4 & .5053 & .4923 & \underline{.5071} & {\bf .5137} & .5060 & .5024 & .4965 & .5051 \\
FOREX\_audsgd-hour-High & 8 & .4961 & .4873 & \underline{.4981} & .4951 & .4940 & .4938 & .4928 & {\bf .5022} \\
FOREX\_audsgd-hour-High & 16 & \underline{.5093} & .5081 & {\bf .5098} & .4998 & .5081 & .5067 & .5055 & .5077 \\
FOREX\_audsgd-hour-High & 32 & .4953 & .4962 & .5009 & {\bf .5028} & .5011 & \underline{.5014} & .5000 & .5004 \\
FOREX\_audsgd-hour-High & 64 & .5000 & .5027 & .4998 & .5020 & {\bf .5085} & \underline{.5046} & .5046 & .5025 \\
waveform-5000 & 4 & .6144 & .6238 & .6552 & .6290 & {\bf .6742} & .6502 & .6410 & \underline{.6728} \\
waveform-5000 & 8 & .6462 & .6864 & \underline{.7202} & .7010 & {\bf .7296} & .7012 & .6868 & .7128 \\
waveform-5000 & 16 & .7234 & .6994 & .7636 & {\bf .7852} & \underline{.7802} & .7560 & .7532 & .7702 \\
waveform-5000 & 32 & .8014 & .7242 & .7840 & \underline{.8030} & {\bf .8056} & .7854 & .7512 & .7992 \\
waveform-5000 & 64 & {\bf .8334} & .7392 & .8102 & .8250 & .8228 & .7940 & .7804 & \underline{.8320} \\
jungle\_chess\_2pcs\_raw\_endgame & 4 & \underline{.4719} & .4269 & .4629 & .4473 & .4645 & {\bf .4719} & .4515 & .4572 \\
jungle\_chess\_2pcs\_raw\_endgame & 8 & {\bf .4943} & .4749 & .4866 & .4709 & .4639 & \underline{.4938} & .4574 & .4813 \\
jungle\_chess\_2pcs\_raw\_endgame & 16 & .5650 & .5079 & .5586 & .5654 & \underline{.5655} & {\bf .5835} & .5248 & .5386 \\
jungle\_chess\_2pcs\_raw\_endgame & 32 & .6045 & .5486 & .5902 & \underline{.6275} & .6028 & {\bf .6325} & .5737 & .5777 \\
jungle\_chess\_2pcs\_raw\_endgame & 64 & .6590 & .5886 & .6037 & .6612 & \underline{.6670} & {\bf .6841} & .6072 & .6051 \\
BNG(cmc) & 4 & .3686 & {\bf .3794} & .3717 & .3383 & .3705 & .3729 & .3670 & \underline{.3773} \\
BNG(cmc) & 8 & \underline{.4186} & .3758 & .3958 & .4170 & .4022 & {\bf .4274} & .4042 & .4066 \\
BNG(cmc) & 16 & {\bf .4658} & .3898 & \underline{.4441} & .4351 & .4279 & .4414 & .4263 & .4130 \\
BNG(cmc) & 32 & {\bf .4836} & .4146 & .4582 & \underline{.4642} & .4453 & .4639 & .4373 & .4508 \\
BNG(cmc) & 64 & {\bf .4981} & .4167 & .4688 & .4652 & .4450 & \underline{.4740} & .4307 & .4535 \\
page-blocks & 4 & .6557 & {\bf .7352} & .6732 & .6782 & .5673 & .6970 & \underline{.7118} & .6857 \\
page-blocks & 8 & .7954 & .7184 & {\bf .8237} & \underline{.8225} & .7934 & .7934 & .7671 & .8040 \\
page-blocks & 16 & .8617 & .8278 & {\bf .8658} & .8349 & .8489 & .8157 & .8311 & \underline{.8650} \\
segment & 4 & {\bf .7273} & .6658 & .6948 & .6442 & \underline{.7009} & .6948 & .6870 & .6528 \\
segment & 8 & {\bf .8004} & .7195 & .7584 & .7593 & \underline{.7675} & .7641 & .7459 & .7385 \\
segment & 16 & {\bf .8558} & .7671 & .7870 & .8143 & .8091 & \underline{.8429} & .8100 & .7918 \\
segment & 32 & {\bf .8861} & .8190 & .8277 & .8532 & .8662 & \underline{.8706} & .8459 & .8338 \\
segment & 64 & {\bf .9035} & .8481 & .8481 & .8861 & .8918 & \underline{.9013} & .8645 & .8654 \\
website\_phishing & 4 & .5845 & .5454 & {\bf .6317} & .5926 & .6155 & .5970 & .6118 & \underline{.6236} \\
website\_phishing & 8 & \underline{.7107} & .6317 & .6745 & .7100 & .6782 & .6952 & .6760 & {\bf .7292} \\
website\_phishing & 16 & .7668 & .6583 & .7247 & \underline{.7668} & .7528 & {\bf .7712} & .7336 & .7203 \\
website\_phishing & 32 & .7970 & .7137 & .7727 & {\bf .8089} & \underline{.7978} & .7970 & .7801 & .7720 \\
website\_phishing & 64 & .8310 & .7520 & .8185 & {\bf .8590} & .8472 & \underline{.8509} & .8015 & .7852 \\
baseball & 4 & .8127 & .7993 & .8179 & .8537 & .8134 & \underline{.8575} & .8515 & {\bf .8881} \\
baseball & 8 & .8597 & .8604 & .8448 & .8813 & .8336 & \underline{.8851} & .8299 & {\bf .9209} \\
baseball & 16 & \underline{.8276} & .8254 & .8142 & .8179 & .8246 & .7776 & .7955 & {\bf .8903} \\
baseball & 32 & .8396 & .8037 & .8507 & .8567 & \underline{.8649} & .8567 & .8299 & {\bf .8873} \\
pendigits & 4 & \underline{.8006} & .7617 & .7828 & .6626 & {\bf .8038} & .7783 & .7154 & .6445 \\
pendigits & 8 & \underline{.8724} & .8272 & .8345 & .7669 & {\bf .8760} & .8392 & .8163 & .7588 \\
pendigits & 16 & {\bf .9361} & .9008 & .8681 & .8581 & \underline{.9269} & .9134 & .8895 & .8645 \\
pendigits & 32 & {\bf .9604} & .9327 & .8980 & .9146 & \underline{.9521} & .9500 & .9321 & .9119 \\
pendigits & 64 & {\bf .9752} & .9543 & .9165 & .9469 & .9656 & \underline{.9663} & .9576 & .9379 \\
Gender\_Gap\_in\_Spanish\_WP & 4 & .3446 & \underline{.3806} & .3501 & .3080 & .3141 & .3549 & .3307 & {\bf .3819} \\
Gender\_Gap\_in\_Spanish\_WP & 8 & .3598 & .3571 & .3533 & \underline{.3762} & .3082 & .3579 & .3697 & {\bf .4166} \\
Gender\_Gap\_in\_Spanish\_WP & 16 & .3796 & \underline{.4065} & .3789 & .3777 & .3975 & .3817 & .3909 & {\bf .4356} \\
Gender\_Gap\_in\_Spanish\_WP & 32 & {\bf .4274} & .3756 & .3792 & .3756 & .3811 & \underline{.4023} & .3716 & .3941 \\
Gender\_Gap\_in\_Spanish\_WP & 64 & {\bf .4080} & .3855 & .3512 & \underline{.3964} & .3964 & .3880 & .3836 & .3893 \\
satimage & 4 & .7613 & .7345 & \underline{.7840} & .6782 & {\bf .7874} & .7779 & .7426 & .6927 \\
satimage & 8 & {\bf .8180} & .8051 & .7927 & .7639 & \underline{.8096} & .8076 & .7843 & .7795 \\
satimage & 16 & \underline{.8219} & .8171 & .8026 & .7871 & .8188 & .8196 & .8022 & {\bf .8255} \\
satimage & 32 & .8443 & \underline{.8454} & .8247 & .8260 & .8423 & .8364 & .8320 & {\bf .8597} \\
satimage & 64 & \underline{.8678} & .8541 & .8350 & .8628 & .8631 & .8631 & .8432 & {\bf .8770} \\
mfeat-fourier & 4 & .5820 & .5355 & .6270 & .5825 & .6280 & \underline{.6745} & .6215 & {\bf .6845} \\
mfeat-fourier & 8 & .6825 & .6460 & .6950 & .6620 & .6945 & \underline{.7315} & .6815 & {\bf .7435} \\
mfeat-fourier & 16 & .7215 & .7170 & .7415 & .7550 & .7470 & \underline{.7850} & .7235 & {\bf .8110} \\
mfeat-fourier & 32 & .7890 & .7740 & .7760 & .7965 & .8030 & \underline{.8145} & .7785 & {\bf .8320} \\
mfeat-fourier & 64 & .8160 & .8170 & .8095 & .8255 & .8240 & \underline{.8435} & .8120 & {\bf .8510} \\
VulNoneVul & 4 & .6945 & \underline{.8462} & .7087 & .5742 & .7152 & .6446 & .6911 & {\bf .8732} \\
VulNoneVul & 8 & .6911 & .6739 & .7136 & .6999 & .6687 & \underline{.7531} & .7066 & {\bf .8200} \\
VulNoneVul & 16 & .7025 & \underline{.8267} & .7935 & .6565 & .7470 & .7317 & .6729 & {\bf .8371} \\
law-school-admission-bianry & 4 & .7502 & .5746 & .6935 & {\bf .8724} & .6842 & \underline{.8622} & .8286 & .7272 \\
law-school-admission-bianry & 8 & .7575 & .6261 & .7080 & {\bf .9664} & .6874 & \underline{.9237} & .8837 & .9006 \\
law-school-admission-bianry & 16 & .8718 & .6934 & .8423 & {\bf .9832} & .8017 & \underline{.9820} & .9367 & .9519 \\
law-school-admission-bianry & 32 & .9510 & .7526 & .9192 & {\bf 1.000} & .8861 & \underline{.9997} & .9900 & .9734 \\
law-school-admission-bianry & 64 & .9582 & .7583 & .9343 & {\bf 1.000} & .9074 & \underline{1.000} & .9953 & .9541 \\
KDD & 4 & .5168 & .5124 & .5227 & \underline{.5503} & .5128 & {\bf .5579} & .5351 & .5416 \\
KDD & 8 & .6371 & .5668 & .6242 & {\bf .6918} & .6191 & .6669 & .6252 & \underline{.6723} \\
KDD & 16 & .6310 & .5686 & .6207 & {\bf .6878} & .6026 & .6385 & .6071 & \underline{.6582} \\
KDD & 32 & .6896 & .6127 & .6620 & {\bf .7349} & .6608 & .6860 & .6548 & \underline{.7327} \\
KDD & 64 & .7259 & .6189 & .6731 & {\bf .7605} & .6763 & \underline{.7311} & .6544 & .7152 \\
\midrule
avg. rank & & 3.828 & 6.440 & 4.716 & 4.403 & 4.463 & 3.440 & 5.284 & {\bf 3.388}
\label{tab:fewshot_tinybench_whole}
\end{longtable}

\end{appendices}

\end{document}